\documentclass[preprint,authoryear,3p]{elsarticle}

\input{epsf}

\usepackage{bold-extra}
\usepackage[english]{babel}
\usepackage[latin1]{inputenc}
\usepackage{natbib}
\usepackage{amssymb}
\usepackage{amsmath}  
\usepackage{color}    
\usepackage{moreverb} 
\usepackage{rotating}
\usepackage{tabularx}
\usepackage{stmaryrd}

\usepackage{pdfpages}
\usepackage{morefloats}
\usepackage{graphics}
\usepackage{bbm}
\usepackage{epsfig}
\usepackage{mathrsfs}
\usepackage{multirow}
\usepackage{amsthm}
\usepackage{graphicx} 

\newtheorem{df}{Definition}
\newtheorem{thm}{Theorem}

\newproof{pf}{Proof}

\newcommand{\abs}[1]{\vert #1 \vert}
\newcommand{\sentences}[1]{\llbracket #1 \rrbracket}
\newcommand{\sentencesw}[1]{\llceil #1 \rrceil}
\newcommand{\absd}[1]{\Vert #1 \Vert}

\newcommand{\ldinfty}[0]{\infty \hspace*{-.3cm} \infty}
\newcommand{\dinfty}[0]{\infty \hspace*{-.325cm} \infty}

\journal{Computer Speech and Language}

\begin{document}

\begin{frontmatter}

\title{Modeling of learning curves with applications to {\sc pos}
  tagging}

\author[UVigo]{Manuel Vilares Ferro\corref{cor}}
\cortext[cor]{Corresponding author: tel. +34 988 387280, fax +34 988 387001.}
\ead{vilares@uvigo.es}
\author[UVigo]{V\'ictor M. Darriba Bilbao}
\ead{darriba@uvigo.es}
\author[UVigo]{Francisco J. Ribadas Pena}
\ead{ribadas@uvigo.es}

\address[UVigo]{Department of Computer Science, University of Vigo \\ Campus As Lagoas s/n, 32004 -- Ourense, Spain}

\begin{abstract}

An algorithm to estimate the evolution of learning curves on the whole
of a training data base, based on the results obtained from a portion
and using a functional strategy, is introduced. We approximate
iteratively the sought value at the desired time, independently of the
learning technique used and once a point in the process, called
prediction level, has been passed. The proposal proves to be
formally correct with respect to our working hypotheses and includes a
reliable proximity condition. This allows the user to fix a
convergence threshold with respect to the accuracy finally achievable,
which extends the concept of stopping criterion and seems to be
effective even in the presence of distorting observations.

Our aim is to evaluate the training effort, supporting decision making
in order to reduce the need for both human and computational resources
during the learning process. The proposal is of interest in at least
three operational procedures. The first is the anticipation of
accuracy gain, with the purpose of measuring how much work is needed
to achieve a certain degree of performance. The second relates the
comparison of efficiency between systems at training time, with the
objective of completing this task only for the one that best suits our
requirements. The prediction of accuracy is also a valuable item of
information for customizing systems, since we can estimate in advance
the impact of settings on both the performance and the development
costs. Using the generation of part-of-speech taggers as an example
application, the experimental results are consistent with our
expectations.

\end{abstract}


\end{frontmatter}

\section{Introduction}
\label{section-introduction}

Creating a labeled training data base often leads to an expensive and
time-consuming task, even more so when we are talking about processes
involving new application domains where the resources are scarce or
even non-existent. Added to this are the usually high costs of the
training process itself, placing us at the origin of a bottleneck in
the generation of tools based on \textit{machine learning} ({\sc ml}),
as in the case of classification. The growing popularity of these
techniques multiplies the need for training material, and reducing the
efforts for its creation and further processing becomes a major
challenge. A particularly sensitive area of work to these
inconveniences is \textit{natural language processing} ({\sc nlp}),
the components of which are increasingly based on {\sc
  ml}~\citep{Biemann:2006:UPT:1557856.1557859,KATRIN08.335}. The
problem is specially delicate on \textit{part-of-speech} ({\sc pos})
tagging, because it relies on the complexity of both the annotation
task and the relations to be captured from learning, but also because
it serves as a first step for other {\sc nlp} functionalities such as
parsing and semantic analysis, so errors at this stage can lower their
performance~\citep{Song:2012:CSP:2390524.2390661}.

One way to save time and resources without loss of learning power is
to anticipate the working configuration that best suits our needs,
avoiding long training processes. This gives a practical meaning to
our work, also providing an objective basis for defining model
selection
criteria~\citep{akaike1973,akaike1974,BurnhamAnderson02,Hurvich1989,Lebreton92}
and model shrinking strategies~\citep{ChenMRSS09,SarikayaCDJ14}. Both
are issues studied in practice from an statistical point of view,
whose errors are expressed in probabilistic
terms~\citep{Blumer:1987:OR:31168.31174,floyd-95,McAllester:1999:PMA:307400.307435},
producing results which have proved to be quite
loose~\citep{Langford:2005:TPP:1046920.1058111} and therefore
justifying the exploration of new treatment options.

To do so, we must first identify both the parameter that will serve as
measure of performance for the system we are generating, generally
called \textit{accuracy}, and the factors that impact
it. Unfortunately, the number and complexity of these latter is often
such that it becomes impossible to weight them. Instead, the most
practical approach may be to make our own judgments in the context of
the task that we are trying to accomplish~\citep{VanHalteren1999}. A
way to do this is to consider the learning process as a single task to
study because it compiles, relates and processes all the information
provided by the user. This in turn means identifying the nature of the
learning signals, in order to avoid inappropriate comparisons between
{\sc ml} approaches using dissimilar ones, since to a large extent it
determines effectiveness. To that end, learning methods are typically
classified into three categories: supervised, semi-supervised and
unsupervised. The latter constitute the only possible choice when no
labelled data are available, or an alternative when they are expensive
to prepare, even though no useful accuracies have been
achieved~\citep{Li2012}. On the contrary, supervised strategies have
proved to be efficient, but they require a large amount of training
input that usually needs to be hand-annotated by human experts, which
is an intensive task in terms of time and expertise. For its part, the
semi-supervised procedures aim to combine the advantages of the
previous ones when only few labelled data are available. Although the
results have been mostly negative~\citep{Sogaard2010}, the research
effort is significant in the case of \textit{active learning} ({\sc
  al})~\citep{Cohn:1994:IGA:189256.189489,Seung:1992:QC:130385.130417},
an iterative approach that interacts with the environment in each
cycle, selecting for annotation the instances which are harder to
identify. Since the latter are assumed to be the most informative
ones, the expected result is the acceleration of the learning
process. Unfortunately, despite the potential of {\sc al}, its
adoption in practice is
questionable~\citep{Attenberg:2011:ILD:1964897.1964906}. In any case
and having made these comments, the study of the accuracy progress
during the learning process through the corresponding \textit{learning
  curve} is possible, which is our starting point.

Looking ahead to {\sc pos} tagging as domain to illustrate the
discussion, the way to evaluate its reliability is to determine how
many of the tags provided are correct, and how many superfluous ones
are eliminated~\citep{VanHalteren1999} in the case of ambiguous
outputs. The absence of the latter simplifies such a evaluation, which
allows us to speak of accuracy in {\sc pos}
tagging~\citep{DeRose1988}. With regard to the factors that influence
it, we should include any variation affecting the linguistic
resources, the tag-set and the method of evaluation
considered. Finally, {\sc pos} tagging based on learning uses a
statistical model built from labelled and/or unlabelled training data,
so that we can consider
unsupervised~\citep{goldwater-griffiths:2007:ACLMain,Merialdo:1994:TET:972525.972526,Ravi2009},
supervised~\citep{Brants2000,Brill1995a,Daelemans1996a,Gimenez2004,Schmid1994,Toutanova2003}
and semi-supervised~\citep{Sogaard2010,Spoustova2009} approaches. In
this latter case, the popularity of {\sc al} is growing in {\sc pos}
tagging
tasks~\citep{Dagan95committee-basedsampling,Haertel:2008:ACS:1557690.1557708,Neubig:2011:PPR:2002736.2002841,Ringger:2007:ALP:1642059.1642075}.

Returning to the general case, one major question for both supervised
and unsupervised approaches is the implementation of a condition to
halt the learning process, namely to detect when it has reached its
maximum or is sufficiently close for our purposes. The first goal
requires a \textit{stopping
  criterion}~\citep{Provost:1999:EPS:312129.312188}, while the second
one implies a more general condition, referred to in this paper as
\textit{proximity criterion} and the consideration of which is a
novelty in the state of the art, to the best of our knowledge. The
same applies to semi-supervised strategies, although we must also add
here the difficulty of designing the mechanism for selecting the instances
to be annotated in every iteration when {\sc al} techniques are
involved, which lies at the root of users' reluctance to adopt these
methods. 

All of that places the definition of stable proximity criteria at the
core of our proposal. We can then talk about its \textit{correctness}
with respect to our working hypotheses when the prior estimation for
accuracy is formally guaranteed in that context. We should also
provide a certain capacity for assimilating the fluctuations in
learning conditions without compromising the correctness, a phenomenon
referred to as \textit{variance}~\citep{Breiman96}, in order to
generate stable results. This is what we call the \textit{robustness}
of the model. Both issues, correctness and robustness, focus our
attention in this paper, the structure of which is described
below. Firstly, Section~\ref{section-state-of-the-art} examines the
methodologies serving as inspiration to solve the question
posed. Next, Section~\ref{section-formal-framework} reviews the
mathematical basis necessary to support our proposal, which we
introduce in Section~\ref{section-abstract-model}. In
Section~\ref{section-testing-frame}, we describe the testing frame for
the experiments illustrated in
Section~\ref{section-experiments}. Finally,
Section~\ref{section-conclusions} presents our final conclusions.

\section{The state of the art}
\label{section-state-of-the-art}

We briefly review now the main research lines in dealing with both
correctness and robustness in the prediction of learning curves,
highlighting their contributions and limitations, as well as their
application on the {\sc nlp} domain. This serve as reference for
contextualizing our work.

\subsection{Working on correctness}

Here, the focus has been given to the convergence, leaving aside the
definition of proximity criteria.  Thus, current {\sc ml} approaches
often take for granted a set of hypotheses ensuring it, such as the
access to independent and identically distributed
observations~\citep{Domingo:2002:ASM:593433.593526,Schutze:2006:PTP:1183614.1183709,KATRIN08.335}
to sample from. This allows us to assume that any learning curve is
monotonic and, given that it is always bounded, the learning process
has a supremum for accuracy and its convergence is
guaranteed. Accordingly, the attention of researchers turns to the
definition of stopping conditions in order to halt the training
procedure once this value has been identified. On the basis that such
functions have an initial steeply sloping portion, a more gently
sloping middle portion, and a final
plateau~\citep{Meek:2002:LSM:944790.944798}, the problem reduces to
detect the latter. Initially treated from a statistical
perspective~\citep{John96staticversus,Valiant:1984:TL:1968.1972}, this
approach faces the approximation of complete learning curves as a
major challenge because their complex functional forms often represent
the plateau as an infinite slight
incline~\citep{FreyFisher99,Last:2009:IDM:1557019.1557076}. So, we may
fit the early part of the curve well, but not the final one. Although
a simple way for minimizing this risk is to increase the number and
size of the sample from which it is built, we should then take into
account the impact in terms of computational efficiency.

For seeking a proper cost/benefit trade-off in the construction of
learning curves, the researchers usually apply the principle of
\textit{maximum expected utility}, which implies expressing the
problem in terms of \textit{decision theory}~\citep{Howard66}. In
practice, the approach differs depending upon the degree of control by
the user on the process. In the absence of this control, the final
cost can be defined as the sum of training data, error and model
induction charges, although the interpretation may vary according to
the strategy considered. The early works try to minimize induction and
error costs while ignoring those of data
acquisition~\citep{Provost:1999:EPS:312129.312188}, typically using a
linear regression function, whose slope is compared to zero. This flaw
was overcome by later proposals, first partially by assuming that the
cost of cases limited the amount of training data and this amount is
already specified~\citep{Weiss:2003:LTD:1622434.1622445} to later
eliminate this restriction~\citep{WeissTian08} and even calculate the
data cost of each label feature separately through a cost-sensitive
learner~\citep{Sheng:2007:PEA:1281192.1281261}. Nonetheless, finding
the global optimum of total cost can in no case be
guaranteed~\citep{Last:2009:IDM:1557019.1557076}, and it is not
infrequent that one wants to stop only when the desired degree of
accuracy is met, which would bring us back to the consideration of
pure stopping criteria.

A separate case is that of {\sc al}, where we select the training
examples to label and even quantify costs associated to specific
feature values, which lead us to take into account the cost of teacher
and the cost of tests~\citep{Turney02}, respectively. However, since
much of the work on {\sc al} assumes a fixed budget~\citep{Kapoor05},
practical stopping criteria often rely simply on measuring the
confidence of the learning process. Given that most of the research
focuses on \textit{pool-based active
  learning}~\citep{Lewis:1994:SAT:188490.188495}, in which the
selection is made from a pool, such a measure applies either over a
separate data set~\citep{Vlachos:2008:SCA:1349893.1350099} or over the
unlabelled data pool~\citep{Zhu:2012:UAL:2093153.2093154}. The
assumption in the first case is that we cannot take advantage of the
remaining instances in the pool when that confidence drops, while in
the second one the starting point is the uncertainty of the
classifier~\citep{Roy:2001:TOA:645530.655646}. The point where the
pool becomes uninformative can also be determined through the gradient
of the performance~\citep{Laws:2008:SCA:1599081.1599140}, whose rising
estimation slows to an almost horizontal slope at about the time when
the learning reaches its peak. We then stop the process when that
gradient approaches zero.

\subsection{Working on robustness}

Turning now to robustness, proposals pass through the generation of
different versions (weak predictors) of the learning curve by changing
the distribution of the training repeatedly, which makes it possible
to combine by aggregation the set of hypotheses so generated. In the
case of \textit{bagging}\footnote{For \textit{bootstrap aggregating}.}
procedures~\citep{Breiman:1996:BP:231986.231989}, the weak predictors
are built in parallel and then combined using voting
(classification)~\citep{Leung:2003:ECV:956750.956825} or averaging
(regression)~\citep{Leite:2007:IPB:1782254.1782263}. On the contrary,
\textit{boosting} algorithms~\citep{Schapire:1990:SWL:83637.83645} do
it sequentially, which allows to adapt the distribution of the
training data base from the performance of the previous weak
predictors. This gives rise to \textit{arcing}\footnote{For
  \textit{adaptive resampling and combining}.}
strategies~\citep{Freund+Schapire:1996}, where increasing weight is
placed on the more frequently misclassified observations. Since these
are the troublesome points, focusing on them may do better than the
neutral bagging approach~\citep{Bauer:1999:ECV:599591.599607},
justifying its
popularity~\citep{Garcia-Pedrajas:2014:BIS:2657522.2658088}. Another
common strategy for increasing stability, specially in {\sc al}, is
the use of thresholds providing a flexibility with regard to variance,
but without warranty of any type. The stability of predictions is
studied on a set of examples, called the stop set, that do not have to
be labeled~\citep{Bloodgood:2009:MSA:1596374.1596384}. At best, the
dynamic update of thresholds is outlined to increase soundness across
changing data sets~\citep{Zhu:2012:UAL:2093153.2093154}.

\subsection{An overview for the {\sc nlp} domain}

The use of learning curves for anticipating the performance of {\sc
  nlp} tools has been the subject of ongoing research during the last
years, mainly in the sphere of \textit{machine translation} ({\sc
  mt}). So, they have been employed for assessing the quality of {\sc
  mt}
systems~\citep{Bertoldi:2012:ELC:2393015.2393076,Turchi:2008:LPM:1626394.1626399},
for optimizing parameter
setting~\citep{Koehn:2003:SPT:1073445.1073462} and for estimating how
many training data are required to achieve a certain degree of
translation accuracy~\citep{Kolachina:2012:PLC:2390524.2390528},
although no formal stopping criteria are described. They were also
used to evaluate the impact of a concrete set of distortion factors on
the performance of a concrete operational
model~\citep{Birch:2008:PSM:1613715.1613809}, albeit with meagre
results. Together, all these works recall the essence of our problem
formulation and serve also as examples of how correctness and
robustness are treated in the prediction of learning curves. So, the
lack of well-founded mathematical models results in proposals whose
sole support comes from a test battery, which in the best case offers a
partial vision of the problem. Therefore, they are a long way from
achieving a solution verifying our requirements on correctness.

In the case of {\sc al} techniques, their popularity is growing in
{\sc pos} tagging
tasks~\citep{Dagan95committee-basedsampling,Haertel:2008:ACS:1557690.1557708,Ringger:2007:ALP:1642059.1642075}
and closely related ones such as named entity
recognition~\citep{Laws:2008:SCA:1599081.1599140,Shen:2004:MAL:1218955.1219030,Tomanek07anapproach}
or word sense
disambiguation~\citep{ChanN07,Chen:2006:ESB:1220835.1220851,Zhu07activelearning},
with the purpose of reducing the annotation effort. The same is true
for information
extraction~\citep{Culotta:2005:RLE:1619410.1619452,Thompson:1999:ALN:645528.657614},
parsing~\citep{Becker:2005:TMA:1642293.1642452,Tang:2002:ALS:1073083.1073105}
or text
classification~\citep{Lewis:1994:SAT:188490.188495,Liere97activelearning,McCallum:1998:EEP:645527.757765,Tong:2002:SVM:944790.944793}
applications. In none of these cases, however, these works have
contributed to the treatment of correctness and robustness beyond what
we have seen so far.

\subsection{Our contribution}

In order to confer reliability on the prediction of accuracy in tools
resulting from {\sc ml} processes, we introduce an iterative
functional architecture as an alternative to the classic statistical
techniques. This is defined on a sequence of approximations for the
partial learning curves, which are calculated from an increasing set
of observations. On the basis of a set of working hypotheses widely
recognised in both learning curves and training data bases, the
correctness of the method is theoretically proved. We can then, in
contrast to earlier works, define a proximity criterion to stop the
learning once a degree of accuracy fixed by the user is
reached. Regarding robustness against variations in the working
hypotheses, we propose a anchoring mechanism which formally limits
their impact, and is fully compatible with the basic algorithm.

\section{The formal framework}
\label{section-formal-framework}

The aim now is to describe our abstract model on a mathematical basis
that would enable us to prove its correctness. We choose a function,
which must be continuous so that it provides sustainability to estimate
in advance the learning curve for accuracy. Since the approximations
of that function can be modelled from partial learning curves, it
seems natural to raise its calculation as the convergence of the
sequence of such approximations while the training process advances.

\subsection{The mathematical support}

We first recall some notions~\citep{Apostol} on the theory of
sequences in the real metric space $(\mathbb{R}, \abs{\;})$, where
$\abs{\;}$ denotes the Euclidean distance defined as the absolute
difference, and $\mathbb{R}$ is the set of real numbers. For the sake
of simplicity, we assume familiarity with the concepts of continuity
and derivability of a real function. We denote the set of natural
numbers by $\mathbb{N}$, and we assume that $0 \not\in \mathbb{N}$.

\begin{df}
\label{def-convergence-sequence}
Let $\{x_i\}_{i \in \mathbb{N}}$ be a sequence in 
$(\mathbb{R}, \abs{\;})$, we say that it is a {\em sequence convergent
  to} $x_\infty \in \mathbb{R}$ iff
\begin{equation}
\forall \varepsilon > 0, \exists n \in \mathbb{N}, 
   \forall i \geq n \Rightarrow \abs{x_i - x_\infty} < \varepsilon
\end{equation}
\noindent where $x_\infty$ is called the {\em limit of} $\{x_i\}_{i \in
  \mathbb{N}}$, using the notation $\lim \limits_{i \rightarrow
  \infty} x_i = x_\infty$.
\end{df}

A sequence converges when we can situate, from a given moment, all
its elements as close to the limit as we want to. It may be proved
that any monotonic increasing (resp. decreasing) and upper
(resp. lower) bounded sequence converges to its supremum
(resp. infimum). Since we want to study the convergence of a
collection of curves, we need to extend the concept to sequences of
real functions.

\begin{df}
\label{def-convergence-functional-sequence}
Let $\Delta := \{f:E \subseteq \mathbb{R} \rightarrow \mathbb{R}\}$
and let $\{f_i\}_{i \in \mathbb{N}}, \; f_i \in \Delta$ be a sequence
of functions, we say that it is a {\em sequence pointwise
  convergent to} $f_\infty \in \Delta$ iff
\begin{equation}
\forall x \in E, \varepsilon >0, \exists n_x \in \mathbb{N}, \forall i
\geq n_x \Rightarrow \abs{f_i(x) - f_\infty(x)} < \varepsilon
\end{equation}
where $f_\infty$ is called the {\em pointwise limit of} $\{f_i\}_{i \in
  \mathbb{N}}$, using the notation ${\lim \limits_{i \rightarrow
  \infty}}^p f_i = f_\infty$.
\end{df}

A sequence of functions is pointwise convergent if the sequence of
their values on each instant converges, which implies that we can
calculate the limit point-to-point, although the speed of convergence
may be different in each case. This poses a problem when we want to
use the limit function for prediction purposes because the results
obtained could vary greatly even over points close to the
observations. Namely, in order to provide reliability to our
estimates, we need a criterion making it possible to consider a common
convergence threshold for the whole functional domain considered.

\begin{df}
\label{def-uniform-convergence-functional-sequence}
Let $\Delta := \{f:E \subseteq \mathbb{R} \rightarrow \mathbb{R}\}$
and let $\{f_i\}_{i \in \mathbb{N}}, \; f_i \in \Delta$ be a sequence
of functions, we say that it is a {\em sequence uniformly convergent
  to} $f_\infty \in \Delta$ iff
\begin{equation}
\forall \varepsilon >0, \exists n \in \mathbb{N}, \forall
i \geq n \Rightarrow \abs{f_i(x) - f_\infty(x)} < \varepsilon, \forall x
\in E
\end{equation}
\noindent where $f_\infty$ is called the {\em uniform limit of} $\{f_i\}_{i \in
  \mathbb{N}}$, using the notation ${\lim \limits_{i \rightarrow
  \infty}}^u f_i = f_\infty$.
\end{df}

This identifies functional sequences for which all points converge at
the same pace to the limit, and allows its continuity to be inferred
when the curves in the sequence are continuous. Such a property is also
useful for prediction purposes since continuity is a guarantee of
regularity, matching small variations in the input to small ones in
the output.

\subsection{The working hypotheses}

Since the approximation of partial learning curves are the basis for
our proposal, we need to gather as much information as possible about
their nature in order to identify the functions better adapted to that
requirement. Thus, the point of departure for generating a learning
curve is a sequence of observations calculated from a series of cases
incrementally taken from a training data base over the time, which
should meet certain conditions in order to make the prediction task
possible. Basically, we assume this data base to be independently and
identically
distributed~\citep{Schutze:2006:PTP:1183614.1183709,KATRIN08.335}, in
order to obtain a predictable progression of the estimation trace for
accuracy over a virtually infinite interval. This does not imply a
loss of generality since we can re-order the training data
base~\citep{Clark2010} and generate as many observations as we want.

\begin{figure}[htbp]
\begin{center}
\begin{tabular}{cc}
\hspace*{-.7cm}
\epsfxsize=.52\linewidth
\epsffile{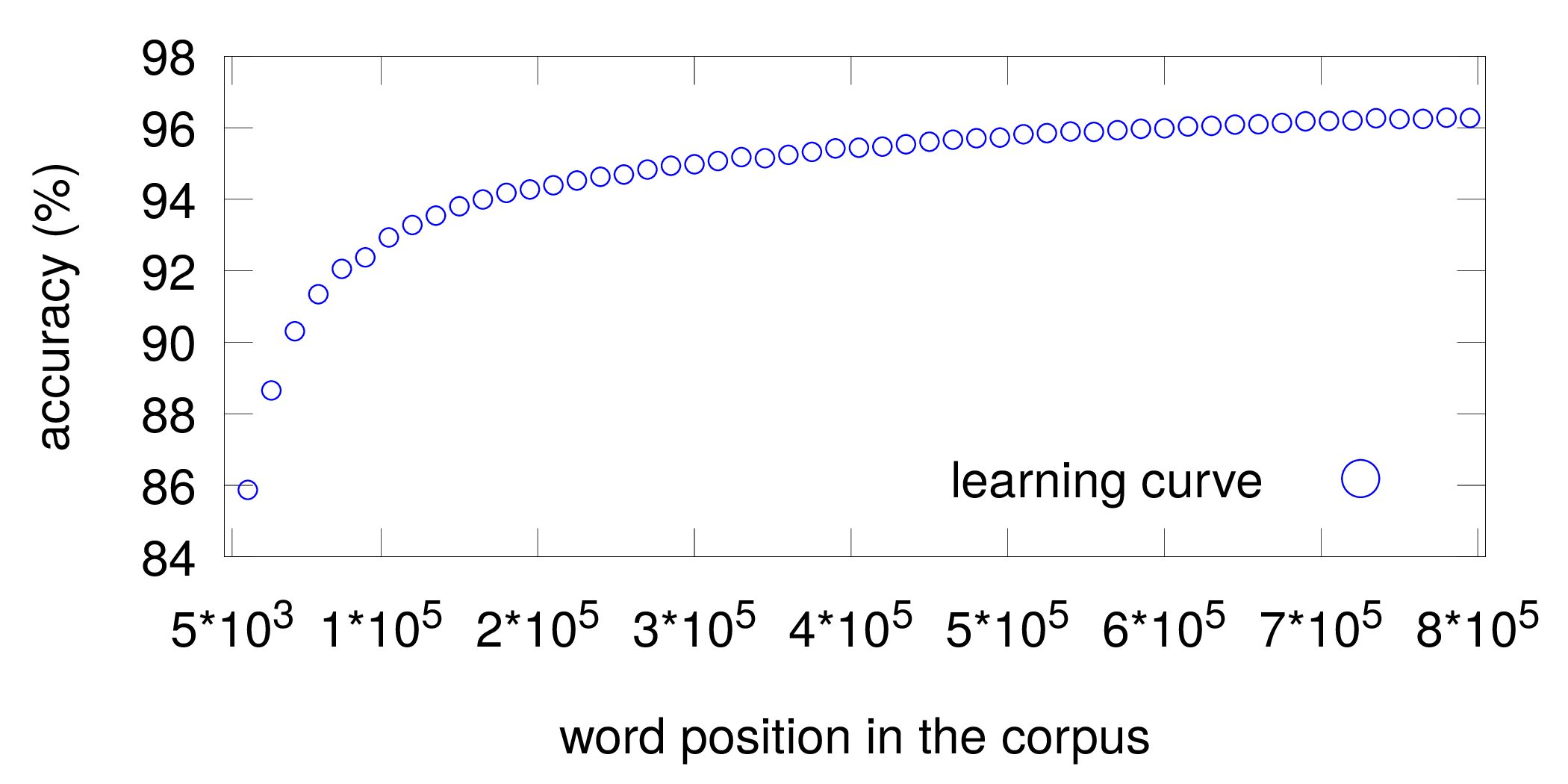} 
&
\hspace*{-.7cm}
\epsfxsize=.52\linewidth
\epsffile{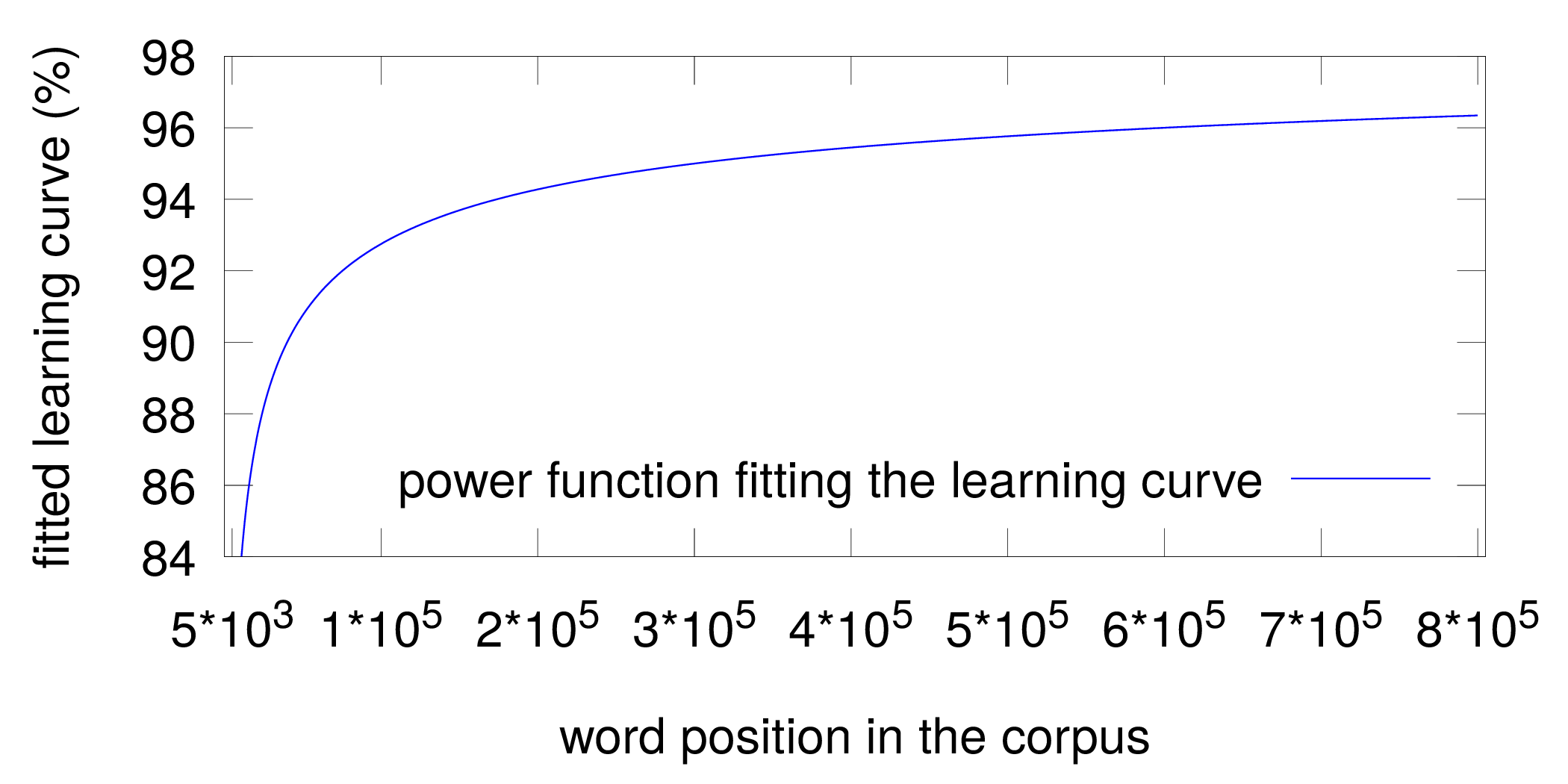}
\end{tabular}
\caption{Learning curve for the training process of fn{\sc
    tbl} on {\sc f}rown corpus, and an accuracy pattern fitting it.}
\label{fig-accuracy-fnTBL-Frown-5000-800000}
\end{center}
\end{figure}

We then accept that a learning curve is a positive definite and
strictly increasing function on $\mathbb{N}$, where natural numbers
represent the sample size, upper bounded by 100. We also assume that
the speed of increase is higher in its first stretch, where the
learning is faster, decreasing as the training process advances and
giving the curve a concave form with a horizontal
asymptote~\citep{FreyFisher99,Last:2009:IDM:1557019.1557076,Meek:2002:LSM:944790.944798}. One
might then argue that such requirements cannot be completely
guaranteed in practice, as it can be observed in the left-most diagram
of Figure~\ref{fig-accuracy-fnTBL-Frown-5000-800000}, which shows the
learning curve for the training process of the {\it fast
  transformation-based learning} (fn{\sc tbl}) tagger~\citep{Ngai2001}
on the \textit{Freiburg-Brown} ({\sc f}rown) corpus of American
English~\citep{Mair2007}. We consider here examples indicated by the
position of a word from the beginning of the text\footnote{All the
  word counts in this paper include punctuation marks.}, thus
delimiting the section of this used to generated them and evidencing
the existence of small irregularities in both concavity and increase.
It is then necessary to take into account that the idealization is
inherent to the scientific method, the objective of which is to lay
the foundations for the correctness, leaving for a subsequent stage
the question of robustness.

\subsection{The notational support}

Having identified the context of the problem, we need to formally
capture the data structures we are going to work with, such as the
collection of instances on which the accuracy measurements are
sequentially applied to generate the observations that will serve as 
starting point for the prediction task.

\begin{df}
\label{def-learning-scheme}
Let ${\mathcal D}$ be a training data base, $\mathcal K \subsetneq
\mathcal D$ a non empty set of initial items and $\sigma:\mathbb{N}
\rightarrow \mathbb{N}$ a function. We define a {\em learning scheme}
for $\mathcal D$ with {\em kernel} $\mathcal K$ and {\em step
  function} $\sigma$, as a triple
$\mathcal{D}^{\mathcal{K}}_{\sigma}=[\mathcal{K},\sigma,\{\mathcal
  D_i\}_{i \in \mathbb{N}}]$ with $\{\mathcal D_i\}_{i \in
  \mathbb{N}}$ an incremental cover of $\mathcal{D}$ verifying:
\begin{equation}
{\mathcal D}_1 := {\mathcal K} \mbox{ and }
{\mathcal D}_i := {\mathcal
  D}_{i-1} \cup {\mathcal I}_{i}, \; \mathcal I_i
\subset {\mathcal D} \setminus {\mathcal
  D}_{i-1}, \; \absd{{\mathcal I}_{i}}=\sigma(i), \; \forall i \geq 2
\end{equation}
\noindent where $\absd{{\mathcal I}_{i}}$ is the cardinality of
${\mathcal I}_{i}$. We refer to $\mathcal{D}_i$ as the {\em individual
  of level} $i$ {\em for} $\mathcal{D}^{\mathcal{K}}_{\sigma}$.
\end{df}

A learning scheme relates a level $i \in \mathbb{N}$ with the position
$x_i := \absd{\mathcal D_i}$ in the training data base of its
corresponding case, thereby determining the sequence of observations
$\{[x_i, {\mathcal A}_{\dinfty{}}[{\mathcal D}^{\mathcal
      {K}}_{\sigma}](x_i)], \; x_i := \absd{\mathcal D_i} \}_{i \in
  \mathbb{N}}$, where ${\mathcal A}_{\dinfty{}}[{\mathcal D}^{\mathcal
    {K}}_{\sigma}](x_i)$ is the accuracy achieved on such instance by
the system we are studying. Namely ${\mathcal A}_{\dinfty{}}[{\mathcal
    D}^{\mathcal {K}}_{\sigma}]$ is the learning curve associated to
the scheme ${\mathcal D}^{\mathcal {K}}_{\sigma}$ that we want to
approximate, and the kernel $\mathcal K$ delimits a portion of
${\mathcal D}$ we believe to be enough to initiate consistent
evaluations. We therefore need a functional pattern serving as model
for fitting these curves. This leads us to consider candidates
verifying our working hypotheses, but also representing real
C-infinity functions in order to provide reliability to our estimates
through their structural smoothness.

\begin{df}
\label{def-accuracy-pattern-fitting}
Let $C^\infty_{(0,\infty)}$ be the real C-infinity functions in
$\mathbb{R}^{+}$, we say that $\pi: \mathbb{R}^{{+}^{n}} \rightarrow
C^\infty_{(0,\infty)}$ is an {\em accuracy pattern} iff $\pi(a_1,
\dots, a_n)$ is positive definite, concave and strictly increasing.
\end{df}

An example of accuracy pattern is the \textit{power family} of curves
$\pi(a,b,c)(x) :=-a * x^{-b}+c$, hereinafter used as running
example. They have \(\lim \limits_{x \rightarrow \infty} \pi(a,b,c)(x)
= c\) as horizontal asymptote and verify:
\begin{equation}
\pi(a,b,c)'(x)=a * b * x^{-(b+1)} > 0 \hspace*{1.5cm} 
\pi(a,b,c)''(x)=-a * b
* (b+1) * x^{-(b+2)} < 0 
\end{equation}
\noindent which guarantees increase and concavity in $\mathbb{R}^{+}$,
respectively. This is illustrated in the right-most curve of
Figure~\ref{fig-accuracy-fnTBL-Frown-5000-800000}, whose goal is fitting
the learning curve represented in the left-hand side.  Here, the
values $a=542.5451$, $b=0.3838$ and $c=99.2876$ are provided by the
\textit{trust region method}~\citep{Branch1999}, a regression
technique minimizing the summed square of \textit{residuals}, namely
the differences between the observed values and the fitted ones.

\begin{figure}[htbp]
\begin{center}
\begin{tabular}{cc}
\hspace*{-.7cm}
\epsfxsize=.52\linewidth
\epsffile{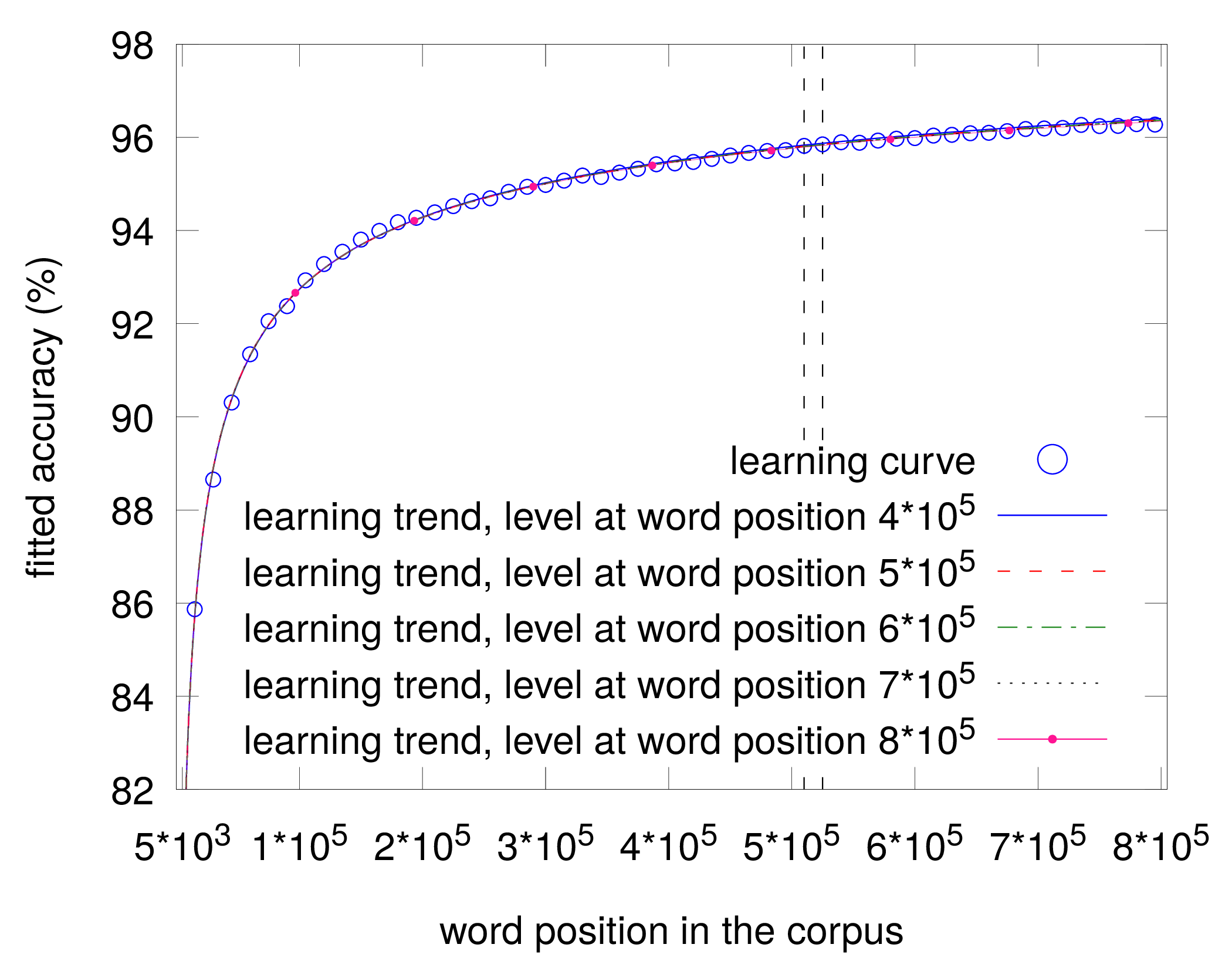}
& 
\hspace*{-.7cm}
\epsfxsize=.52\linewidth
\epsffile{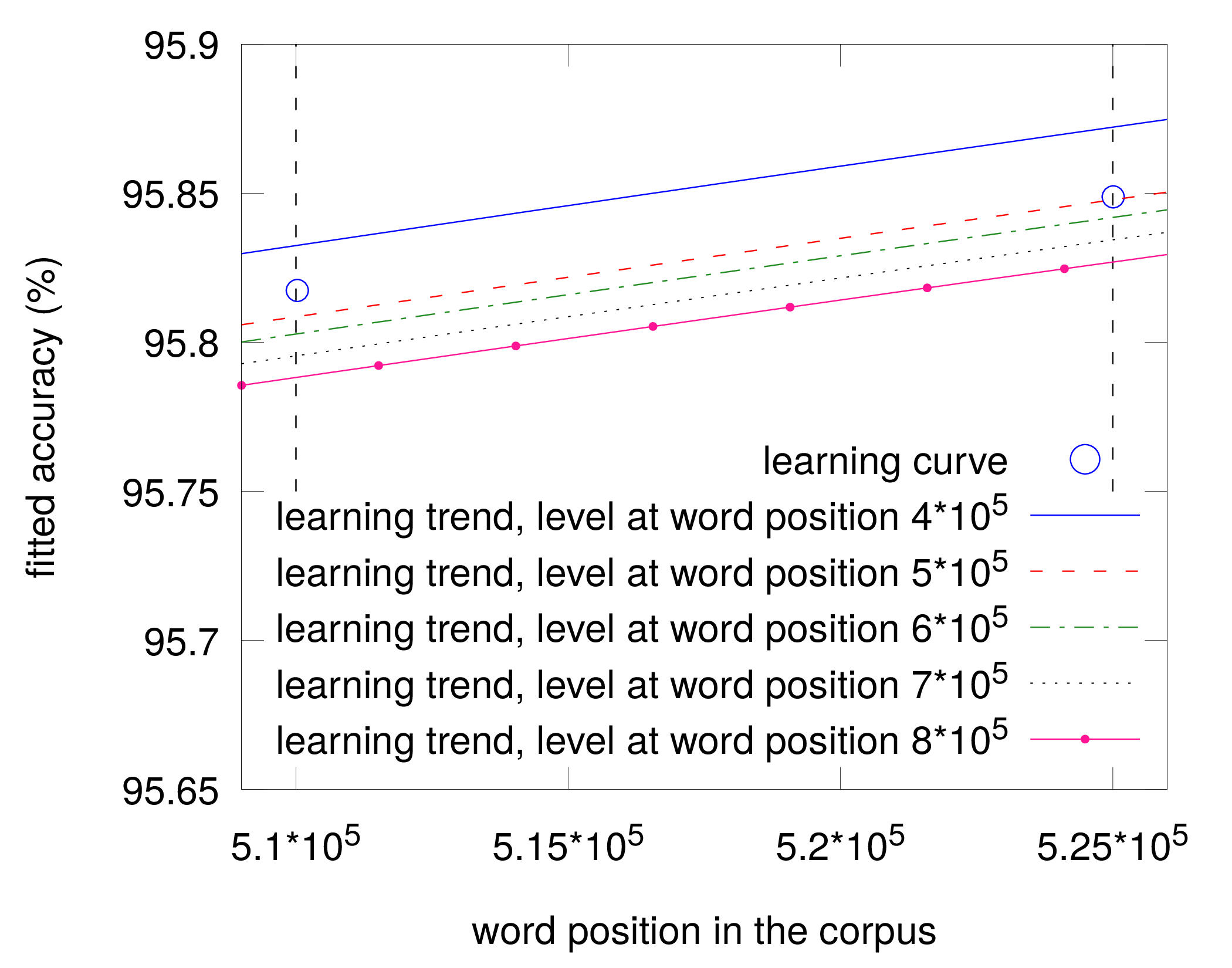}
\end{tabular}
\caption{Learning trace for the training process of fn{\sc tbl} on
  {\sc f}rown, with details in zoom.}
\label{fig-trace-and-level-sequence-nlls-fnTBL-Frown-5000-800000}
\end{center}
\end{figure}

\begin{df}
\label{def-trace}
Let ${\mathcal A}_{\dinfty{}}[\mathcal{D}^{\mathcal{K}}_{\sigma}]$ be
a learning curve, $\pi$ an accuracy pattern and $\ell \in \mathbb{N},
\; \ell \geq 3$ an item position in the training data base
$\mathcal{D}$. We define the {\em learning trend of level} $\ell$ {\em
  for} ${\mathcal A}_{\dinfty{}}[{\mathcal D}^{\mathcal{K}}_{\sigma}]$
   {\em using} $\pi$, as a curve ${\mathcal A}_{\ell}^\pi[{\mathcal
       D}^{\mathcal{K}}_{\sigma}] \in \pi$, fitting the observations
   $\{[x_i, {\mathcal A}_{\dinfty{}}[{\mathcal D}^{\mathcal
         {K}}_{\sigma}](x_i)], \; x_i := \absd{\mathcal D_i}
   \}_{i=1}^{\ell}$. A sequence of learning trends ${\mathcal
     A}^\pi[{\mathcal D}^{\mathcal {K}}_{\sigma}] :=\{{\mathcal
     A}_{\ell}^\pi[{\mathcal D}^{\mathcal{K}}_{\sigma}]\}_{\ell \in
     \mathbb{N}}$ is called a {\em learning trace}. We denote by
   $\rho_\ell(i) := [{\mathcal A}_{\dinfty{}}[{\mathcal
         D}^{\mathcal{K}}_{\sigma}] - {\mathcal
       A}_{\ell}^\pi[{\mathcal
         D}^{\mathcal{K}}_{\sigma}]](\absd{{\mathcal D}_i})$ the {\em
     residual of} ${\mathcal A}_{\ell}^\pi[{\mathcal
       D}^{\mathcal{K}}_{\sigma}]$ {\em at the level} $i \in
   \mathbb{N}$. We refer to $\{\alpha_\ell\}_{\ell \in \mathbb{N}}$ as
   the {\em asymptotic backbone} of ${\mathcal A}^\pi[{\mathcal
       D}^{\mathcal {K}}_{\sigma}]$, where $y = \alpha_\ell$ is the
   asymptote of ${\mathcal A}_\ell^\pi[{\mathcal D}^{\mathcal
       {K}}_{\sigma}]$.
\end{df}

The learning trends also fit, namely approximate, the partial learning
curves beyond their own level. However, this is not enough to provide
a credible prediction for accuracy because to do so requires
information on their evolution during the training process. Learning
traces solve this question, allowing the extraction of a sequence of
fitted values for a case, representing the evolution of the estimation
at that instant. Continuing with the example for the tagger fn{\sc
  tbl} and the corpus {\sc f}rown,
Figure~\ref{fig-trace-and-level-sequence-nlls-fnTBL-Frown-5000-800000}
shows a portion of the learning trace associated to a kernel and
constant step function $5*10^3$, the levels of which are indicated by
the corresponding word position in the corpus, including also a more
detailed view. Finally, we can only consider learning trends from the
level $\ell =3$ because we need at least three observations to
generate a curve.

\section{The abstract model}
\label{section-abstract-model}

We lay the theoretical foundations of our proposal to later interpret
them from a operational point of view. The first step is proving its
correctness, which also allows us to discuss it against the
deviations introduced by real observations.

\subsection{Correctness}

Since the intention is for us to reliably approximate the learning
curve ${\mathcal A}_{\dinfty{}}[{\mathcal D}^{\mathcal{K}}_{\sigma}]$
by means of the limit function of the sequence ${\mathcal
  A}^\pi[{\mathcal D}^{\mathcal {K}}_{\sigma}] :=\{{\mathcal
  A}_{i}^\pi[{\mathcal D}^{\mathcal{K}}_{\sigma}]\}_{i \in
  \mathbb{N}}$ of learning trends incrementally built from the
observations, we start by studying its uniform convergence. As notational
facility, we extend the natural order in $\mathbb{N}$ in such a way
that $\ldinfty{} > \infty > i > 0, \; \forall i \in \mathbb{N}$.

\begin{thm}
\label{th-uniform-convergence-trace} 
Let ${\mathcal A}^\pi[{\mathcal D}^{\mathcal {K}}_{\sigma}]$ be a
learning trace, with or without anchors. Then its asymptotic backbone
is monotonic and ${\mathcal A}_{\infty}^\pi[{\mathcal
    D}^{\mathcal{K}}_{\sigma}] := {\lim \limits_{i \rightarrow
    \infty}}^u {\mathcal A}_{i}^\pi[{\mathcal
    D}^{\mathcal{K}}_{\sigma}]$ exists, is positive definite,
increasing, continuous and upper bounded by 100 in $(0, \infty)$.\end{thm}

\begin{pf}
Having fixed a level $i \in \mathbb{N}$, the fitting
algorithm minimizes a weighting function on the set of residuals
$\{\rho_i(j)\}_{j \leq i}$ in order to generate a learning trend
${\mathcal A}_{i}^\pi[{\mathcal D}^{\mathcal{K}}_{\sigma}]$, such that
$\sum_{j \leq i} \rho_i(j) = 0$. Consequently, the latter intersects
necessarily the learning curve ${\mathcal A}_{\dinfty{}}[{\mathcal
    D}^{\mathcal {K}}_{\sigma}]$. Since the concavity of both does not
vary in $(0, \infty)$, they cross at one ($q_{\dinfty{}}^i$) or two
($p_{\dinfty{}}^i$ and $q_{\dinfty{}}^i$) points in that interval. As
shown in the left-hand-side of
Figs.~\ref{fig-residual-subintervals-decreasing}
and~\ref{fig-residual-subintervals-increasing}, this delimits two
($B_{\dinfty{}}^i$ and $C_{\dinfty{}}^i$) or three ($A_{\dinfty{}}^i,
\; B_{\dinfty{}}^i$ and $C_{\dinfty{}}^i$) sub-intervals. So,
$A_{\dinfty{}}^i := (0, p_{\dinfty{},x}^{i}]$, $B_{\dinfty{}}^i :=
  (p_{\dinfty{},x}^i,q_{\dinfty{},x}^i]$ and $C_{\dinfty{}}^i :=
    (q_{\dinfty{},x}^i,\infty)$, where $p_{\dinfty{}}^i :=
    (p_{\dinfty{},x}^i,p_{\dinfty{},y}^i)$, $q_{\dinfty{}}^i :=
    (q_{\dinfty{},x}^i,q_{\dinfty{},y}^i)$, and the sign of residuals
    in $B_{\dinfty{}}^i$ is different from that of $A_{\dinfty{}}^i$
    and $C_{\dinfty{}}^i$. Thus, the trend in $C_{\dinfty{}}^i$ is
    either above or below the observation and $0 < p_{\dinfty{},x}^i <
    q_{\dinfty{},x}^i < \absd{\mathcal D_i}$. \\

\begin{figure}[htbp]
\begin{center}
\epsfxsize=.95\linewidth
\epsffile{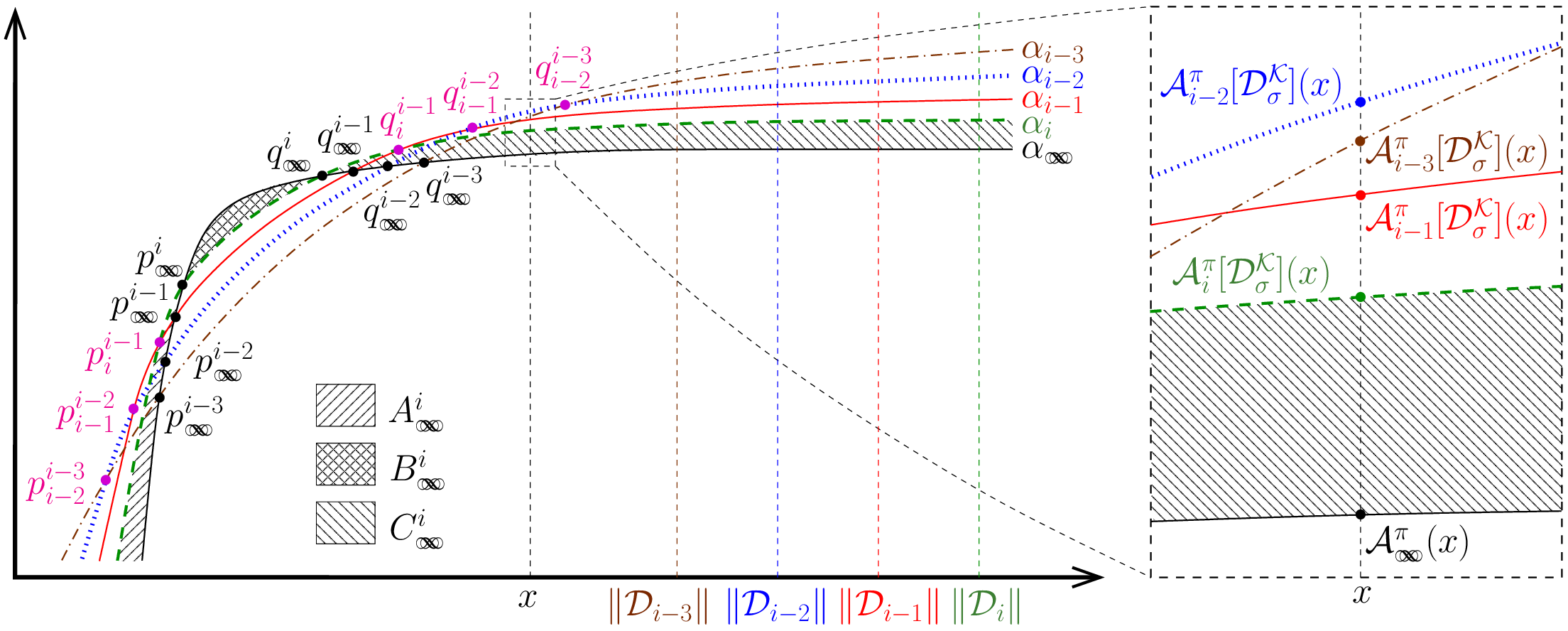} 
\caption{An example of construction for a decreasing learning trace,
  with details in zoom.}
\label{fig-residual-subintervals-decreasing}
\end{center}
\end{figure}

\noindent As learning curves and trends are strictly monotonic
(increasing), the impact of the observations in the fitting process always
applies in one direction from the first position $q_{\dinfty{},x}^1$,
increasing as the levels ascend. Accordingly, the asymptotic backbone
$\alpha := \{\alpha_i\}_{i \in \mathbb{N}}$ is also monotonic. More
specifically, as illustrated in
Figure~\ref{fig-residual-subintervals-decreasing}
(resp. Figure~\ref{fig-residual-subintervals-increasing}), $\alpha$ is
lower (resp. upper) bounded by the asymptotic value of the learning
curve $\alpha_{\dinfty{}}$ if the sequence is decreasing
(resp. increasing). Given that this bound is, in fact, an infimum
(resp. a supremum), we can thus conclude that $\alpha$ converges
monotonically to $\alpha_{\dinfty{}}$. \\

\begin{figure}[htbp]
\begin{center}
\epsfxsize=.95\linewidth
\epsffile{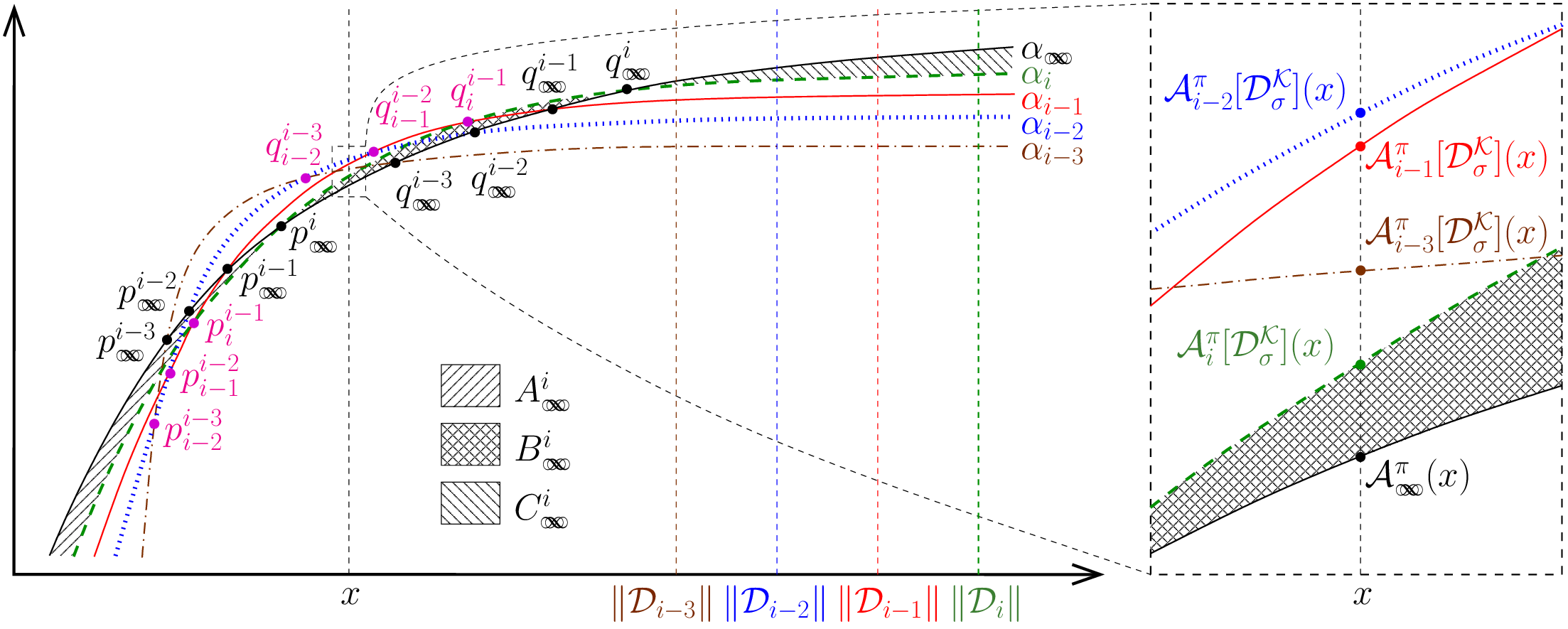} 
\caption{An example of construction for an increasing learning trace,
  with details in zoom.}
\label{fig-residual-subintervals-increasing}
\end{center}
\end{figure}

\noindent On the other hand, ${\mathcal A}_{i}^\pi[{\mathcal
    D}^{\mathcal{K}}_{\sigma}]$ is itself a learning trend
approximating ${\mathcal A}_{j}^\pi[{\mathcal
    D}^{\mathcal{K}}_{\sigma}], \ i < j$. Following the same reasoning
used before with regard to the learning curve, they cross at one
($q_j^{i}$) or two ($p_j^{i}$ and $q_j^{i}$) points in $(0,
\infty)$. This delimits two ($B_j^i$ and $C_j^i$) or three ($A_j^i, \;
B_j^i$ and $C_j^i$) sub-intervals. So, $A_j^i := (0, p_{j,x}^i]$,
  $B_j^i := (p_{j,x}^i,q_{j,x}^i]$ and $C_j^i := (q_{j,x}^i,\infty)$,
    where $p_j^i := (p_{j,x}^i,p_{j,y}^i)$, $q_j^i :=
    (q_{j,x}^i,q_{j,y}^i)$, and $0 < p_{j,x}^i < q_{j,x}^i <
    \absd{\mathcal D_j}$. \\

\noindent Let $\{(q_{i,x}^{i-1},q_{i,y}^{i-1})\}_{i \geq 4}$
(resp. $\{(p_{i,x}^{i-1},p_{i,y}^{i-1})\}_{i \geq 4}$) then be the
sequence of the last (resp. first, if they exist) points of
intersection between a trend ${\mathcal A}_{i}^\pi[{\mathcal
    D}^{\mathcal{K}}_{\sigma}]$ and the previous one ${\mathcal
  A}_{i-1}^\pi[{\mathcal D}^{\mathcal{K}}_{\sigma}]$. Given that
$\alpha$ converges monotonically, $\{q_{i,y}^{i-1}\}_{i \geq 4}$ also
does so, as shown in Figs.~\ref{fig-residual-subintervals-decreasing}
and~\ref{fig-residual-subintervals-increasing} regardless of the type
of monotony exhibited by the asymptotic backbone. For the same reason,
if it exists, $\{p_{i,y}^{i-1}\}_{i \geq 4}$ converges to a point less
than or equal to the one for $\{q_{i,y}^{i-1}\}_{i \geq 4}$. Since
$\{p_{i,y}^{i-1}\}_{i \geq 4}$ and $\{q_{i,y}^{i-1}\}_{i \geq 4}$ are
monotonic, $\{p_{i,x}^{i-1}\}_{i \geq 4}$ and $\{q_{i,x}^{i-1}\}_{i
  \geq 4}$ are also because the trends are strictly
increasing. Moreover, $0 < p_{i,x}^{i-1} < q_{j,x}^{j-1} <
\absd{\mathcal D_j}, \; \forall i, j \in \mathbb{N}, \; i \neq j$
because otherwise $\{q_{i,y}^{i-1}\}_{i \geq 4}$ would not be
monotonic. This implies that $\{p_{i,x}^{i-1}\}_{i \geq 4}$ is also
lower (resp. upper) bounded by $0$ (resp. $q_{2,x}^1$) and, therefore,
$\lim \limits_{i \rightarrow \infty} p_{i,x}^{i-1}$ exists. \\

\noindent Thus, as illustrated in the right-hand-side of
Figs.~\ref{fig-residual-subintervals-decreasing}
and~\ref{fig-residual-subintervals-increasing}, in $(0, \lim
\limits_{j \rightarrow \infty} p_{j,x}^{j-1})$ the sequences
$\{{\mathcal A}_{i}^\pi[{\mathcal D}^{\mathcal{K}}_{\sigma}](x)\}_{i
  \in \mathbb{N}}$ are monotonic from $i$ such that $x <
p_{i,x}^{i-1}$; in $[\lim \limits_{j \rightarrow \infty}
  p_{j,x}^{j-1}, \lim \limits_{j \rightarrow \infty} q_{j,x}^{j-1})$
  from $i$ such that $x < q_{i,x}^{i-1}$; and in $[\lim \limits_{j
      \rightarrow \infty} q_{j,x}^{j-1}, \infty)$ if
    $\{q_{j,x}^{j-1}\}_{j \geq 4}$ converges. Since they are also
    bounded because the learning trends are particular configurations
    of $\pi$, ${\mathcal A}_{\infty}^\pi[{\mathcal
        D}^{\mathcal{K}}_{\sigma}](x) := \lim \limits_{i \rightarrow
      \infty} {\mathcal A}_{i}^\pi[{\mathcal
        D}^{\mathcal{K}}_{\sigma}](x)$ is a well defined, positive
    definite and increasing function. Trivially, ${\mathcal
      A}_{\infty}^\pi[{\mathcal D}^{\mathcal{K}}_{\sigma}] = {\lim
      \limits_{i \rightarrow \infty}}^p {\mathcal A}_{i}^\pi[{\mathcal
        D}^{\mathcal{K}}_{\sigma}]$ in $(0, \infty)$. Given that
    $\{\alpha_i\}_{i \in \mathbb{N}}$ converges to
    $\alpha_{\dinfty{}}$ and the observations are upper bounded by
    100, we have that
\begin{equation}
\exists n_\alpha \in \mathbb{N}, \; \forall i \geq n_\alpha
\Rightarrow \abs{{\mathcal A}_{i}^\pi[{\mathcal
    D}^{\mathcal{K}}_{\sigma}](x) - \alpha_{\dinfty{}}} \leq \varepsilon = \abs{100 -
  \alpha_{\dinfty{}}}, \; \forall x \in (0, \infty)
\end{equation}
\noindent or, in other words, 
\begin{equation}
\exists n_\alpha \in \mathbb{N}, \; \forall i \geq n_\alpha \Rightarrow
\abs{{\mathcal A}_{i}^\pi[{\mathcal D}^{\mathcal{K}}_{\sigma}](x)} \leq
\abs{\alpha_i} \leq 100, \; \forall x \in (0, \infty)
\end{equation}
\noindent which implies that ${\mathcal A}_{\infty}^\pi[{\mathcal
    D}{\mathcal{K}}_{\sigma}]$ is also upper bounded by 100. \\

\noindent To prove now that the convergence is uniform, it is
sufficient to take into account that
\begin{equation}
\lim \limits_{i \rightarrow \infty} (\sup_{x \in (0, \infty)} \abs{{\mathcal
      A}_{i}^\pi[{\mathcal D}^{\mathcal{K}}_{\sigma}](x) - {\mathcal
      A}_{\infty}^\pi[{\mathcal D}^{\mathcal{K}}_{\sigma}](x)}) = 0
\end{equation}
because, by construction, no vertical asymptotic behavior is
observable in the trace. As all the trends are continuous, then so is
their uniform limit ${\mathcal A}_{\infty}^\pi[{\mathcal
    D}^{\mathcal{K}}_{\sigma}]$. $\blacksquare$
\end{pf}

All this provides us with an abstract model to estimate the learning
curve ${\mathcal A}_{\dinfty{}}[{\mathcal D}^{\mathcal{K}}_{\sigma}]$
over the training data base by iteratively approximating the function
${\mathcal A}_{\infty}^\pi[{\mathcal D}^{\mathcal{K}}_{\sigma}]$,
while there is a need for measuring the convergence (resp. error)
threshold at each stage of the process, in order to give it a
practical sense. Namely, after fixing a level $i \in \mathbb{N}$, we
have to calculate an upper bound for the distance between ${\mathcal
  A}_{j}^\pi[{\mathcal D}^{\mathcal{K}}_{\sigma}]$ and ${\mathcal
  A}_{\infty}^\pi[{\mathcal D}^{\mathcal{K}}_{\sigma}]$
(resp. ${\mathcal A}_{\dinfty{}}[{\mathcal
    D}^{\mathcal{K}}_{\sigma}]$) in the interval $[\absd{{\mathcal
      D}_j}, \infty), \; \forall j \geq i$. In other words, we need to
  define a proximity criterion. With a view to simplifying the
  wording, we also denote ${\mathcal A}_{\dinfty{}}[{\mathcal
      D}^{\mathcal{K}}_{\sigma}]$ by ${\mathcal
    A}_{\dinfty{}}^\pi[{\mathcal D}^{\mathcal{K}}_{\sigma}]$ whatever
  the accuracy pattern $\pi$.

\begin{thm}
\label{th-correctness-trace} 
{\em (Correctness)} Let ${\mathcal A}^\pi[{\mathcal D}^{\mathcal
    {K}}_{\sigma}]$ be a learning trace, with $y = \alpha_i$ the
asymptote for ${\mathcal A}_i^\pi[{\mathcal D}^{\mathcal
    {K}}_{\sigma}], \; \forall i \in \mathbb{N} \cup \{\infty,
\dinfty{}\}$ and $(q_{i,x}^{i-1},q_{i,y}^{i-1})$ the last point in
${\mathcal A}_i^\pi[{\mathcal D}^{\mathcal {K}}_{\sigma}] \cap
       {\mathcal A}_{i-1}^\pi[{\mathcal D}^{\mathcal {K}}_{\sigma}],
       \; \forall i \geq 4$. We then have that
\begin{equation}
\label{equation-correctness-trace-decreasing}
\abs{[{\mathcal A}_k^\pi[{\mathcal D}^{\mathcal {K}}_{\sigma}] -
    {\mathcal A}_j^\pi[{\mathcal D}^{\mathcal {K}}_{\sigma}]](x)} \leq
\varepsilon_i := \abs{q_{i,y}^{i-1} - \alpha_i}, \; \forall k,j \geq i
\geq 4, \; \forall x \in [q_{i,x}^{i-1}, \infty)
\end{equation}
\begin{equation}
\label{equation-correctness-trace-increasing}
\mbox{\em (resp. } \abs{[{\mathcal A}_k^\pi[{\mathcal D}^{\mathcal
        {K}}_{\sigma}] - {\mathcal A}_j^\pi[{\mathcal D}^{\mathcal
        {K}}_{\sigma}]](x)} \leq \varepsilon_i := 
\abs{q_{\dinfty{},y}^{i} - \alpha_{\dinfty{}}}, \; \forall k,j \geq i
\geq 1, \; \forall x \in [q_{\dinfty{},x}^i, \infty) \mbox{\em )}
\end{equation}
\noindent if $\{\alpha_i\}_{i \in \mathbb{N}}$ is decreasing
{\em (resp.} increasing{\em )}, with $\{\varepsilon_i\}_{i \in \mathbb{N}
  \cup \{\infty, \dinfty{}\}}$ monotonic decreasing and convergent
to $0$.
\end{thm}

\begin{pf}
Let us first suppose $\{\alpha_i\}_{i \in \mathbb{N}}$ is
decreasing. As shown in the left-hand-side of
Figure~\ref{fig-residual-subintervals-decreasing}, we follow from the
construction of ${\mathcal A}_{\infty}^\pi[{\mathcal
    D}^{\mathcal{K}}_{\sigma}]$ that
\begin{equation}
q_{i,y}^{i-1} := {\mathcal A}_i^\pi[{\mathcal D}^{\mathcal
    {K}}_{\sigma}](q_{i,x}^{i-1}) \leq {\mathcal A}_{l}^\pi[{\mathcal
    D}^{\mathcal{K}}_{\sigma}](x) \leq \alpha_i, \; \; \forall l \geq
i \geq 4, \; \forall x \in [q_{i,x}^{i-1}, \infty)
\end{equation}
\noindent from which
Equation~\ref{equation-correctness-trace-decreasing} is
trivial. Following
Theorem~\ref{th-uniform-convergence-trace}, as
$\{q_{i,y}^{i-1}\}_{i \geq 4}$ is increasing and $\{\alpha_i\}_{i \in
  \mathbb{N}}$ decreasing with $q_{i,y}^{i-1} < \alpha_i$, the
sequence $\{\varepsilon_i\}_{i \in \mathbb{N} \cup \{\infty,
  \dinfty{}\}}$ is decreasing. Since $\lim \limits_{i \rightarrow
  \infty} \varepsilon_i := \lim \limits_{i \rightarrow \infty}
q_{i,y}^{i-1} - \lim \limits_{i \rightarrow \infty} \alpha_i =
\alpha_{\dinfty{}} - \alpha_{\dinfty{}} = 0 $, the thesis is then
proved. \\

\noindent In an analogous manner, as shown in the left-hand-side of
Figure~\ref{fig-residual-subintervals-increasing} and when $\{\alpha_i\}_{i
  \in \mathbb{N}}$ increasing, it verifies
\begin{equation}
q_{\dinfty{},y}^{i} := {\mathcal A}_{\dinfty{}}[{\mathcal D}^{\mathcal
    {K}}_{\sigma}](q_{\dinfty{},x}^{i}) \leq {\mathcal A}_{l}^\pi[{\mathcal
    D}^{\mathcal{K}}_{\sigma}](x) \leq \alpha_{\dinfty{}}, \; \; \forall l \geq
i \geq 1, \; \forall x \in [q_{\dinfty{},x}^i, \infty)
\end{equation}
\noindent from which
Equation~\ref{equation-correctness-trace-increasing}
is trivial. Following a similar reasoning to the one applied to
$\{q_{i,y}^{i-1}\}_{i \geq 4}$ in
Theorem~\ref{th-uniform-convergence-trace}, we deduce that
$\{q_{\dinfty{},y}^{i}\}_{i \geq 1}$ is monotonic (decreasing), so
$\{\varepsilon_i\}_{i \in \mathbb{N} \cup \{\infty, \dinfty{}\}}$ is
monotonic decreasing. Finally, $\lim \limits_{i \rightarrow \infty}
\varepsilon_i := \lim \limits_{i \rightarrow \infty} {\mathcal
  A}_{\dinfty{}}[{\mathcal D}^{\mathcal
    {K}}_{\sigma}](q_{\dinfty{},y}^{i}) - \alpha_{\dinfty{}} = \lim
\limits_{x \rightarrow \infty} {\mathcal A}_{\dinfty{}}[{\mathcal
    D}^{\mathcal {K}}_{\sigma}](x) - \alpha_{\dinfty{}} :=
\alpha_{\dinfty{}} - \alpha_{\dinfty{}} = 0$. $\blacksquare$
\end{pf}

Given that \(q_{i,x}^{j} < \min_{} \{\absd{{\mathcal D}_i},
\absd{{\mathcal D}_j}\}, \; \forall i, j \in \mathbb{N} \cup \{\infty,
\dinfty{}\}, i \neq j\), this result meets our
requirements. Unfortunately, it has only a practical reading in the
first case, when the convergence is decreasing. Otherwise, the upper
bound depends on $\alpha_{\dinfty{}}$, which is the final value for
accuracy we want to estimate and thus is unknown. In order to break
the deadlock, we have no option but to find a criterion for correctness
based on the individual behavior of each learning trend as part of the
approximation process.

\begin{df}
\label{def-layer-of-convergence-learning-trend}
Let ${\mathcal A}^\pi[{\mathcal D}^{\mathcal {K}}_{\sigma}]$ be a
learning trace with asymptotic backbone $\{\alpha_i\}_{i \in
  \mathbb{N}}$. We define the {\em layer of convergence for}
${\mathcal A}_i^\pi[{\mathcal D}^{\mathcal {K}}_{\sigma}], \; i \in
\mathbb{N}$ as the value $\chi({\mathcal A}_i^\pi[{\mathcal
    D}^{\mathcal {K}}_{\sigma}]) := \abs{{\mathcal A}_i^\pi[{\mathcal
      D}^{\mathcal {K}}_{\sigma}](\absd{{\mathcal D}_i}) - \alpha_i}$.
\end{df}

This concept allows us to measure the contribution of each learning
trend to the convergence process, which provides the key for a
practical interpretation of the correctness, regardless of the type of
monotony associated to the asymptotic backbone.

\begin{thm}
\label{th-layered-correctness-trace}
{\em (Layered Correctness)} Let ${\mathcal A}^\pi[{\mathcal
    D}^{\mathcal {K}}_{\sigma}]$ be a learning trace with asymptotic
backbone $\{\alpha_i\}_{i \in \mathbb{N}}$. We then have that
\begin{equation}
\forall \varepsilon > 0, \; \exists n \in \mathbb{N}, \mbox{ such that
} [\chi({\mathcal A}_i^\pi[{\mathcal D}^{\mathcal {K}}_{\sigma}])
  \leq \varepsilon \Leftrightarrow i \geq n] 
\end{equation}
\end{thm}

\begin{pf}
Following the proof of
Theorem~\ref{th-uniform-convergence-trace} both sequences
$\{\alpha_i\}_{i \in \mathbb{N}}$ and $\{{\mathcal A}_i^\pi[{\mathcal
    D}^{\mathcal {K}}_{\sigma}](\absd{{\mathcal D}_i})\}_{i \in
  \mathbb{N}}$ converge to $\alpha_{\dinfty{}}$. Consequently
$\{\chi({\mathcal A}_i^\pi[{\mathcal D}^{\mathcal
    {K}}_{\sigma}]\}_{i \in \mathbb{N}}$ converges to 0 and, having
fixed $\varepsilon > 0$, we can consider the first level $n$ for which
$\chi({\mathcal A}_n^\pi[{\mathcal D}^{\mathcal {K}}_{\sigma}]) \leq
\varepsilon$. It then trivially verifies the inequality
$\chi({\mathcal A}_i^\pi[{\mathcal D}^{\mathcal {K}}_{\sigma}]) >
\varepsilon$, when $i < n$. \\

\noindent To prove now that $\chi({\mathcal A}_i^\pi[{\mathcal
    D}^{\mathcal {K}}_{\sigma}]) \leq \varepsilon, \; \forall i \geq
n$, it is enough to take into account that, by construction, the
sequence $\{{\mathcal A}_i^{\pi'}[{\mathcal D}^{\mathcal
    {K}}_{\sigma}](\absd{{\mathcal D}_i})\}_{i \in \mathbb{N}}$ is
monotonic decreasing and the trends are particular configurations of
the same accuracy pattern. $\blacksquare$
\end{pf}

Intuitively, this result determines the level from which the trends to
be calculated estimate the final accuracy with a gain, on the current
approximation, below a given threshold. As the layers converge
strictly decreasing to 0 in synchrony with the uniform convergence of
the learning trace, they introduce an alternative to absolute error and
convergence thresholds, which in this case have proved to be
impracticable. All this makes it possible for us to fully exploit the
model of convergence, giving practical sense to the
proposal. Nonetheless, the real nature of the observations obliges us to
keep in mind some additional considerations.

\subsection{Robustness}

Although the previous results certify that our proposal is
theoretically correct with respect to its original specification,
ensuring the termination in all cases, its practical application is
subject to conditions that could affect this correctness. In
particular, our abstract model has been built on an ideal
conceptualisation. This implies the assumption of a series of formal
properties for the learning curves, although they may slightly diverge
from this modelisation. The problem moves then away from our working
hypotheses, making it necessary to assess how far the methodology
described is engaged. For that, we henceforth assume that any learning
curve ${\mathcal A}_{\dinfty{}}[{\mathcal D}^{\mathcal {K}}_{\sigma}]$
is positive definite and upper bounded by 100, two conditions whose
compliance we can always guarantee, but only quasi-strictly increasing
and concave. These premises are now our \textit{testing hypotheses},
which capture the notion of irregular observation.

The new conditions can alter the monotony of the asymptotic backbone,
translating it into a quasi-monotony. This is not a minor problem
because we are talking about the key to proving the uniform
convergence in a learning trace, while this type of disorders generate
only local disturbs and does not affect the convergence itself. The
question then focuses on how to reduce the impact on the proximity
criterion associated to the correctness, as is the case for the layers
of convergence. For our study, we distinguish two types of
alterations, according to their position in relation to the {\em
  working level}, namely the instance from which these possible
dysfunctions would have an acceptable impact. Unfortunately, its
optimal location depends on unpredictable factors such as the
magnitude, evolution and the very existence of these
disorders. Consequently, a theoretically well-founded characterization
of this level is impossible and the formalization of heuristic
criteria is the only way out.

\begin{df}
\label{def-level-of-work-trace}
Let ${\mathcal A}^\pi[{\mathcal D}^{\mathcal {K}}_{\sigma}]$ be a
learning trace with asymptotic backbone $\{\alpha_i\}_{i \in
  \mathbb{N}}$, $\nu \in (0, 1)$, $\varsigma \in \mathbb{N}$ and
$\lambda \in \mathbb{N} \cup \{0\}$. We define the {\em working level
  for} ${\mathcal A}^\pi[{\mathcal D}^{\mathcal {K}}_{\sigma}]$ {\em
  with verticality threshold} $\nu$, {\em slow down} $\varsigma$
{\em and look-ahead} $\lambda$, as the smallest
$\omega(\nu,\varsigma,\lambda) \in \mathbb{N}$ verifying
\begin{equation}
\label{equation-permissible-verticality-trace}
\frac{\sqrt[\varsigma]{\nu}}{1 - \nu} \geq \frac{\abs{\alpha_{i+1} -
    \alpha_{i}}}{\absd{{\mathcal D}_{i+1}} - \absd{{\mathcal D}_{i}}},
\; \forall i \in \mathbb{N} \mbox{ such that } \omega(\nu,\varsigma,\lambda) \leq i \leq
\omega(\nu,\varsigma,\lambda) + \lambda
\end{equation}
\noindent The smallest $\wp(\nu,\varsigma,\lambda) \geq
\omega(\nu,\varsigma,\lambda)$ with
$\alpha_{\wp(\nu,\varsigma,\lambda)} \leq 100$ is the {\em prediction
  level for} ${\mathcal A}^\pi[{\mathcal D}^{\mathcal {K}}_{\sigma}]$.
\end{df}

\begin{figure}[htbp]
\begin{center}
\begin{tabular}{cc}
\hspace*{-.7cm}
\epsfxsize=.52\linewidth
\epsffile{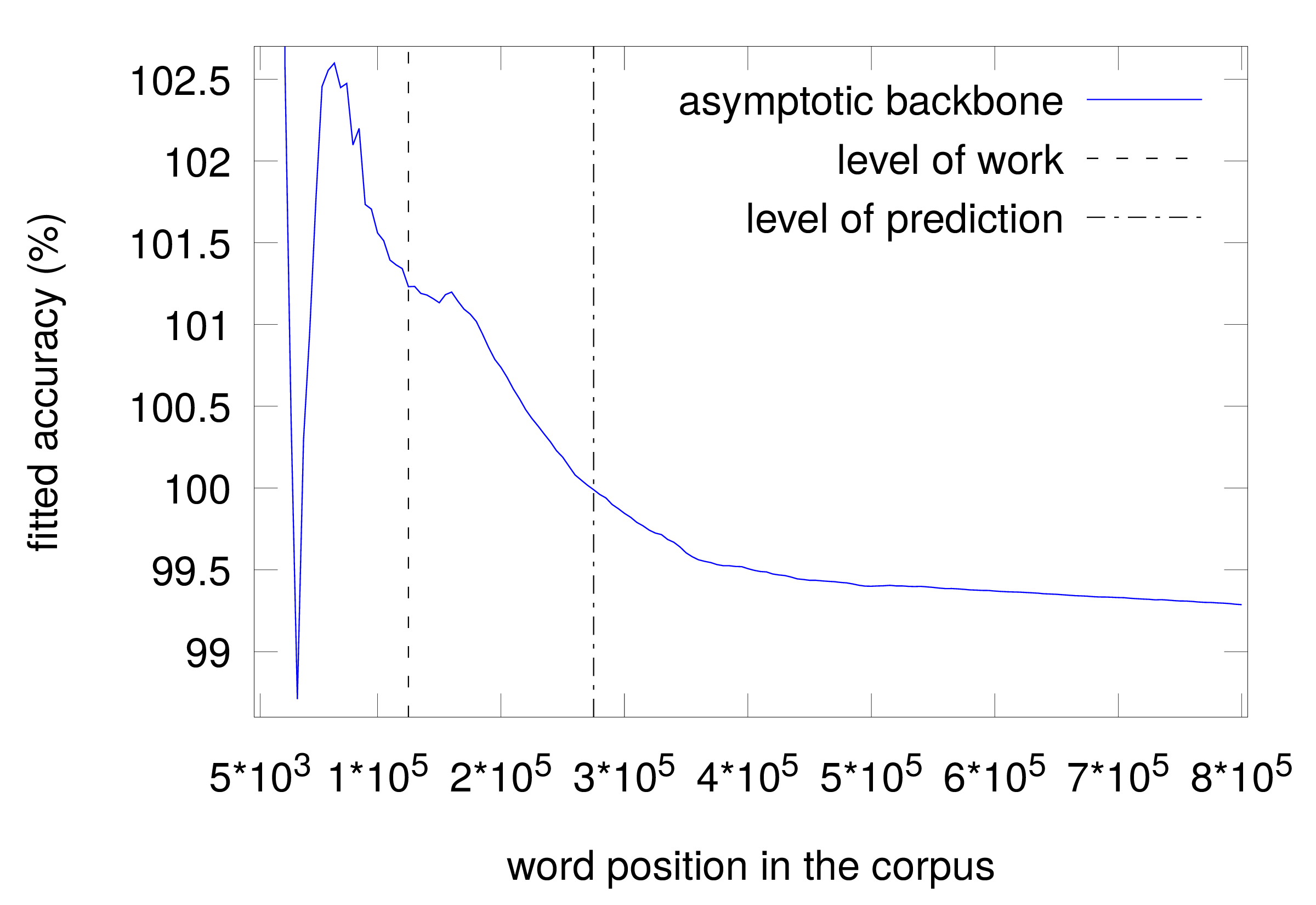} 
&
\hspace*{-.7cm}
\epsfxsize=.52\linewidth
\epsffile{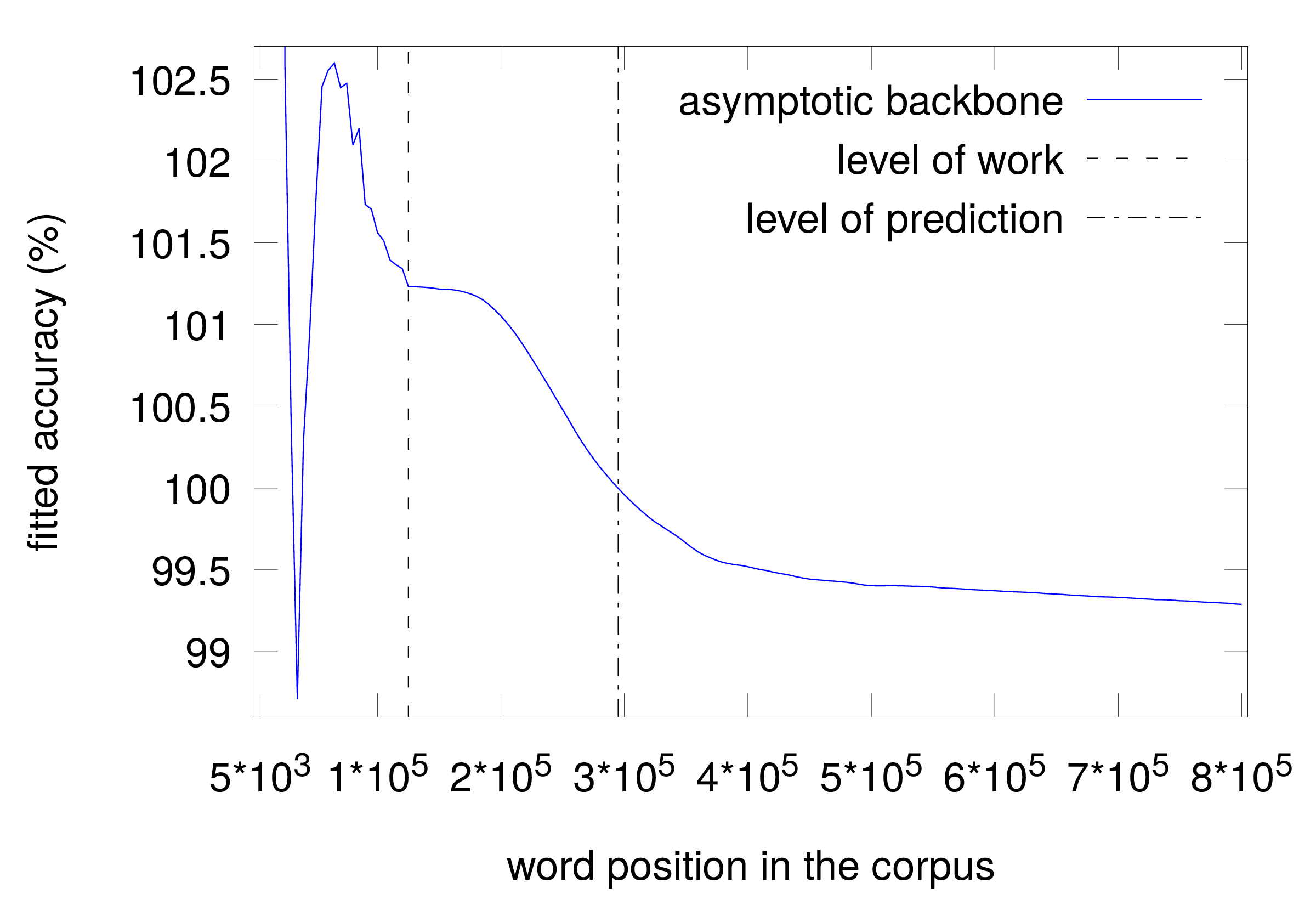}
\end{tabular}
\caption{Working and prediction levels for asymptotic backbones built 
  without and with anchors for fn{\sc tbl} on {\sc f}rown.}
\label{fig-lashes-trace-and-level-sequence-nlls-fnTBL-frown-5000-800000}
\end{center}
\end{figure}

We therefore determine the working level from the normalization $\nu
\in (0, 1)$ of the maximum permissible absolute value for the slope of
the straight line joining two consecutive points on the asymptotic
backbone. Since such values decrease as the training advances and the
irregularities are inversely proportional to the degree of learning
achieved, both magnitudes correlate. This allows us to reasonably
categorize such alterations, using an optional extra degree of
verification given by the look-ahead window. Retaking our example, the
difference between the alterations in the monotony of the asymptotic
backbone before and after the working level can be seen in the
left-most diagram of
Figure~\ref{fig-lashes-trace-and-level-sequence-nlls-fnTBL-frown-5000-800000},
with $\nu=2*10^{-5}$, $\varsigma=1$ and $\lambda=5$ as
parameters. Furthermore, although the condition on the upper bound for
the prediction level does not have any impact on the convergence we
are studying, it allows us to focus on those learning trends that
could be considered as feasible approximations for accuracy, whose
highest value is 100. In practice, the working level is often aligned
with the prediction one because the location of the former usually
permits us to gain the time needed to stabilize the convergence
process below such a maximum. Trivially, when the working hypotheses
are fulfilled, the condition characterizing the working level verifies
from the first operative level $\ell=3$, whatever the parameters
considered.

\subsubsection{Irregularities before the working level}

The low number of observations available, combined with the fact that
they are associated with steep slopes, has here a multiplying effect
on the irregularities on the asymptotic backbone, causing wide
fluctuations. The use of large enough sets of observations to generate
the learning trends should help to mitigate the problem, which is
formally equivalent to estimate the working level. That is, the only
effective strategy in this case to avoid alterations in the monotony
of the asymptotic backbone is to discard all the trends associated to
pre-working levels, as can be seen from the left-most diagram of
Figure~\ref{fig-lashes-trace-and-level-sequence-nlls-fnTBL-frown-5000-800000}.

\subsubsection{Irregularities after the working level}

The irregularities after the working level should be less pronounced,
below the verticality threshold chosen, unless they are caused by a
radical and continued shift away from the working hypotheses. In this
case, the monotony of the asymptotic backbone can even be reversed
until the training process allows the system to rebalance it. We are
then talking about a problem outside our testing scenario, because its
treatment requires the resampling of the training data. Otherwise,
when we deal with small variations in the monotony, the access to an
adequate extra observation could facilitate the realignment of the
asymptotic backbone. We must here take into account that, being the
learning trends the result of a fitting process, the impact of that
additional example will be greater the further the associated instance
is. In order to formalize this idea, we first extend the notion of
learning trace.

\begin{df}
\label{def-anchoring-learning-trace}
Let ${\mathcal A}^\pi[{\mathcal D}^{\mathcal {K}}_{\sigma}]$ be a
learning trace with working level $\omega(\nu,\varsigma,\lambda)$, and
$\{\hat{\mathcal A}_{\ell}^\pi[{\mathcal
    D}^{\mathcal{K}}_{\sigma}](\infty)\}_{\ell >
  \omega(\nu,\varsigma,\lambda)} \subset \mathbb{R}^+$ a sequence. A
{\em learning trend of level} $\ell > \omega(\nu,\varsigma,\lambda)$
{\em with anchor} $\hat{\mathcal A}_\ell^\pi[{\mathcal D}^{\mathcal
    {K}}_{\sigma}](\infty)$ {\em for} ${\mathcal
  A}_{\dinfty{}}[{\mathcal D}^{\mathcal{K}}_{\sigma}]$ {\em using the
  accuracy pattern} $\pi$, is a curve $\hat{\mathcal
  A}_{\ell}^\pi[{\mathcal D}^{\mathcal{K}}_{\sigma}] \in \pi$ fitting
the observations $\{[x_i, {\mathcal A}_{\dinfty{}}[{\mathcal
      D}^{\mathcal {K}}_{\sigma}](x_i)], \; x_i := \absd{\mathcal D_i}
\}_{i=1}^{\ell} \cup \; [\infty, \hat{\mathcal A}_{\ell}^\pi[{\mathcal
      D}^{\mathcal{K}}_{\sigma}](\infty)]$. We then denote by
$\hat\rho_\ell(i) := [{\mathcal A}_{\dinfty{}}[{\mathcal
      D}^{\mathcal{K}}_{\sigma}] - \hat{\mathcal
    A}_{\ell}^\pi[{\mathcal
      D}^{\mathcal{K}}_{\sigma}]](\absd{{\mathcal D}_i})$ the {\em
  residual of} $\hat{\mathcal A}_{\ell}^\pi[{\mathcal
    D}^{\mathcal{K}}_{\sigma}]$ {\em at the level} $i \in \mathbb{N}$,
by $\hat\rho_\ell(\infty) := \hat{\mathcal A}_{\ell}^\pi[{\mathcal
    D}^{\mathcal{K}}_{\sigma}](\infty) - \hat\alpha_\ell$ its {\em
  residual at the point of infinity} and by $y = \hat\alpha_\ell$ its
asymptote.

When $\{\hat\alpha_\ell\}_{\ell > \omega(\nu,\varsigma,\lambda)}$ is
positive definite and converges monotonically to the asymptotic value
$\alpha_{\dinfty{}}$ of ${\mathcal A}_{\dinfty{}}^\pi[{\mathcal
    D}^{\mathcal {K}}_{\sigma}]$, we say that $\hat{\mathcal
  A}^\pi[{\mathcal D}^{\mathcal {K}}_{\sigma}] := \{\hat{\mathcal
  A}_{\ell}^\pi[{\mathcal D}^{\mathcal{K}}_{\sigma}]\}_{\ell >
  \omega(\nu,\varsigma,\lambda)}$ is an {\em anchoring learning
  trace}, whose {\em reference} is $[{\mathcal A}^\pi[{\mathcal
      D}^{\mathcal {K}}_{\sigma}], \omega(\nu,\varsigma,\lambda)]$ 
with $\{\hat\alpha_\ell\}_{\ell >
  \omega(\nu,\varsigma,\lambda)}$ its {\em asymptotic backbone}.
\end{df}

The only difference between an anchoring learning trace and a standard
one is the use of extra fitting points in the infinity in order to
generate its learning trends. In particular, as the properties of
monotony and convergence for the asymptotic backbone must be
preserved, the previous findings on uniform convergence, correctness
and layer correctness for learning traces also
verify. Accordingly, the conclusions and even the formal proofs for
Theorems~\ref{th-uniform-convergence-trace},~\ref{th-correctness-trace}
and~\ref{th-layered-correctness-trace} remain fully valid when we use
anchors, which is why we do not repeat them. At this point, the next
step in achieving a mechanism to soften the impact of irregular
observations in the asymptotic backbone is to determine the real
potential of an anchor in this regard.

\begin{thm}
\label{th-residuals-anchoring-traces}
Let $\hat{\mathcal A}^\pi[{\mathcal D}^{\mathcal {K}}_{\sigma}]$ be an
anchoring learning trace. We have that $\hat\rho_i(\infty) = -
\sum_{j \leq i} \hat\rho_i(j), \; \forall i >
\omega(\nu,\varsigma,\lambda)$, and $\lim \limits_{i \rightarrow
  \infty} \hat\rho_i(\infty) = 0$.
\end{thm}

\begin{pf}
Let us assume $\{\hat\alpha_i\}_{i \in \mathbb{N}}$ the asymptotic
backbone for $\hat{\mathcal A}^\pi[{\mathcal D}^{\mathcal
    {K}}_{\sigma}]$. Having fixed a level $i >
\omega(\nu,\varsigma,\lambda)$, the fitting mechanism ensures that
$\sum_{j \leq i} \hat\rho_i(j) + \hat\rho_i(\infty) = 0$ because the
sum total of residuals on $\hat{\mathcal A}_i^\pi[{\mathcal
    D}^{\mathcal {K}}_{\sigma}]$ must be null, which proves the first
part of the wording. As $\hat\rho_i(\infty) := \hat{\mathcal
  A}_i^\pi[{\mathcal D}^{\mathcal {K}}_{\sigma}](\infty) -
\hat\alpha_i, \; \forall i > \omega(\nu,\varsigma,\lambda)$, with both
sequences $\{\hat{\mathcal A}_i^\pi[{\mathcal D}^{\mathcal
    {K}}_{\sigma}](\infty)\}_{i > \omega(\nu,\varsigma,\lambda)}$ and
$\{\hat\alpha_i\}_{i > \omega(\nu,\varsigma,\lambda)}$ converging to
$\alpha_{\dinfty{}}$, we conclude that $\lim \limits_{i \rightarrow
  \infty} \hat\rho_i(\infty) = 0$.  $\blacksquare$
\end{pf}

This suggests that an intelligent use of anchors can reduce the
magnitude of distortions in the asymptotic backbone, by compensating
for their associated residuals. Thus, having fixed a learning trend,
the degree of smoothing applicable to the irregularities correlates
with its residual at the point of infinity. In other words, the better
the estimation of the asymptotes by the anchors, the lower the impact
of inconsistencies on the monotony of the asymptotic backbone. The
question now is, therefore, how to choose the optimal sequence of
anchors. Considering that the abstract model is iterative, our
approach should take into account the impact of the new observation
incorporated at each cycle, which excludes strategies based on static
analysis.

\begin{thm}
\label{th-anchoring-trace-correction}
Let $\hat{\mathcal A}^\pi[{\mathcal D}^{\mathcal {K}}_{\sigma}]$ be an
anchoring learning trace with asymptotic backbone $\{\hat\alpha_i\}_{i
  > \omega(\nu,\varsigma,\lambda)}$. We then have that:
\begin{equation}
\label{equation-anchoring-trace-correction}  
   \; \forall i > \omega(\nu,\varsigma,\lambda), \; \hat{\mathcal
     A}_{i+1}^\pi[{\mathcal D}^{\mathcal {K}}_{\sigma}](\infty) \leq
   \hat\alpha_{i} - \sum_{j \leq i+1} \hat\rho_{i+1}(j) \; \;
   (\mbox{\em resp. } \hat{\mathcal A}_{i+1}^\pi[{\mathcal
       D}^{\mathcal {K}}_{\sigma}](\infty) \geq \hat\alpha_{i} -
   \sum_{j \leq i+1} \hat\rho_{i+1}(j))
\end{equation}
\noindent if $\{\hat\alpha_i\}_{i > \omega(\nu,\varsigma,\lambda)}$
decreasing {\em (resp.} increasing{\em )}.
\end{thm}

\begin{pf}
In the decreasing case, we have that $\hat\alpha_{i+1} \leq
\hat\alpha_{i}, \; \forall i > \omega(\nu,\varsigma,\lambda)$. As
$\hat{\mathcal A}_{i+1}^\pi[{\mathcal
    D}^{\mathcal{K}}_{\sigma}](\infty) - \hat\alpha_{i+1} :=
\hat\rho_{i+1}(\infty) = - \sum_{j \leq i+1} \hat\rho_{i+1}(j), \;
\forall i > \omega(\nu,\varsigma,\lambda)$, the thesis derives
immediately. The increasing case is analogous. $\blacksquare$
\end{pf}

Since Theorem~\ref{th-residuals-anchoring-traces} stated
that, excluding the one at the point of infinity, the residuals
accumulated by the observations of the trends in an anchoring learning
trace converge to zero, this last result opens the doors to a
systematic way to generate anchors from the last asymptotic values 
calculated.

\begin{thm}
\label{th-canonical-anchoring-trace} 
Let ${\mathcal A}^\pi[{\mathcal D}^{\mathcal {K}}_{\sigma}]$ be a
learning trace with asymptotic backbone $\{\alpha_i\}_{i \in
  \mathbb{N}}$, and $\{\hat{\mathcal A}_i^\pi[{\mathcal D}^{\mathcal
    {K}}_{\sigma}](\infty)\}_{i > \omega(\nu,\varsigma,\lambda)}$ the
sequence defined by
\begin{equation}
   \hat{\mathcal A}_{\omega(\nu,\varsigma,\lambda)+1}^\pi[{\mathcal
       D}^{\mathcal{K}}_{\sigma}](\infty) :=
   \alpha_{\omega(\nu,\varsigma,\lambda)} \hspace*{.75cm}
   \hat{\mathcal A}_{i+1}^\pi[{\mathcal
       D}^{\mathcal{K}}_{\sigma}](\infty) := \hat\alpha_{i} := \lim
   \limits_{x \rightarrow \infty} \hat{\mathcal A}_{i}^\pi[{\mathcal
       D}^{\mathcal{K}}_{\sigma}](x)
\end{equation}
\noindent with $\hat{\mathcal A}_{i}^\pi[{\mathcal
    D}^{\mathcal{K}}_{\sigma}]$ a curve fitting $\{[x_j, {\mathcal
    A}_{\dinfty{}}[{\mathcal D}^{\mathcal {K}}_{\sigma}](x_j)], \; x_j
:= \absd{\mathcal D_j} \}_{j=1}^{i} \cup \; [\infty, \hat{\mathcal
    A}_{i}^\pi[{\mathcal D}^{\mathcal{K}}_{\sigma}](\infty)]$,
$\forall i > \omega(\nu,\varsigma,\lambda)$. Then
$\alpha_{\omega(\nu,\varsigma,\lambda) + i} \leq
\hat\alpha_{\omega(\nu,\varsigma,\lambda) + i}$ {\em (resp.}
$\alpha_{\omega(\nu,\varsigma,\lambda) + i} \geq
\hat\alpha_{\omega(\nu,\varsigma,\lambda) + i}${\em)}, $\forall i \in
\mathbb{N}$, when $\{\alpha_i\}_{i \in \mathbb{N}}$ is decreasing {\em
  (resp.} increasing {\em)}. Furthermore, $\{\hat{\mathcal
  A}_i^\pi[{\mathcal D}^{\mathcal {K}}_{\sigma}]\}_{i >
  \omega(\nu,\varsigma,\lambda)}$ is an anchoring learning trace of
reference $[{\mathcal A}^\pi[{\mathcal D}^{\mathcal
      {K}}_{\sigma}],\omega(\nu,\varsigma,\lambda)]$. We call
$\{\hat{\mathcal A}_i^\pi[{\mathcal D}^{\mathcal
    {K}}_{\sigma}](\infty)\}_{i > \omega(\nu,\varsigma,\lambda)}$ the
set of {\em canonical anchors} for $\hat{\mathcal A}^\pi[{\mathcal
    D}^{\mathcal {K}}_{\sigma}]$.
\end{thm}

\begin{pf}
\noindent As both the decreasing and the increasing cases are
analogous, we only detail the former. We first demonstrate both that
$\alpha_{\omega(\nu,\varsigma,\lambda) + i} \leq
\hat\alpha_{\omega(\nu,\varsigma,\lambda) + i}$ and $\hat{\mathcal
  A}_{\omega(\nu,\varsigma,\lambda) + i+1}^\pi[{\mathcal D}^{\mathcal
    {K}}_{\sigma}](\infty) \leq \hat{\mathcal
  A}_{\omega(\nu,\varsigma,\lambda) + i}^\pi[{\mathcal D}^{\mathcal
    {K}}_{\sigma}](\infty), \; \forall i \in \mathbb{N}$. We do so by
induction on $i$. Let us first assume $i=1$, as
$\hat\alpha_{\omega(\nu,\varsigma,\lambda) + 1} := \lim \limits_{x
  \rightarrow \infty} \hat{\mathcal A}_{\omega(\nu,\varsigma,\lambda)
  + 1}^\pi[{\mathcal D}^{\mathcal {K}}_{\sigma}](x)$ and the latter is
a curve fitting the set of values
\begin{equation}
\{[x_j, {\mathcal A}_{\dinfty{}}[{\mathcal D}^{\mathcal
      {K}}_{\sigma}](x_j)], \; x_j := \absd{\mathcal D_j}
\}_{j=1}^{\omega(\nu,\varsigma,\lambda)+1} \; \cup \; \{[\infty, \hat{\mathcal
  A}_{\omega(\nu,\varsigma,\lambda) + 1}^\pi[{\mathcal
    D}^{\mathcal{K}}_{\sigma}](\infty)]\}
\end{equation}
\noindent or, in other words, the series 
\begin{equation}
\{[x_j, {\mathcal A}_{\dinfty{}}[{\mathcal D}^{\mathcal
      {K}}_{\sigma}](x_j)], \; x_j := \absd{\mathcal D_j}
\}_{j=1}^{\omega(\nu,\varsigma,\lambda)+1}
\cup
\{[\infty, \alpha_{\omega(\nu,\varsigma,\lambda)}]\}
\end{equation}
\noindent where, as $\{\alpha_i\}_{i \in \mathbb{N}}$ is 
decreasing, we have that $\lim \limits_{x \rightarrow \infty}
{\mathcal A}_{\omega(\nu,\varsigma,\lambda) + 1}^\pi[{\mathcal
    D}^{\mathcal {K}}_{\sigma}](x):=
\alpha_{\omega(\nu,\varsigma,\lambda) + 1} \leq
\alpha_{\omega(\nu,\varsigma,\lambda)}$ and we immediately deduce
\begin{equation}
\begin{array}{llll}
\alpha_{\omega(\nu,\varsigma,\lambda)+1} & \leq & \lim \limits_{x
  \rightarrow \infty} \hat{\mathcal
  A}_{\omega(\nu,\varsigma,\lambda)+1}[{\mathcal
    D}^{\mathcal{K}}_{\sigma}](x) :=
\hat\alpha_{\omega(\nu,\varsigma,\lambda)+1} := \hat{\mathcal
  A}_{\omega(\nu,\varsigma,\lambda) + 2}[{\mathcal
    D}^{\mathcal{K}}_{\sigma}](\infty) & \leq \\
& \leq &
\alpha_{\omega(\nu,\varsigma,\lambda)} := \hat{\mathcal
  A}_{\omega(\nu,\varsigma,\lambda)+1}[{\mathcal
    D}^{\mathcal{K}}_{\sigma}](\infty)
\end{array}
\end{equation}
\noindent which proves this first case. \\

\noindent Let us now assume that $\hat{\mathcal
  A}_{\omega(\nu,\varsigma,\lambda) + i+1}^\pi[{\mathcal D}^{\mathcal
    {K}}_{\sigma}](\infty) \leq \hat{\mathcal
  A}_{\omega(\nu,\varsigma,\lambda) + i}^\pi[{\mathcal D}^{\mathcal
    {K}}_{\sigma}](\infty), \; \forall i < n \in \mathbb{N}$. We must
now prove that $\hat{\mathcal A}_{\omega(\nu,\varsigma,\lambda) +
  n+1}^\pi[{\mathcal D}^{\mathcal {K}}_{\sigma}](\infty) \leq
\hat{\mathcal A}_{\omega(\nu,\varsigma,\lambda) + n}^\pi[{\mathcal
    D}^{\mathcal {K}}_{\sigma}](\infty)$. \\

\noindent As $\hat\alpha_{\omega(\nu,\varsigma,\lambda) + n} :=
\lim \limits_{x \rightarrow \infty} \hat{\mathcal
  A}_{\omega(\nu,\varsigma,\lambda) + n}^\pi[{\mathcal D}^{\mathcal
    {K}}_{\sigma}](x)$ and the latter is a curve fitting the 
set of values
\begin{equation}
\{[x_j, {\mathcal A}_{\dinfty{}}[{\mathcal D}^{\mathcal
      {K}}_{\sigma}](x_j)], \; x_j := \absd{\mathcal D_j}\}_{j=1}^{\omega(\nu,\varsigma,\lambda)+n} \; \cup \; \{[\infty, \hat{\mathcal
  A}_{\omega(\nu,\varsigma,\lambda) + n}^\pi[{\mathcal
    D}^{\mathcal{K}}_{\sigma}](\infty)]\}
\end{equation}
\noindent or, in other words, the series 
\begin{equation}
\{[x_j, {\mathcal A}_{\dinfty{}}[{\mathcal D}^{\mathcal
      {K}}_{\sigma}](x_j)], \; x_j := \absd{\mathcal D_j}\}_{j=1}^{\omega(\nu,\varsigma,\lambda)+n} \; \cup
\{[\infty, \hat\alpha_{\omega(\nu,\varsigma,\lambda) + n -1}]\}
\end{equation}
\noindent where, as $\{\alpha_i\}_{i \in \mathbb{N}}$ is 
decreasing, we have that $\lim \limits_{x \rightarrow \infty}
{\mathcal A}_{\omega(\nu,\varsigma,\lambda) + n}^\pi[{\mathcal
    D}^{\mathcal {K}}_{\sigma}](x):=\alpha_{\omega(\nu,\varsigma,\lambda)+n}
\leq \alpha_{\omega(\nu,\varsigma,\lambda)+n-1}$. Since, by induction
hypothesis, it also verifies that
$\alpha_{\omega(\nu,\varsigma,\lambda)+n-1} \leq
\hat\alpha_{\omega(\nu,\varsigma,\lambda)+n-1}$, we immediately deduce
\begin{equation}
\begin{array}{llll}
\alpha_{\omega(\nu,\varsigma,\lambda)+n-1} & \leq & \lim \limits_{x
  \rightarrow \infty} \hat{\mathcal
  A}_{\omega(\nu,\varsigma,\lambda)+n}[{\mathcal
    D}^{\mathcal{K}}_{\sigma}](x) :=
\hat\alpha_{\omega(\nu,\varsigma,\lambda)+n} := \hat{\mathcal
  A}_{\omega(\nu,\varsigma,\lambda) + n+1}[{\mathcal
    D}^{\mathcal{K}}_{\sigma}](\infty)
& \leq \\
& \leq &
\hat\alpha_{\omega(\nu,\varsigma,\lambda)+n-1} := \hat{\mathcal
  A}_{\omega(\nu,\varsigma,\lambda)+n}[{\mathcal
    D}^{\mathcal{K}}_{\sigma}](\infty)
\end{array}
\end{equation}
\noindent which proves that $\{\hat{\alpha}_i\}_{i >
  \omega(\nu,\varsigma,\lambda)}$ is, as the reference asymptotic
backbone $\{\alpha_i\}_{i \in \mathbb{N}}$, positive definite and
monotonic decreasing. Given that it is a infimum for this
sequence, the latter converges to $\alpha_{\dinfty{}}$ and the thesis
is proved. $\blacksquare$
\end{pf}

Regardless of the consideration of other types of anchors, we are
already in a position to transfer their practical use for smoothing
small irregularities, as should be usual once passed the working
level. Obviously, faced with the impossibility of representing the
value $\infty$ on the computer, we locate the anchors in a case as far
as possible for practical purposes. Returning to our running example
and identifying such a case with the word position $10^{200}$, this
technique is illustrated in
Figure~\ref{fig-lashes-trace-and-level-sequence-nlls-fnTBL-frown-5000-800000},
where the effects of using anchors in its right-most diagram are in
contrast with the results obtained in their absence and shown in the
left-most one. The result also shows that, as might have been
expected, anchors can cause changes in the speed of convergence of the
learning trace. In fact, given that these structures are necessarily
estimated from the experience provided by previously computed learning
trends, anchoring strategies tend to be conservative. In practice,
this has meant the slowing down of the process, as has just been
proved for the canonical case.

\section{The testing frame}
\label{section-testing-frame}

Given a training data base $\mathcal{D}$, the goal is to illustrate how
far in advance and how well a learning curve
$\mathcal{A}_{\dinfty{}}[\mathcal{D}^{\mathcal{K}}_{\sigma}]$, built
from a kernel $\mathcal{K}$ and using a step function $\sigma$, can be
approximated. We apply our proposal based on the convergence of
learning traces, without and with anchors.

\subsection{The monitoring structure}

As evaluation basis we introduce the {\em run}, a tuple
$\mathcal{E}=[\mathcal{A}^\pi[\mathcal{D}^{\mathcal{K}}_{\sigma}],
  \wp(\nu,\varsigma,\lambda),\tau]$ characterized by a learning trace
$\mathcal{A}^\pi[\mathcal{D}^{\mathcal{K}}_{\sigma}]$, a prediction
level $\wp(\nu,\varsigma,\lambda)$ and a convergence threshold
$\tau$. We then formalize our experimental study on two collections of
runs, ${\mathcal F} =\{{\mathcal E}_i\}_{i \in I}$ and ${\mathcal G}
=\{{\mathcal E}_j\}_{j \in J}$, of standard and anchoring learning
traces, respectively. The anchors are located in a case $10^{200}$,
sufficiently far for giving practical sense to the concept. With the
aim of avoiding distortions due to the lack of uniformity in the
testing frame, a common kernel $\mathcal{K}$, accuracy pattern $\pi$,
step function $\sigma$, verticality threshold $\nu$, slow down
$\varsigma$ and look-ahead $\lambda$ are used. In order to make the
results comparable, we also use a common value for the convergence
threshold $\tau$ on those runs involving the same training data base
and system tested, whether they are in ${\mathcal F}$ or ${\mathcal
  G}$.

In practice, we are interested in studying each run
$\mathcal{E}=[\mathcal{A}^\pi[\mathcal{D}^{\mathcal{K}}_{\sigma}],
  \wp(\nu,\varsigma,\lambda),\tau]$ from the level in which
predictions are below the convergence threshold $\tau$ and which we
call \textit{convergence level} ({\sc cl}evel) from now on. So, once
the prediction level ({\sc pl}evel) is found during the computation
of the learning trace
$\mathcal{A}^\pi[\mathcal{D}^{\mathcal{K}}_{\sigma}]$, we begin to
check the layer of convergence. When it reaches the convergence
threshold $\tau$, the corresponding trend
$\mathcal{A}_{\mbox{\footnotesize {\sc
      cl}evel}}^\pi[\mathcal{D}^{\mathcal{K}}_{\sigma}]$ is selected
for predicting the learning curve
$\mathcal{A}_{\dinfty{}}[\mathcal{D}^{\mathcal{K}}_{\sigma}]$, which
implies that the iterative process of approximation is stopped at that
instant. For the collection of runs ${\mathcal F} =\{{\mathcal
  E}_i\}_{i \in I}$ (resp. ${\mathcal G} =\{{\mathcal E}_j\}_{j \in
  J}$), monitoring is then applied to those learning trends
$\{\mathcal{A}_{\mbox{\footnotesize {\sc
      cl}evel}_i}^\pi[\mathcal{D}^{\mathcal{K}}_{\sigma}]\}_{i \in I}$
(resp. $\{\mathcal{A}_{\mbox{\footnotesize {\sc
      cl}evel}_j}^\pi[\mathcal{D}^{\mathcal{K}}_{\sigma}]\}_{j \in
  J}$) on a common set of \textit{control levels} for the training
data base, which are extracted from a finite sub-interval of the
\textit{prediction windows} $\{[\mbox{{\sc cl}evel}_i, \infty)\}_{i
    \in I}$ (resp. $\{[\mbox{{\sc cl}evel}_j, \infty)\}_{j \in
      J}$). We baptize these sets as \textit{control sequences} and in
    each of their levels the accuracy (Ac) and its estimate (EAc) are
    computed for each run. In order to guarantee the soundness of the
    results, we use 6 decimal digits, though only two are represented
    for reasons of space and visibility.

\subsection{The quality metrics}

We want to measure both the reliability of our estimates for accuracy
regarding the actual values and their robustness against variations in
our working hypotheses, in all case studies. For that, two groups of
metrics are used.

\subsubsection{Measuring the reliability}

A way of measuring the reliability is through the {\em mean absolute
  percent error} ({\sc mape})~\citep{Vandome63}. For every run
$\mathcal{E}$ and level $i$ of a control sequence $\mathcal{S}$, we
first compute the \textit{percentage error} ({\sc pe}) as the
difference between the EAc calculated from
$\mathcal{A}_{\mbox{\footnotesize {\sc cl}evel}_{\mathcal
    E}}^\pi[\mathcal{D}^{\mathcal{K}}_{\sigma}](i)$ and the Ac from
$\mathcal{A}_{\dinfty{}}[\mathcal{D}^{\mathcal{K}}_{\sigma}](i)$. We
can then express the {\sc mape} as the arithmetic mean of the unsigned
{\sc pe}, as
\begin{equation}
\mbox{\sc pe}(\mathcal{E})(i) := 100 *
\frac{[\mathcal{A}_{\mbox{\footnotesize
    {\sc cl}evel}_{\mathcal E}}^\pi -
      \mathcal{A}_{\dinfty{}}][\mathcal{D}^{\mathcal{K}}_{\sigma}](i)}
     {\mathcal{A}_{\dinfty{}}[\mathcal{D}^{\mathcal{K}}_{\sigma}](i)}, \;
     \mathcal{E} = [{\mathcal A}^\pi[{\mathcal D}^{\mathcal{K}}_{\sigma}],
                    \wp(\nu,\varsigma,\lambda), \tau],
     \; i \in {\mathcal S}
\end{equation}
\begin{equation}
\mbox{\sc mape}(\mathcal{E})({\mathcal S}) := 
\frac{100}{\absd{\mathcal S}} * 
\sum_{i \in {\mathcal S}} \abs{\mbox{\sc pe}(\mathcal{E})(i)}
\end{equation}

\noindent Intuitively, the error in the estimates done over a control
sequence is, on average, proportional to the {\sc mape}. Nonetheless,
while at first sight this measure could appear to be sufficient for
our purposes, it only provides quantitative information about the
estimation process. We also need to determine to what extent those
deviations impact decision-making on accuracy-based criteria, which is
a qualitative perspective.

In this context, once a set of runs ${\mathcal H}$ working on a common
training data base have been fixed, we are interested in calculating
the percentage of those for which such errors do not cause wrong
decisions to be made in looking for the run with the best performance.
To this end, having fixed a control sequence ${\mathcal S}$, the
reliability of one of these runs in relation to the others depends on
its estimates not altering the relative position of its learning curve
with respect to the learning curves associated to the rest of runs
throughout ${\mathcal S}$. We then say that such run is
\textit{locally reliable in} ${\mathcal H}$ \textit{along} ${\mathcal
  S}$. From this perspective, the fact that different runs have
similar {\sc mape} values may help to improve the reliability of one
from the other, but only when the corresponding asymptotic backbones
have the same type of monotony\footnote{This implies that the learning
  traces approximate the learning curves from a common relative
  position to all the runs.}. The reason is that the learning trends
used for predicting are then close to geometric translations of the
learning curves, on the basis of a common shifting vector. Otherwise,
the proximity between {\sc mape}s does not allow conclusions to be
drawn on reliability, which is unfortunately our case.

This justifies the need for a complementary evaluation view,
independent of the {\sc mape} concept and based on the local reliability
condition, though taking into account that calling for compliance with
the latter may be sometimes unrealistic. The reason is that it applies
to an entire control sequence, which may be an excessive degree of
requirement when comparing runs whose learning curves intersect within
their domain. More specifically, the error in the estimate of an
intersection point should be lower than the distance between its
neighbouring control levels, which is highly unlikely if such distance
is short. Therefore, we are here more interested in assessing the \textit{rate
of distortion} introduced when comparing runs, understood as the
percentage of control levels in which the estimates of a run preserve
the relative position of its learning curve with respect to the
other ones associated to the rest of runs.

Accordingly, we distinguish two testing scenarios from runs in
${\mathcal F}$ and ${\mathcal G}$. The former refers to the compliance
for the local reliability condition when the corresponding learning
curves are disjoint. The second analyzes the impact of prediction
errors in the comparison of runs whose learning curves
intersect. Given a control sequence ${\mathcal S}$ and a set
${\mathcal H}$ of runs defined on the same training data base, our
primary reference in both scenarios is the \textit{reliability
  estimation} ({\sc re}) \textit{of two runs} ${\mathcal
  E},\tilde{\mathcal E} \in {\mathcal H}$ \textit{on} $i \in {\mathcal
  S}$, defined as
\begin{equation}
\label{equation-re-intersection}
\begin{array}{c}
\mbox{{\sc re}}(\mathcal{E},\tilde{\mathcal{E}})(i) :=
\left\{ \begin{array}{l}
          1 \mbox{ if } \; [[\mathcal{A}_{\dinfty{}}-
                            \tilde{\mathcal{A}}_{\dinfty{}}]
              * [\mathcal{A}_{\mbox{\footnotesize {\sc cl}evel}_{\mathcal E}}^\pi -
                    \tilde{\mathcal{A}}_{\mbox{\footnotesize
    {\sc cl}evel}_{\tilde{\mathcal E}}}^\pi]][\mathcal{D}^{\mathcal{K}}_{\sigma}](i)
              \geq 0 \\
          0 \mbox{ otherwise}
        \end{array}
\right.
\end{array}
\end{equation}
\noindent with $\mathcal{E} = [{\mathcal A}^\pi[{\mathcal
      D}^{\mathcal{K}}_{\sigma}], \wp(\nu,\varsigma,\lambda), \tau]$,
$\tilde{\mathcal E} = [\tilde{\mathcal A}^\pi[{\mathcal
      D}^{\tilde{\mathcal K}}_{\tilde{\sigma}}],
  \wp(\tilde{\nu},\tilde{\varsigma},\tilde{\lambda}), \tilde{\tau}]$
and $\mathcal{E} \neq \tilde{\mathcal E}$. Having fixed a control
level, this Boolean function verifies if the estimates for
${\mathcal E}$ and $\tilde{\mathcal E}$ preserve the relative
positions of the corresponding observations, and we can extend it to
the control sequence.

\begin{df}
\label{def-reliability-estimation-ratio}
Let ${\mathcal E}$ and $\tilde{\mathcal {E}}$ runs on a control
sequence ${\mathcal S}$. We define the {\em reliability estimation
  ratio} {\sc (rer)} {\em of} ${\mathcal E}$ {\em and}
$\tilde{\mathcal E}$ {\em for} ${\mathcal S}$ as
\begin{equation}
\label{equation-rer}
\mbox{\sc rer}(\mathcal{E},\tilde{\mathcal{E}})({\mathcal S}) := 100 *
\frac{\sum_{i \in {\mathcal S}} \mbox{\sc re}(\mathcal{E},\tilde{\mathcal{E}})(i)}{\absd{{\mathcal S}}}
\end{equation}
\end{df}

Although the {\sc rer} covers our requirements to measure the
performance in the second scenario, this is not the case for the first
one, where we need to calculate the number of runs in a set ${\mathcal
  H}$ with regard to which the estimates for a given one $\mathcal{E}$
are reliable on the whole of the control sequence $\mathcal{S}$
considered.

\begin{df}
\label{def-decision-making-reliability}
Let ${\mathcal H} =\{{\mathcal E}_k\}_{k \in K}$ and ${\mathcal E}
\not\in {\mathcal H}$ a set of runs and a run, respectively, on a
control sequence ${\mathcal S}$. We define the {\em decision-making
  reliability ({\sc dmr}) \em of} ${\mathcal E}$ {\em on}
${\mathcal H}$ {\em for} ${\mathcal S}$ as
\begin{equation}
\label{equation-dmr}
\mbox{\sc dmr}(\mathcal{E},\mathcal{H})({\mathcal S}) := 100 *
\frac{\absd{\mathcal{E}_k \in \mathcal{H}, \; \mbox{{\sc
        rer}}(\mathcal{E},\mathcal{E}_k)({\mathcal S}) = 100}} 
     {\absd{\mathcal{S}}}
\end{equation}
\end{df}

Together these metrics offer a good overview of the prediction
reliability achieved. Thus, {\sc mape} gives us a way to
quantitatively evaluate our estimates regardless of the scenario
considered, while {\sc dmr} and {\sc rer} provide a qualitative point
of view. Once a set of runs defined on the same training data base has
been fixed, {\sc dmr} focuses on the reliability of one of these with
respect to the rest along complete control sequences, while {\sc rer}
provides a more realistic view when comparing two runs involving
intersecting learning curves.

\subsubsection{Measuring the robustness}

Since the stability of a run correlates to the degree of monotony in
the asymptotic backbone, a way to measure it is to calculate the
percentage of monotonic elements in the latter.  We are only
interested in those elements computed between the working and the
convergence level, because is in this interval where the approximation
performs effectively.

\begin{df}
\label{def-robustness-rate}
Let ${\mathcal E}$ be a run with asymptotic backbone
$\{\alpha_\ell\}_{\ell \in \mathbb{N}}$, and $\mbox{\em {\sc
    cl}evel}_{\mathcal E}$ and $\mbox{\em {\sc wl}evel}_{\mathcal E}$
its convergence and working levels, respectively. We define the {\em
  robustness rate ({\sc rr}) of} ${\mathcal E}$ as 
\begin{equation}
\label{equation-rr}
\mbox{\sc rr}(\mathcal{E}) := 100 *
\frac{\absd{\mu}}
     {\absd{\{\alpha_i, \; \mbox{\em {\sc wl}evel}_{\mathcal E} \leq i \leq
\mbox{\em {\sc cl}evel}_{\mathcal E}\}}}
\end{equation}

\noindent with $\mu$ the longest maximum monotonic subsequence of
$\{\alpha_i, \; \mbox{\em {\sc wl}evel}_{\mathcal E} \leq i \leq
\mbox{\em {\sc cl}evel}_{\mathcal E}\}$.
\end{df}

Having fixed a run, its tolerance for alterations in the working
hypotheses will be greater, the higher its {\sc rr}. This provides an
efficient criterion for checking the degree of robustness on which we
can count.

\section{The experiments}

\label{section-experiments}

We focus on the prediction of learning curves associated to {\sc
  ml}-based tagger generation, a demanding task in the domain of {\sc
  nlp}, to illustrate our proposal. To that end, we first introduce
the linguistic resources and settings used to later analyze the
results on the basis of the testing frame described. 

\subsection{The linguistics resources and settings}

We need a collection of corpora as training data bases on a number of
target languages, as well as tag-sets, taggers and a methodology to
compute the accuracy values we use for both computing and evaluating
predictions. As target languages we consider English and Spanish. The
former is the most widely studied and best understood one. Spanish is
one of the languages with the highest growth and development potential
in {\sc nlp} and, in contrast to English, it has a complex
derivational paradigm.

\subsubsection{The corpora}

With regard to the corpora, we have selected them together with their
associated tag-sets from the most popular ones in the domain for each
target language:

\begin{enumerate}
\item The {\sc a}n{\sc c}ora~\citep{Taule2008} treebank includes a
  section for Spanish, previously used as a resource in the shared
  tasks of {\sc c}o{\sc nnl}~\citep{Hajic2009} and {\sc s}em{\sc
    e}val~\citep{Recasens2010}. It has served as a training and
  testing resource for {\sc pos} tagging~\citep{Hulden2012},
  parsing~\citep{Popel2013} and semantic annotation~\citep{Mukund2010}
  tasks. Since its tag-set has been developed for languages
  morphologically richer than English, {\sc a}n{\sc c}ora has the most
  detailed annotation of the corpora considered and is the only one to
  follow the {\sc e}agles recommendations~\citep{Monachini1996}. Its
  280 tags~\citep{Taule2008} cover the main {\sc pos} classes used in
  Spanish as well as sub-classes and morphological features,
  accessible on {\em clic.ub.edu/corpus/webfm\_send/18}. The number of
  words in the Spanish section of the treebank is over 515,000.

\item The Freiburg-Brown of American English~\citep{Mair2007} ({\sc
  f}rown) matches the composition and style of the {\sc b}rown
  corpus~\citep{Kucera1967}. It has been used in linguistic studies
  with a more theoretical purpose~\citep{Leech2009,Mair2006}, which in
  our opinion entails an interesting counterpoint to the other two
  corpora considered, which are more oriented to {\sc nlp}-related
  applications. The associated tag-set is the {\sc ucrel} C8. With 169
  tags accessible on {\em ucrel.lancs.ac.uk/claws8tags.pdf}, it was
  selected as the common tag-set for the Brown family of
  corpora~\citep{Hinrichs2010}. The size of the {\sc f}rown corpus is
  over 1,165,000 words.

\item The section with news items from the \textit{Wall Street
  Journal} ({\sc wsj}) included in the {\sc p}enn
  treebank~\citep{Marcus1999}, is a popular corpus in {\sc nlp} for
  both {\sc pos}
  tagging~\citep{Brants2000,Collins2002,Gimenez2004,Ratnaparki1996,Toutanova2003}
  and parsing purposes~\citep{Charniak2000,Petrov2006}. It has also
  been used in the shared tasks of prestigious events in the domain of
  natural language learning, such as {\sc c}o{\sc
    nnl}~\citep{Hajic2009,Nivre2007,Surdeanu2008}, and semantic
  evaluation, such as {\sc s}em{\sc e}val~\citep{Yuret2010}. The
  tag-set associated to this corpus has 45 tags, accessible on {\em
    www.comp.leeds.ac.uk/ccalas/tagsets/upenn.html} and covering basic
  {\sc pos} classes in English along with some morphological
  information. This is the simplest of the tag-sets we consider. The
  {\sc wsj} section is the biggest of our corpora with over 1,170,000
  words.

\end{enumerate}

\noindent Both {\sc p}enn and {\sc a}n{\sc c}ora are treebanks
annotated with {\sc pos} tags as well as syntactic structures. By
stripping them of the latter, they can be used to train {\sc pos}
tagging systems. In order to ensure well-balanced corpora, we also
have scrambled them at sentence level before testing. Thus, the levels
in the learning traces refer to word positions in the scrambled
versions of {\sc a}n{\sc c}ora, {\sc f}rown and {\sc p}enn.

\subsubsection{The {\sc pos} tagging systems}

We avoid rule-based taggers, where the absence of a training phase
leaves the consideration of accuracy as a predictable magnitude void
of content. We then focus on systems built from supervised
learning. In contrast to taggers resulting from unsupervised
techniques, these make it possible to work with predefined tag-sets,
which facilitates both the evaluation and the comprehension of the
results. Furthermore, they have proved to be the best-performing
tagging proposals, placing them as a reference for any testing
purpose. We have selected a broad range of proposals covering the most
representative architectures:

\begin{enumerate}

\item In the category of stochastic methods and as representative of
  the \textit{hidden M\'arkov models} ({\sc hmm}s), we have chosen
  {\sc t}n{\sc t}~\citep{Brants2000}. We also include here the {\sc
    t}ree{\sc t}agger~\citep{Schmid1994}, which uses decision trees to
  generate the {\sc hmm}, and {\sc m}orfette~\citep{Chrupala2008}, an
  averaged perceptron approach~\citep{Collins2002}. To illustrate the
  \textit{maximum entropy models} ({\sc mem}s), we chose to work with
         {\sc mxpost}~\citep{Ratnaparki1996} and the associated to
         {\sc a}pache {\sc o}pen{\sc nlp} ({\sc o}pen{\sc nlp} {\sc
           m}ax{\sc e}nt) (see {\em opennlp.apache.org/}). Finally,
         the {\sc s}tanford {\sc pos} tagger~\citep{Toutanova2003} is
         based on a \textit{conditional M\'arkov model}, which
         combines features of {\sc hmm}s and {\sc mem}s.

\item Under the heading of other {\sc pos} tagging methods, the
  possibilities are many and various. As an example of
  transformation-based learning, we take fn{\sc tbl}~\citep{Ngai2001},
  an updated version of the classic {\sc b}rill~\citep{Brill1995a}. In
  relation to memory-based learning, the representative is the
  \textit{memory-based tagger} ({\sc mbt})~\citep{Daelemans1996a},
  while we chose {\sc svmt}ool~\citep{Gimenez2004} to describe the
  behaviour with respect to a support vector machine
  technique. Finally, we use a perceptron-based training method with
  look-ahead, through {\sc lapos}~\citep{Tsuruoka2011}.
\end{enumerate}

\subsubsection{The testing space}

In order to avoid learning dysfunctions resulting from sentence
truncation, we take a particular class of learning scheme, allowing us
to reap the maximum from the training process. So, given a corpus
$\mathcal D$, a kernel $\mathcal K \subsetneq \mathcal D$ and a step
function $\sigma$, we build the set of individuals $\{\mathcal
D_i\}_{i \in \mathbb{N}}$ as follows:
\begin{equation}
\begin{array}{l}
\hspace*{-.15cm} {\mathcal D}_i := \sentences{{\mathcal C}_i}, \; \forall i \in
\mathbb{N}, \mbox{ such that } {\mathcal C}_1 := {\mathcal K} \mbox{
  and } {\mathcal C}_i := {\mathcal C}_{i-1} \cup {\mathcal I}_{i}, \;
\mathcal I_i \subset {\mathcal D} \setminus {\mathcal C}_{i-1}, \;
\absd{{\mathcal I}_{i}} := \sigma(i), \; \forall i \geq 2
\end{array}
\end{equation}
\noindent where $\sentences{{\mathcal C}_{i}}$ denotes the minimal set of
sentences including ${\mathcal C}_{i}$.

In this way, with respect to the setting of runs, the size of the
kernels is $5*10^3$ words and the constant step function $5*10^3$
locates the instances. We deem both choices conservative, since much
smaller kernels and increases are possible. In addition, we have found
that reasonably good values for the prediction levels can be obtained
with $\nu=2*10^{-5}$, $\varsigma=1$ and $\lambda=5$. Following
previous works on the performance of different models for fitting
learning curves~\citep{Gu:2001:MCP:645940.671380}, particularly in the
{\sc nlp} domain~\citep{Kolachina:2012:PLC:2390524.2390528}, the power
law family has been chosen for $\pi$. As regresion technique for
approximating the partial learning curves we consider the
\textit{trust region method}~\citep{Branch1999}.

Since we are trying to assess the validity or our proposal on finite
corpora, it makes sense to study the prediction capacity within their
boundaries. We then limit the scope in measuring the layer of
convergence introduced in
Definition~\ref{def-layer-of-convergence-learning-trend} for a
learning trend ${\mathcal A}_i^\pi[{\mathcal D}^{\mathcal
    {K}}_{\sigma}]$. Formally, the asymptotic value $\alpha_i$ is
replaced by the one reached at the end of the corpus we are working
on. So, if $\sentencesw{\ell}$ denotes the position of the first
sentence-ending beyond the $\ell$-th word, the layer of convergence
for ${\mathcal A}_i^\pi[{\mathcal D}^{\mathcal {K}}_{\sigma}]$ is now
expressed as follows:
\begin{equation}
\chi^{\ell}({\mathcal A}_i^\pi[{\mathcal D}^{\mathcal {K}}_{\sigma}]) :=
\abs{{\mathcal A}_i^\pi[{\mathcal D}^{\mathcal
      {K}}_{\sigma}](\absd{{\mathcal D}_i}) - \pmb{{\mathcal
    A}_i^\pi[{\mathcal D}^{\mathcal
      {K}}_{\sigma}](\sentencesw{\ell})}}
\end{equation}
\noindent with $\ell=5*10^5$ for runs involving the corpus {\sc
  a}n{\sc c}ora, and $\ell=8*10^5$ for the rest. The term represented
in bold font is the updated one.

The sampling window comprises the interval $[\sentencesw{5*10^3},
  \sentencesw{5*10^5}]$ (resp. $[\sentencesw{5*10^3},
  \sentencesw{8*10^5}]$) for {\sc a}n{\sc c}ora (resp. for {\sc f}rown
and {\sc p}enn), whilst the control levels are taken from control
sequences in $[\sentencesw{3*10^5}, \sentencesw{5*10^5}]$
(resp. $[\sentencesw{3*10^5}, \sentencesw{8*10^5}]$), matching the
location for instances. With the goal of conferring additional
stability on our measures, we opt for a $k$-fold cross
validation~\citep{Clark2010} for computing the samples, due to its
good adaptation to small data sets, an advantage in our context. We
have chosen $k$=10, which has been used before in {\sc pos} tagging
evaluation~\citep{Daelemans1996a,Giesbrecht2009}.

\subsection{The analysis of the results}

As mentioned, our experiments apply from two complementary points of
view, quantitative and qualitative, according in this latter case to
the presence or absence of intersection points between the learning
curves involved in the analysis and generated from a common training
data base.

\subsubsection{The sets of runs}

We start from two collections of runs, ${\mathcal F} =\{{\mathcal
  E}_i\}_{i \in I}$ and ${\mathcal G} =\{{\mathcal E}_j\}_{j \in J}$,
in order to illustrate the predictability of standard and anchoring
learning traces, respectively. Since the use of anchors has been
designed as a mechanism to reinforce the robustness of the
approximations, and Theorem~\ref{th-canonical-anchoring-trace} states
that it can introduce a delay on the convergence, we are particularly
interested in comparing the performance in both cases. We then
consider runs in ${\mathcal G}$ that are \textit{homologous} to those
in ${\mathcal F}$, namely they are exclusively distinguishable by the
use of anchors. The detail of the monitoring is compiled separately
for each case study. So, the collection ${\mathcal F}$
(resp. ${\mathcal G}$) of runs without (resp. with) anchors is shown
in Table~\ref{table-statistics-on-accuracy-no-anchor}
(resp. Table~\ref{table-statistics-on-accuracy-anchor}), also
including in passing the data relative to tests involving disjoint
learning curves along the control sequences. Similarly, the data for
the experiments involving intersecting learning curves for runs in
${\mathcal F}$ (resp. ${\mathcal G}$), can be seen in
Table~\ref{table-statistics-on-relative-positions-no-anchor}
(resp. Table~\ref{table-statistics-on-relative-positions-anchor}). We
include for each run the {\sc pl}evel and the {\sc cl}evel, as well as
the values for Ac and EAc on some of the control levels considered to
later calculate the results for {\sc mape}, {\sc dmr} and {\sc rr}. In
order to facilitate understanding, all those levels are indicated as
previously by their associated word positions in the corpus, which is
denoted by using a superscript {\bf wp} in the identification labels.

These runs combine all the taggers and corpora previously introduced,
though we have discarded some cases due to their high prediction
level, which prevents us from evaluating them from the observations
available. So, the run for {\sc t}ree{\sc t}agger on {\sc a}n{\sc
  c}ora without (resp. with) anchors does not even reach this level on
the interval covered by the sampling set, which is why it has not been
included in the collection ${\mathcal F}$ (resp. ${\mathcal G}$) and
its entry on Table~\ref{table-statistics-on-accuracy-no-anchor}
(resp. Table~\ref{table-statistics-on-accuracy-anchor}) is empty. In
the case of the run for {\sc t}ree{\sc t}agger on {\sc f}rown without
(resp. with) anchors, the prediction level is 795,029
(resp. 800,010). This places us at the limit of our observations,
therefore making it impossible an appropriate evaluation, which also
justifies its exclusion from collection ${\mathcal F}$
(resp. ${\mathcal G}$), as we can see in
Table~\ref{table-statistics-on-accuracy-no-anchor}
(resp. Table~\ref{table-statistics-on-accuracy-anchor}). Consequently,
the entries in
Tables~\ref{table-statistics-on-relative-positions-no-anchor}
and~\ref{table-statistics-on-relative-positions-anchor} for the pairs
in which those runs are involved are discarded. This is the case of
{\sc mbt} and {\sc t}ree{\sc t}agger on {\sc a}n{\sc c}ora, and {\sc
  mxpost} and {\sc t}ree{\sc t}agger on {\sc f}rown.

\subsubsection{The quantitative study}

\begin{figure}[htbp]
\begin{center}
\begin{tabular}{cc}
\hspace*{-.7cm}
\epsfxsize=.52\linewidth
\epsffile{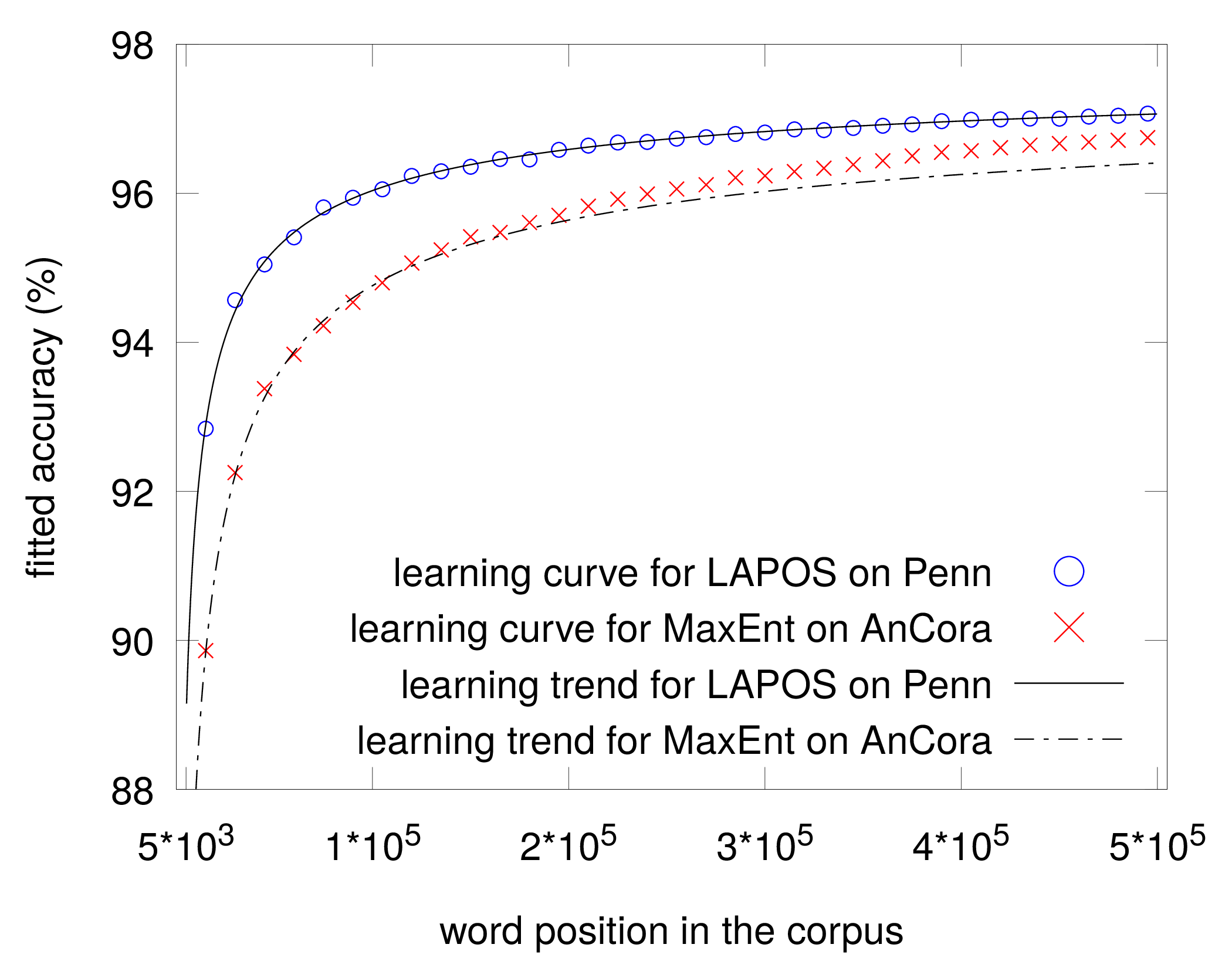}
& 
\hspace*{-.7cm}
\epsfxsize=.52\linewidth
\epsffile{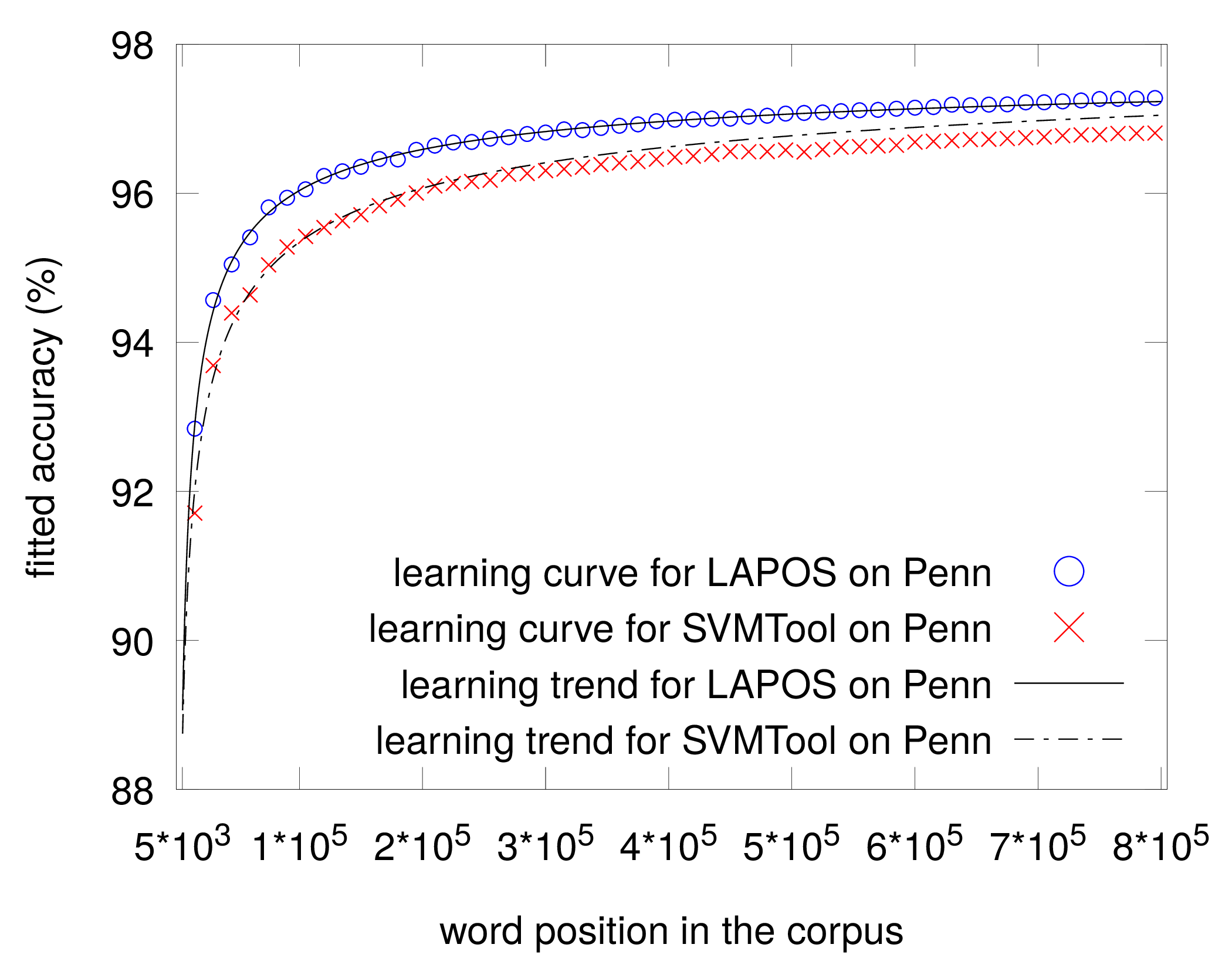}
\end{tabular}
\caption{Learning trends, without anchors, for the best and worst {\sc mape}s.}
\label{fig-best-worst-MAPE-Ac-EAc-500000-800000-no-anchor}
\end{center}
\end{figure}

Our reference performance metric is now the {\sc mape}, and we are
interested in values that are as small as possible. Regarding the
collection of runs ${\mathcal F}$, when anchors are discarded, the
results are shown in Figure~\ref{fig-dmr-mape-statistics-no-anchor},
from the data compiled in
Table~\ref{table-statistics-on-accuracy-no-anchor}. Here, {\sc mape}s
range from 0.01 (resp 0.02) for {\sc lapos} on {\sc p}enn, to 0.30 for
{\sc m}ax{\sc e}nt on {\sc a}n{\sc c}ora (resp. 0.20 for {\sc svmt}ool
on {\sc p}enn) in the interval $[\sentencesw{3*10^5},
  \sentencesw{5*10^5}]$ (resp. $[\sentencesw{3*10^5},
  \sentencesw{8*10^5}]$). Those results are illustrated in
Figure~\ref{fig-best-worst-MAPE-Ac-EAc-500000-800000-no-anchor}, showing
the learning curves and learning trends used for prediction on the
runs with best and worst {\sc mape} on both control sequences. As we
have already done, the observations are generated considering the
portion of the corpus taken from its beginning until the word position
indicated on the horizontal axis. Finally, 57.14\% (resp. 52.63\%) of
{\sc mape} values in this set of runs belong to the interval $[0,
  0.10]$, a proportion that reaches 96.43\% (resp. 100\%) in $[0,
  0.20]$.

\begin{table*}[htbp]
\begin{center}
\begin{scriptsize}
\begin{tabular}{@{\hspace{0pt}}l@{\hspace{3pt}}l@{\hspace{4pt}}
                r@{\hspace{0pt}}c@{\hspace{2pt}}  
                c@{\hspace{0pt}}c@{\hspace{2pt}}  
                c@{\hspace{0pt}}c@{\hspace{2pt}}  
                c@{\hspace{4pt}}c@{\hspace{3pt}}c@{\hspace{4pt}}|@{\hspace{4pt}}                  
                                c@{\hspace{3pt}}c@{\hspace{3pt}}c@{\hspace{2pt}}                  
                c@{\hspace{0pt}}c@{\hspace{2pt}}r@{\hspace{0pt}}c@{\hspace{2pt}}                  
                c@{\hspace{4pt}}c@{\hspace{3pt}}c@{\hspace{4pt}}|@{\hspace{4pt}}                  
                                c@{\hspace{3pt}}c@{\hspace{3pt}}c@{\hspace{2pt}}                  
                c@{\hspace{0pt}}c@{\hspace{2pt}}r@{\hspace{0pt}}c@{\hspace{2pt}}r@{\hspace{0pt}}} 

& & $\mbox{\bf{\scshape pl}evel}^{\tiny \mbox{\bf wp}}$\hspace{4pt} & & \boldmath$\mathbf\tau$ & & $\mbox{\bf{\scshape cl}evel}^{\tiny \mbox{\bf wp}}$ & & \multicolumn{5}{c}{$\mbox{\bf Control-Level}^{\tiny \mbox{\bf wp}}$} & & \bf\scshape mape & & \multicolumn{1}{c}{\bf\scshape dmr} & &\multicolumn{5}{c}{$\mbox{\bf Control-Level}^{\tiny \mbox{\bf wp}}$} & & \bf\scshape mape & & \multicolumn{1}{c}{\bf\scshape dmr} & & \multicolumn{1}{c}{\bf\scshape rr} \\
\cline{3-3} \cline{5-5} \cline{7-7} \cline{9-13} \cline{15-15} \cline{17-17} \cline{19-23} \cline{25-25} \cline{27-27} \cline{29-29}

& & & & & & & & \multicolumn{3}{c}{$\sentencesw{3*10^5}$} & \multicolumn{2}{c}{$\sentencesw{5*10^5}$} & & & & & & \multicolumn{3}{c}{$\sentencesw{6*10^5}$} & \multicolumn{2}{c}{$\sentencesw{8*10^5}$} & & & & & & \rule{0pt}{3.5ex} \\ 
\cline{3-3} \cline{5-5} \cline{7-7} \cline{9-13} \cline{15-15} \cline{17-17} \cline{19-23} \cline{25-25} \cline{27-27} \cline{29-29}

& & & & & & & & \multicolumn{2}{c}{\bf Ac} & \bf EAc & \bf Ac & \bf EAc & & & & & & \multicolumn{2}{c}{\bf Ac} & \bf EAc & \bf Ac & \bf EAc & & & & & & \rule{0pt}{3ex} \\
\cline{3-3} \cline{5-5} \cline{7-7} \cline{9-13} \cline{15-15} \cline{17-17} \cline{19-23} \cline{25-25} \cline{27-27} \cline{29-29}

\multirow{3}{*}{\begin{sideways}{\bf fn\textsc{tbl}}~~\end{sideways}}
& \textsc{a}n\textsc{c}ora & 100,034\hspace{2pt} & & 0.83 & & 230,019 & & & 96.67 & 96.59 & 97.15 & 97.09 & & 0.05 & & 88.89 & & & & & & & & & & \rule{0pt}{3.5ex} & & 88.89 \\
& \textsc{f}rown & 275,023\hspace{2pt} & & 1.50 & & 290,002 & & & 94.98 & 95.09 & 95.78 & 95.90 & & 0.14 & & 100.00 & & & 95.98 & 96.15 & 96.31 & 96.52 & & 0.16 & & 100.00 & & 88.24 \\
& \textsc{p}enn & 95,007\hspace{2pt} & & 0.58 & & 285,005 & & & 96.10 & 96.04 & 96.43 & 96.35 & & 0.05 & & 100.00 & & & 96.53 & 96.45 & 96.72 & 96.58 & & 0.09 & & 87.50 & & 74.36 \\[0.5ex]

\cline{3-3} \cline{5-5} \cline{7-7} \cline{9-13} \cline{15-15} \cline{17-17} \cline{19-23} \cline{25-25} \cline{27-27} \cline{29-29}

\multirow{3}{*}{\begin{sideways}{\bf \textsc{lapos}}~~\end{sideways}}
& \textsc{a}n\textsc{c}ora & 60,019\hspace{2pt} & & 0.36 & & 280,005 & & & 97.55 & 97.47 & 97.90 & 97.78 & & 0.10 & & 100.00 & & & & & & & & & & \rule{0pt}{3.5ex} & & 88.89 \\
& \textsc{f}rown & 90,000\hspace{2pt} & & 1.27 & & 230,005 & & & 96.31 & 96.37 & 96.82 & 96.90 & & 0.07 & & 100.00 & & & 96.98 & 97.07 & 97.21 & 97.30 & & 0.08 & & 100.00 & & 86.21 \\
& \textsc{p}enn & 65,003\hspace{2pt} & & 0.93 & & 140,002 & & & 96.81 & 96.83 & 97.07 & 97.06 & & 0.01 & & 100.00 & & & 97.15 & 97.13 & 97.28 & 97.23 & & 0.02 & & 100.00 & & 50.00 \\[0.5ex]

\cline{3-3} \cline{5-5} \cline{7-7} \cline{9-13} \cline{15-15} \cline{17-17} \cline{19-23} \cline{25-25} \cline{27-27} \cline{29-29}

\multirow{3}{*}{\begin{sideways}{\bf \textsc{m}ax\textsc{e}nt}\end{sideways}}
& \textsc{a}n\textsc{c}ora & 70,009\hspace{2pt} & & 1.07 & & 155,053 & & & 96.23 & 96.02 & 96.77 & 96.41 & & 0.30 & & 77.78 & & & & & & & & & & \rule{0pt}{3.5ex} & & 72.22 \\
& \textsc{f}rown & 160,004\hspace{2pt} & & 1.70 & & 245,011 & & & 94.32 & 94.40 & 95.11 & 95.16 & & 0.08 & & 100.00 & & & 95.33 & 95.39 & 95.69 & 95.73 & & 0.06 & & 100.00 & & 100.00 \\
& \textsc{p}enn & 95,007\hspace{2pt} & & 0.60 & & 270,033 & & & 95.95 & 95.88 & 96.34 & 96.18 & & 0.11 & & 100.00 & & & 96.45 & 96.27 & 96.63 & 96.40 & & 0.17 & & 100.00 & & 100.00 \\[0.5ex]

\cline{3-3} \cline{5-5} \cline{7-7} \cline{9-13} \cline{15-15} \cline{17-17} \cline{19-23} \cline{25-25} \cline{27-27} \cline{29-29}

\multirow{3}{*}{\begin{sideways}{\bf \textsc{mbt}}~~\end{sideways}}
& \textsc{a}n\textsc{c}ora & 70,009\hspace{2pt} & & 0.47 & & 280,005 & & & 96.10 & 96.00 & 96.63 & 96.40 & & 0.18 & & 100.00 & & & & & & & & & & \rule{0pt}{3.5ex} & & 100.00 \\
& \textsc{f}rown & 215,030\hspace{2pt} & & 1.95 & & 255,003 & & & 93.58 & 93.61 & 94.52 & 94.51 & & 0.04 & & 100.00 & & & 94.77 & 94.80 & 95.17 & 95.22 & & 0.04 & & 100.00 & & 100.00 \\
& \textsc{p}enn & 75,035\hspace{2pt} & & 1.66 & & 195,007 & & & 95.24 & 95.32 & 95.76 & 95.91 & & 0.12 & & 100.00 & & & 95.89 & 96.10 & 96.13 & 96.37 & & 0.17 & & 100.00 & & 80.00 \\[0.5ex]

\cline{3-3} \cline{5-5} \cline{7-7} \cline{9-13} \cline{15-15} \cline{17-17} \cline{19-23} \cline{25-25} \cline{27-27} \cline{29-29}

\multirow{3}{*}{\begin{sideways}{\bf \textsc{m}orfette}\end{sideways}}
& \textsc{a}n\textsc{c}ora & 65,035\hspace{2pt} & & 0.73 & & 175,002 & & & 97.18 & 97.03 & 97.52 & 97.33 & & 0.18 & & 77.78 & & & & & & & & & & \rule{0pt}{3.5ex} & & 86.96 \\
& \textsc{f}rown & 100,009\hspace{2pt} & & 1.43 & & 240,053 & & & 95.65 & 95.65 & 96.23 & 96.27 & & 0.04 & & 100.00 & & & 96.39 & 96.47 & 96.69 & 96.75 & & 0.06 & & 100.00 & & 51.72 \\
& \textsc{p}enn & 75,035\hspace{2pt} & & 0.52 & & 210,013 & & & 96.47 & 96.44 & 96.74 & 96.64 & & 0.07 & & 100.00 & & & 96.81 & 96.70 & 96.96 & 96.77 & & 0.11 & & 87.50 & & 96.43 \\[3ex]

\cline{3-3} \cline{5-5} \cline{7-7} \cline{9-13} \cline{15-15} \cline{17-17} \cline{19-23} \cline{25-25} \cline{27-27} \cline{29-29}

\multirow{3}{*}{\begin{sideways}{\bf \textsc{mxpost}}\end{sideways}}
& \textsc{a}n\textsc{c}ora & 80,023\hspace{2pt} & & 1.15 & & 205,013 & & & 96.56 & 96.57 & 97.08 & 97.15 & & 0.04 & & 88.89 & & & & & & & & & & \rule{0pt}{3.5ex} & & 42.31 \\
& \textsc{f}rown & 110,017\hspace{2pt} & & 2.84 & & 150,014 & & & 94.75 & 94.65 & 95.49 & 95.45 & & 0.07 & & 88.89 & & & 95.74 & 95.70 & 96.09 & 96.05 & & 0.05 & & 100.00 & & 88.89 \\
& \textsc{p}enn & 85,013\hspace{2pt} & & 1.40 & & 140,002 & & & 96.11 & 96.18 & 96.52 & 96.54 & & 0.04 & & 100.00 & & & 96.59 & 96.64 & 96.74 & 96.79 & & 0.04 & & 100.00 & & 100.00 \\[0.5ex]

\cline{3-3} \cline{5-5} \cline{7-7} \cline{9-13} \cline{15-15} \cline{17-17} \cline{19-23} \cline{25-25} \cline{27-27} \cline{29-29}

\multirow{3}{*}{\begin{sideways}{\bf \textsc{s}tanford}\end{sideways}}
& \textsc{a}n\textsc{c}ora & 40,010\hspace{2pt} & & 0.51 & & 255,008 & & & 96.86 & 96.72 & 97.31 & 97.09 & & 0.18 & & 88.89 & & & & & & & & & & \rule{0pt}{3.5ex} & & 93.18 \\
& \textsc{f}rown & 120,003\hspace{2pt} & & 1.91 & & 180,021 & & & 95.46 & 95.55 & 96.08 & 96.19 & & 0.13 & & 100.00 & & & 96.27 & 96.39 & 96.56 & 96.68 & & 0.12 & & 77.78 & & 89.47 \\
& \textsc{p}enn & 90,031\hspace{2pt} & & 0.98 & & 150,031 & & & 96.41 & 96.35 & 96.72 & 96.62 & & 0.08 & & 100.00 & & & 96.81 & 96.70 & 96.95 & 96.82 & & 0.11 & & 100.00 & & 53.85 \\[3ex]

\cline{3-3} \cline{5-5} \cline{7-7} \cline{9-13} \cline{15-15} \cline{17-17} \cline{19-23} \cline{25-25} \cline{27-27} \cline{29-29}

\multirow{3}{*}{\begin{sideways}{\bf \textsc{svmt}ool}\end{sideways}}
& \textsc{a}n\textsc{c}ora & 70,009\hspace{2pt} & & 0.76 & & 200,051 & & & 97.03 & 96.91 & 97.47 & 97.30 & & 0.14 & & 87.50 & & & & & & & & & & \rule{0pt}{3.5ex} & & 92.59 \\
& \textsc{f}rown & 205,004\hspace{2pt} & & 1.41 & & 260,002 & & & 95.77 & 95.82 & 96.37 & 96.50 & & 0.12 & & 100.00 & & & 96.54 & 96.70 & 96.79 & 97.01 & & 0.15 & & 100.00 & & 100.00 \\
& \textsc{p}enn & 130,008\hspace{2pt} & & 1.25 & & 155,010 & & & 96.30 & 96.41 & 96.56 & 96.77 & & 0.16 & & 77.78 & & & 96.69 & 96.88 & 96.81 & 97.05 & & 0.20 & & 77.78 & & 100.00 \\[2ex]

\cline{3-3} \cline{5-5} \cline{7-7} \cline{9-13} \cline{15-15} \cline{17-17} \cline{19-23} \cline{25-25} \cline{27-27} \cline{29-29}

\multirow{3}{*}{\begin{sideways}{\bf \textsc{t}n\textsc{t}}~~\end{sideways}}
& \textsc{a}n\textsc{c}ora & 60,019\hspace{2pt} & & 0.77 & & 175,002 & & & 97.11 & 97.03 & 97.47 & 97.35 & & 0.09 & & 100.00 & & & & & & & & & & \rule{0pt}{3.5ex} & & 62.50 \\
& \textsc{f}rown & 95,018\hspace{2pt} & & 1.51 & & 205,004 & & & 95.67 & 95.74 & 96.25 & 96.31 & & 0.09 & & 87.50 & & & 96.42 & 96.48 & 96.63 & 96.74 & & 0.08 & & 87.50 & & 100.00 \\
& \textsc{p}enn & 60,015\hspace{2pt} & & 0.51 & & 230,002 & & & 96.13 & 96.10 & 96.39 & 96.32 & & 0.04 & & 85.71 & & & 96.45 & 96.39 & 96.57 & 96.47 & & 0.06 & & 100.00 & & 45.71 \\[0.5ex]

\cline{3-3} \cline{5-5} \cline{7-7} \cline{9-13} \cline{15-15} \cline{17-17} \cline{19-23} \cline{25-25} \cline{27-27} \cline{29-29}

\multirow{3}{*}{\begin{sideways}{\bf \textsc{t}ree\textsc{t}agger}\end{sideways}}
& \textsc{a}n\textsc{c}ora & ~\hspace{2pt} & & 0.55 & & ~ & & & 96.09 & & 96.67 & & & & & & & & & & & & & & & & & \rule{0pt}{3.5ex} \\
& \textsc{f}rown & 795,029\hspace{2pt} & & 2.00 & & ~ & & & 94.65 & & 95.47 & & & & & & & & 95.75 & & 96.06 & & & & & & & \\
& \textsc{p}enn & 60,015\hspace{2pt} & & 1.32 & & 220,032 & & & 95.13 & 95.07 & 95.83 & 95.60 & & 0.20 & & 100.00 & & & 95.94 & 95.77 & 96.11 & 96.01 & & 0.18 & & 100.00 & & 39.39 \\[6ex]

\cline{3-3} \cline{5-5} \cline{7-7} \cline{9-13} \cline{15-15} \cline{17-17} \cline{19-23} \cline{25-25} \cline{27-27} \cline{29-29}

\end{tabular}
\end{scriptsize}
\end{center}
\caption{Monitoring of runs, without anchors, involving disjoint learning curves
  along the control sequences.}
\label{table-statistics-on-accuracy-no-anchor}
\end{table*}

\begin{figure}[htbp]
\begin{center}
\begin{tabular}{cc}
\hspace*{-.7cm}
\epsfxsize=.52\linewidth
\epsffile{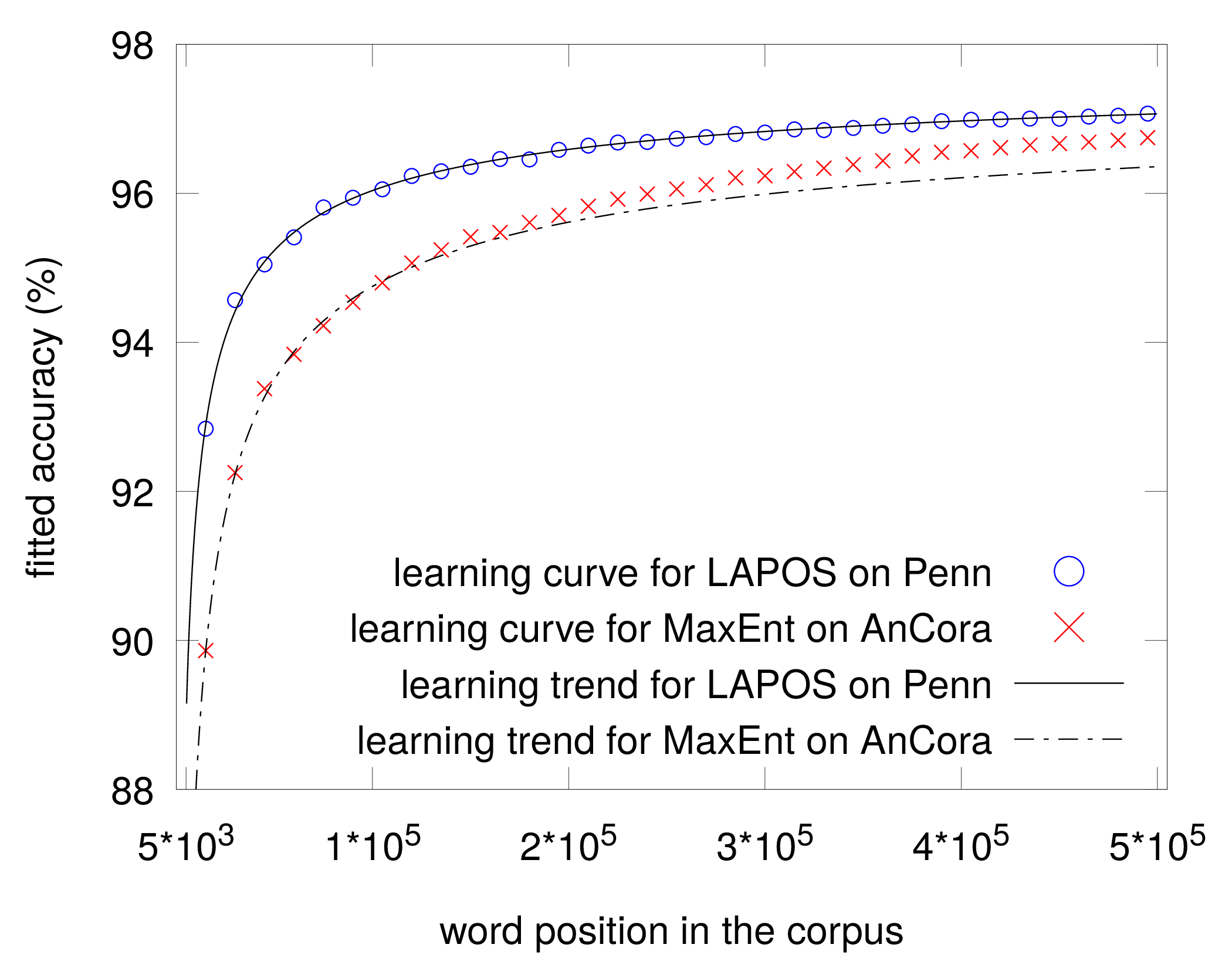}
& 
\hspace*{-.7cm}
\epsfxsize=.52\linewidth
\epsffile{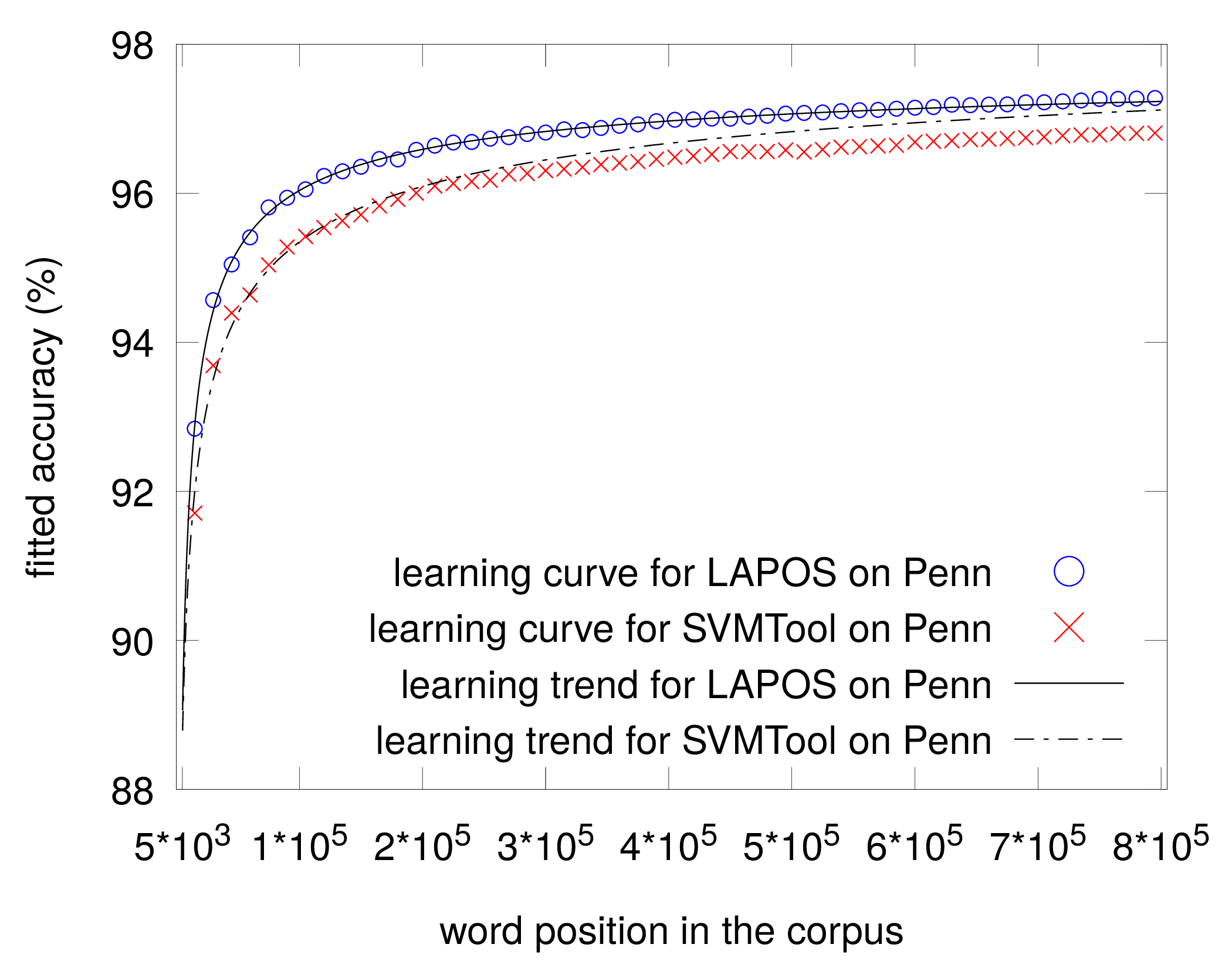}
\end{tabular}
\caption{Learning trends, with anchors, for the best and worst {\sc mape}s.}
\label{fig-best-worst-MAPE-Ac-EAc-500000-800000-anchor}
\end{center}
\end{figure}

Focusing now on the collection of runs ${\mathcal G}$, when anchors
are used, the results are shown in
Figure~\ref{fig-dmr-mape-statistics-anchor} from the data compiled in
Table~\ref{table-statistics-on-accuracy-anchor}. This time the {\sc
  mape}s vary from 0.01 (resp. 0.02) for {\sc lapos} on {\sc p}enn, to
0.35 for {\sc m}ax{\sc e}nt on {\sc a}n{\sc c}ora (resp. 0.26 for {\sc
  svmt}ool on {\sc p}enn) in the interval $[\sentencesw{3*10^5},
  \sentencesw{5*10^5}]$ (resp. $[\sentencesw{3*10^5},
  \sentencesw{8*10^5}]$). The trends involved are shown in
Figure~\ref{fig-best-worst-MAPE-Ac-EAc-500000-800000-anchor}, once again
considering word positions in the text to indicate the portion from the
beginning of the corpus used to generate the observations. Out of the
total of {\sc mape} values for runs in the collection, 53.57\%
(resp. 57.89\%) of them are in the interval $[0, 0.10]$, increasing to
92.86\% (resp. 94.74\%) in the interval $[0, 0.20]$.

While the results are promising for both collections $\mathcal F$ and
$\mathcal G$, which leads us to argue for the goodness of the proposal
on the quantitative plane, we can observe some variations by studying
pairs of homologous runs. We then find that {\sc mape}s are lower in
$\mathcal F$, when anchors are discarded, for 67.86\% (resp 73.68\%)
of those pairs in the interval $[\sentencesw{3*10^5},
  \sentencesw{5*10^5}]$ (resp. $[\sentencesw{3*10^5},
  \sentencesw{8*10^5}]$), and higher only in 10.71\% (resp 15.79\%) of
them. This imbalance is due to the delay in convergence introduced by
anchoring, a mechanism that increases robustness at the price of a
possible slowdown of convergence. Nevertheless, the effects of this
delay seem minor, given that the average difference between the {\sc
  mape}s of homologous runs is 0.02 (resp. 0.03) in the interval
$[\sentencesw{3*10^5}, \sentencesw{5*10^5}]$
(resp. $[\sentencesw{3*10^5}, \sentencesw{8*10^5}]$). At the same
time, a common feature for runs in both $\mathcal F$ and $\mathcal G$
is that, when the length of the corpora allows us to overlay the
control sequences, the {\sc mape}s tend not to increase significantly
between them, illustrating the reliability of calculations.

\begin{table*}[htbp]
\begin{center}
\begin{scriptsize}
\begin{tabular}{@{\hspace{0pt}}l@{\hspace{3pt}}l@{\hspace{4pt}}
                r@{\hspace{0pt}}c@{\hspace{2pt}}  
                c@{\hspace{0pt}}c@{\hspace{2pt}}  
                c@{\hspace{0pt}}c@{\hspace{2pt}}  
                c@{\hspace{4pt}}c@{\hspace{3pt}}c@{\hspace{4pt}}|@{\hspace{4pt}}                  
                                c@{\hspace{3pt}}c@{\hspace{3pt}}c@{\hspace{2pt}}                  
                c@{\hspace{0pt}}c@{\hspace{2pt}}r@{\hspace{0pt}}c@{\hspace{2pt}}                  
                c@{\hspace{4pt}}c@{\hspace{3pt}}c@{\hspace{4pt}}|@{\hspace{4pt}}                  
                                c@{\hspace{3pt}}c@{\hspace{3pt}}c@{\hspace{2pt}}                  
                c@{\hspace{0pt}}c@{\hspace{2pt}}r@{\hspace{0pt}}c@{\hspace{2pt}}r@{\hspace{0pt}}} 

& & $\mbox{\bf{\scshape pl}evel}^{\tiny \mbox{\bf wp}}$\hspace{4pt} & & \boldmath$\mathbf\tau$ & & $\mbox{\bf{\scshape cl}evel}^{\tiny \mbox{\bf wp}}$ & & \multicolumn{5}{c}{$\mbox{\bf Control-Level}^{\tiny \mbox{\bf wp}}$} & & \bf\scshape mape & & \multicolumn{1}{c}{\bf\scshape dmr} & &\multicolumn{5}{c}{$\mbox{\bf Control-Level}^{\tiny \mbox{\bf wp}}$} & & \bf\scshape mape & & \multicolumn{1}{c}{\bf\scshape dmr} & & \multicolumn{1}{c}{\bf\scshape rr} \\
\cline{3-3} \cline{5-5} \cline{7-7} \cline{9-13} \cline{15-15} \cline{17-17} \cline{19-23} \cline{25-25} \cline{27-27} \cline{29-29}

& & & & & & & & \multicolumn{3}{c}{$\sentencesw{3*10^5}$} & \multicolumn{2}{c}{$\sentencesw{5*10^5}$} & & & & & & \multicolumn{3}{c}{$\sentencesw{6*10^5}$} & \multicolumn{2}{c}{$\sentencesw{8*10^5}$} & & & & & & \rule{0pt}{3.5ex} \\ 
\cline{3-3} \cline{5-5} \cline{7-7} \cline{9-13} \cline{15-15} \cline{17-17} \cline{19-23} \cline{25-25} \cline{27-27} \cline{29-29}

& & & & & & & & \multicolumn{2}{c}{\bf Ac} & \bf EAc & \bf Ac & \bf EAc & & & & & & \multicolumn{2}{c}{\bf Ac} & \bf EAc & \bf Ac & \bf EAc & & & & & & \rule{0pt}{3ex} \\
\cline{3-3} \cline{5-5} \cline{7-7} \cline{9-13} \cline{15-15} \cline{17-17} \cline{19-23} \cline{25-25} \cline{27-27} \cline{29-29}

\multirow{3}{*}{\begin{sideways}{\bf fn\textsc{tbl}}~~\end{sideways}}
& \textsc{a}n\textsc{c}ora & 100,034\hspace{2pt} & & 0.83 & & 230,019 & & & 96.67 & 96.58 & 97.15 & 97.08 & & 0.06 & & 88.89 & & & & & & & & & & \rule{0pt}{3.5ex} & & 100.00 \\
& \textsc{f}rown & 295,019\hspace{2pt} & & 1.50 & & 295,019 & & & 94.98 & 95.10 & 95.78 & 95.91 & & 0.15 & & 100.00 & & & 95.98 & 96.17 & 96.31 & 96.55 & & 0.18 & & 100.00 & & 97.14 \\
& \textsc{p}enn & 95,007\hspace{2pt} & & 0.58 & & 285,005 & & & 96.10 & 96.04 & 96.43 & 96.35 & & 0.05 & & 100.00 & & & 96.53 & 96.44 & 96.72 & 96.58 & & 0.09 & & 87.50 & & 100.00 \\[0.5ex]

\cline{3-3} \cline{5-5} \cline{7-7} \cline{9-13} \cline{15-15} \cline{17-17} \cline{19-23} \cline{25-25} \cline{27-27} \cline{29-29}

\multirow{3}{*}{\begin{sideways}{\bf \textsc{lapos}}~~\end{sideways}}
& \textsc{a}n\textsc{c}ora & 60,019\hspace{2pt} & & 0.36 & & 280,005 & & & 97.55 & 97.47 & 97.90 & 97.78 & & 0.10 & & 100.00 & & & & & & & & & & \rule{0pt}{3.5ex} & & 82.22 \\
& \textsc{f}rown & 90,000\hspace{2pt} & & 1.27 & & 230,005 & & & 96.31 & 96.37 & 96.82 & 96.91 & & 0.08 & & 100.00 & & & 96.98 & 97.07 & 97.21 & 97.31 & & 0.09 & & 100.00 & & 72.41 \\
& \textsc{p}enn & 65,003\hspace{2pt} & & 0.93 & & 140,002 & & & 96.81 & 96.83 & 97.07 & 97.06 & & 0.01 & & 100.00 & & & 97.15 & 97.13 & 97.28 & 97.23 & & 0.02 & & 100.00 & & 93.75 \\[0.5ex]

\cline{3-3} \cline{5-5} \cline{7-7} \cline{9-13} \cline{15-15} \cline{17-17} \cline{19-23} \cline{25-25} \cline{27-27} \cline{29-29}

\multirow{3}{*}{\begin{sideways}{\bf \textsc{m}ax\textsc{e}nt}\end{sideways}}
& \textsc{a}n\textsc{c}ora & 70,009\hspace{2pt} & & 1.07 & & 150,022 & & & 96.23 & 95.99 & 96.77 & 96.36 & & 0.35 & & 77.78 & & & & & & & & & & \rule{0pt}{3.5ex} & & 82.35 \\
& \textsc{f}rown & 160,004\hspace{2pt} & & 1.70 & & 250,001 & & & 94.32 & 94.41 & 95.11 & 95.17 & & 0.09 & & 100.00 & & & 95.33 & 95.41 & 95.69 & 95.76 & & 0.08 & & 100.00 & & 100.00 \\
& \textsc{p}enn & 95,007\hspace{2pt} & & 0.60 & & 265,005 & & & 95.95 & 95.87 & 96.34 & 96.17 & & 0.12 & & 100.00 & & & 96.45 & 96.26 & 96.63 & 96.38 & & 0.18 & & 100.00 & & 100.00 \\[0.5ex]

\cline{3-3} \cline{5-5} \cline{7-7} \cline{9-13} \cline{15-15} \cline{17-17} \cline{19-23} \cline{25-25} \cline{27-27} \cline{29-29}

\multirow{3}{*}{\begin{sideways}{\bf \textsc{mbt}}~~\end{sideways}}
& \textsc{a}n\textsc{c}ora & 70,009\hspace{2pt} & & 0.47 & & 280,005 & & & 96.10 & 95.99 & 96.63 & 96.39 & & 0.19 & & 100.00 & & & & & & & & & & \rule{0pt}{3.5ex} & & 100.00 \\
& \textsc{f}rown & 255,003\hspace{2pt} & & 1.95 & & 260,002 & & & 93.58 & 93.64 & 94.52 & 94.56 & & 0.08 & & 100.00 & & & 94.77 & 94.85 & 95.17 & 95.29 & & 0.09 & & 100.00 & & 100.00 \\
& \textsc{p}enn & 75,035\hspace{2pt} & & 1.66 & & 175,019 & & & 95.24 & 95.18 & 95.76 & 95.72 & & 0.06 & & 100.00 & & & 95.89 & 95.89 & 96.13 & 96.13 & & 0.03 & & 100.00 & & 90.48 \\[0.5ex]

\cline{3-3} \cline{5-5} \cline{7-7} \cline{9-13} \cline{15-15} \cline{17-17} \cline{19-23} \cline{25-25} \cline{27-27} \cline{29-29}

\multirow{3}{*}{\begin{sideways}{\bf \textsc{m}orfette}\end{sideways}}
& \textsc{a}n\textsc{c}ora & 65,035\hspace{2pt} & & 0.73 & & 175,002 & & & 97.18 & 97.02 & 97.52 & 97.32 & & 0.19 & & 77.78 & & & & & & & & & & \rule{0pt}{3.5ex} & & 95.65 \\
& \textsc{f}rown & 100,009\hspace{2pt} & & 1.43 & & 240,053 & & & 95.65 & 95.65 & 96.23 & 96.28 & & 0.05 & & 100.00 & & & 96.39 & 96.48 & 96.69 & 96.76 & & 0.07 & & 100.00 & & 62.07 \\
& \textsc{p}enn & 75,035\hspace{2pt} & & 0.52 & & 210,013 & & & 96.47 & 96.44 & 96.74 & 96.64 & & 0.07 & & 100.00 & & & 96.81 & 96.70 & 96.96 & 96.77 & & 0.12 & & 87.50 & & 100.00 \\[3ex]

\cline{3-3} \cline{5-5} \cline{7-7} \cline{9-13} \cline{15-15} \cline{17-17} \cline{19-23} \cline{25-25} \cline{27-27} \cline{29-29}

\multirow{3}{*}{\begin{sideways}{\bf \textsc{mxpost}}\end{sideways}}
& \textsc{a}n\textsc{c}ora & 80,023\hspace{2pt} & & 1.15 & & 205,013 & & & 96.56 & 96.57 & 97.08 & 97.15 & & 0.04 & & 88.89 & & & & & & & & & & \rule{0pt}{3.5ex} & & 65.38 \\
& \textsc{f}rown & 110,017\hspace{2pt} & & 2.84 & & 150,014 & & & 94.75 & 94.66 & 95.49 & 95.47 & & 0.05 & & 88.89 & & & 95.74 & 95.71 & 96.09 & 96.06 & & 0.04 & & 100.00 & & 100.00 \\
& \textsc{p}enn & 85,013\hspace{2pt} & & 1.40 & & 140,002 & & & 96.11 & 96.21 & 96.52 & 96.58 & & 0.08 & & 87.50 & & & 96.59 & 96.68 & 96.74 & 96.84 & & 0.08 & & 87.50 & & 100.00 \\[0.5ex]

\cline{3-3} \cline{5-5} \cline{7-7} \cline{9-13} \cline{15-15} \cline{17-17} \cline{19-23} \cline{25-25} \cline{27-27} \cline{29-29}

\multirow{3}{*}{\begin{sideways}{\bf \textsc{s}tanford}\end{sideways}}
& \textsc{a}n\textsc{c}ora & 40,010\hspace{2pt} & & 0.51 & & 250,008 & & & 96.86 & 96.70 & 97.31 & 97.06 & & 0.20 & & 77.78 & & & & & & & & & & \rule{0pt}{3.5ex} & & 88.37 \\
& \textsc{f}rown & 145,014\hspace{2pt} & & 1.91 & & 190,034 & & & 95.46 & 95.59 & 96.08 & 96.26 & & 0.19 & & 77.78 & & & 96.27 & 96.46 & 96.56 & 96.77 & & 0.19 & & 77.78 & & 95.24 \\
& \textsc{p}enn & 90,031\hspace{2pt} & & 0.98 & & 150,031 & & & 96.41 & 96.36 & 96.72 & 96.63 & & 0.07 & & 100.00 & & & 96.81 & 96.72 & 96.95 & 96.84 & & 0.09 & & 100.00 & & 69.23 \\[3ex]

\cline{3-3} \cline{5-5} \cline{7-7} \cline{9-13} \cline{15-15} \cline{17-17} \cline{19-23} \cline{25-25} \cline{27-27} \cline{29-29}

\multirow{3}{*}{\begin{sideways}{\bf \textsc{svmt}ool}\end{sideways}}
& \textsc{a}n\textsc{c}ora & 70,009\hspace{2pt} & & 0.76 & & 200,051 & & & 97.03 & 96.90 & 97.47 & 97.28 & & 0.16 & & 87.50 & & & & & & & & & & \rule{0pt}{3.5ex} & & 100.00 \\
& \textsc{f}rown & 230,005\hspace{2pt} & & 1.41 & & 265,000 & & & 95.77 & 95.83 & 96.37 & 96.51 & & 0.13 & & 100.00 & & & 96.54 & 96.72 & 96.79 & 97.02 & & 0.17 & & 100.00 & & 100.00 \\
& \textsc{p}enn & 130,008\hspace{2pt} & & 1.25 & & 160,001 & & & 96.30 & 96.45 & 96.56 & 96.83 & & 0.20 & & 77.78 & & & 96.69 & 96.95 & 96.81 & 97.12 & & 0.26 & & 77.78 & & 100.00 \\[2ex]

\cline{3-3} \cline{5-5} \cline{7-7} \cline{9-13} \cline{15-15} \cline{17-17} \cline{19-23} \cline{25-25} \cline{27-27} \cline{29-29}

\multirow{3}{*}{\begin{sideways}{\bf \textsc{t}n\textsc{t}}~~\end{sideways}}
& \textsc{a}n\textsc{c}ora & 60,019\hspace{2pt} & & 0.77 & & 170,007 & & & 97.11 & 97.00 & 97.47 & 97.31 & & 0.13 & & 87.50 & & & & & & & & & & \rule{0pt}{3.5ex} & & 52.17 \\
& \textsc{f}rown & 95,018\hspace{2pt} & & 1.51 & & 210,009 & & & 95.67 & 95.77 & 96.25 & 96.35 & & 0.13 & & 87.50 & & & 96.42 & 96.53 & 96.63 & 96.79 & & 0.12 & & 87.50 & & 100.00 \\
& \textsc{p}enn & 60,015\hspace{2pt} & & 0.51 & & 230,002 & & & 96.13 & 96.10 & 96.39 & 96.32 & & 0.04 & & 85.71 & & & 96.45 & 96.38 & 96.57 & 96.47 & & 0.07 & & 100.00 & & 48.57 \\[0.5ex]

\cline{3-3} \cline{5-5} \cline{7-7} \cline{9-13} \cline{15-15} \cline{17-17} \cline{19-23} \cline{25-25} \cline{27-27} \cline{29-29}

\multirow{3}{*}{\begin{sideways}{\bf \textsc{t}ree\textsc{t}agger}\end{sideways}}
& \textsc{a}n\textsc{c}ora & ~\hspace{2pt} & & 0.55 & & ~ & & & 96.09 & & 96.67 & & & & & & & & & & & & & & & & & \rule{0pt}{3.5ex} \\
& \textsc{f}rown & 800,010\hspace{2pt} & & 2.00 & & ~ & & & 94.65 & & 95.47 & & & & & & & & 95.75 & & 96.06 & & & & & & & \\
& \textsc{p}enn & 60,015\hspace{2pt} & & 1.32 & & 220,032 & & & 95.13 & 95.06 & 95.83 & 95.59 & & 0.21 & & 100.00 & & & 95.94 & 95.76 & 96.11 & 95.99 & & 0.19 & & 100.00 & & 69.70 \\[6ex]

\cline{3-3} \cline{5-5} \cline{7-7} \cline{9-13} \cline{15-15} \cline{17-17} \cline{19-23} \cline{25-25} \cline{27-27} \cline{29-29}

\end{tabular}
\end{scriptsize}
\end{center}
\caption{Monitoring of runs, with anchors, involving disjoint learning curves
  along the control sequences.}
\label{table-statistics-on-accuracy-anchor}
\end{table*}

\subsubsection{The qualitative study}

Our reference performance metrics are here the {\sc dmr} and the {\sc
  rer}, depending on whether the testing scenario considered involves
disjoint learning curves or not. Unlike {\sc mape} values, we are now
interested in those close to 100, the maximum possible for both
metrics.

\paragraph{Runs involving disjoint learning curves} 

\begin{figure}[htbp]
\begin{center}
\hspace*{-1cm}
\epsfxsize=1.025\linewidth
\epsffile{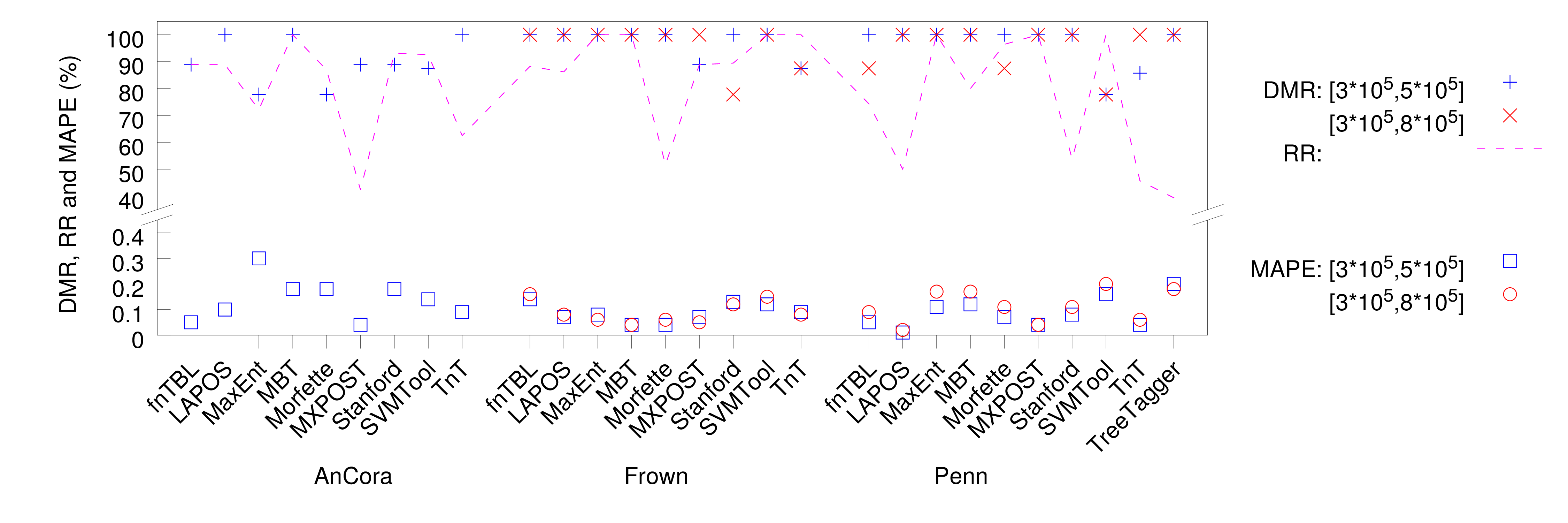}
\caption{{\sc mape}s and {\sc rr}s for runs without anchors. {\sc
    dmr}s when excluding crossing learning curves along the control
  sequences.}
\label{fig-dmr-mape-statistics-no-anchor}
\end{center}
\end{figure}

\begin{figure}[htbp]
\begin{center}
\hspace*{-1cm}
\epsfxsize=1.025\linewidth
\epsffile{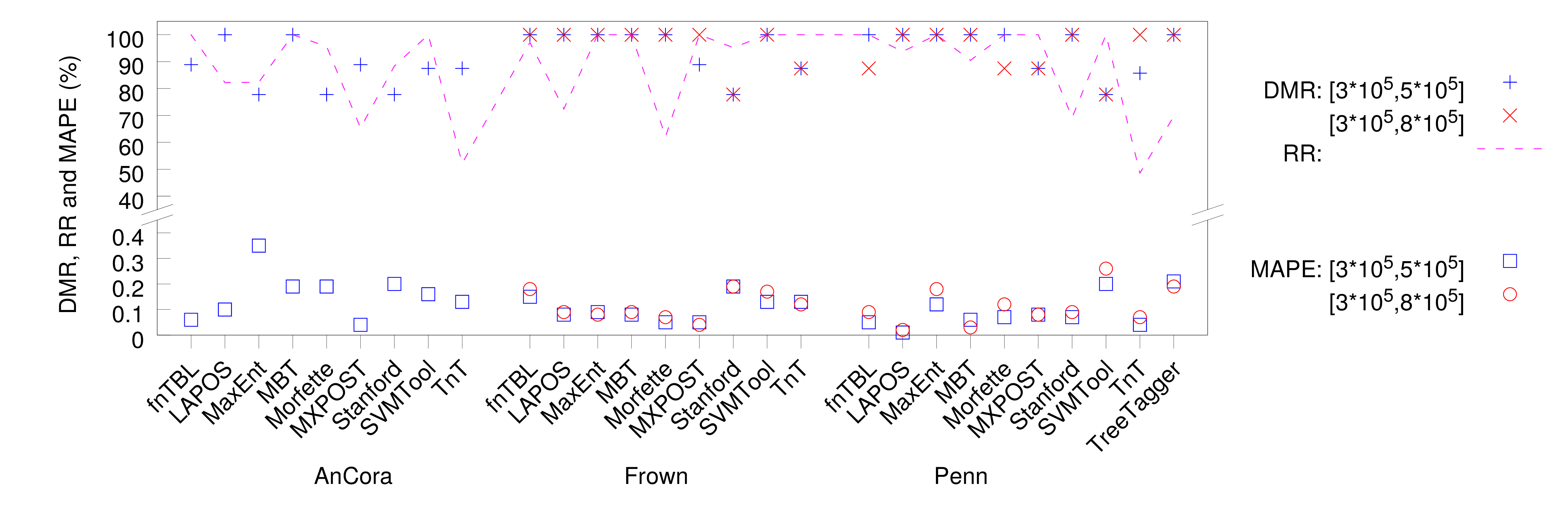}
\caption{{\sc mape}s and {\sc rr}s for runs with anchors. {\sc dmr}s
  when excluding crossing learning curves along the control sequences.}
\label{fig-dmr-mape-statistics-anchor}
\end{center}
\end{figure}

As regards the collection ${\mathcal F}$ of runs based on learning
traces without anchors, {\sc dmr}s in
Figure~\ref{fig-dmr-mape-statistics-no-anchor} are taken from the data
in Table~\ref{table-statistics-on-accuracy-no-anchor} and range from
77.78 to 100 in both control sequences. Moreover, 85.71\%
(resp. 89.47\%) of these values belong to the interval $[87.50, 100]$
for the control sequence in $[\sentencesw{3*10^5},
  \sentencesw{5*10^5}]$ (resp. $[\sentencesw{3*10^5},
  \sentencesw{8*10^5}]$). Similar results, shown in
Figure~\ref{fig-dmr-mape-statistics-anchor} from the data compiled in
Table~\ref{table-statistics-on-accuracy-anchor}, were obtained on the
collection ${\mathcal G}$ of runs with anchoring learning traces.
This time, {\sc dmr} values range between 77.78 and 100 in both
control sequences, with 78.57\% (resp. 89.47\%) of them in the
interval $[87.50, 100]$ for the control levels in
$[\sentencesw{3*10^5}, \sentencesw{5*10^5}]$
(resp. $[\sentencesw{3*10^5}, \sentencesw{8*10^5}]$). The {\sc dmr}s
prove, therefore, to be reasonably high in all the case studies. With
regard to the overlap of control sequences, the values remain stable.

\begin{table*}[htbp]
\begin{center}
\begin{scriptsize}
\begin{tabular}{@{\hspace{0pt}}l@{\hspace{3pt}}l@{\hspace{4pt}}
                r@{\hspace{0pt}}c@{\hspace{2pt}}  
                c@{\hspace{0pt}}c@{\hspace{2pt}}  
                c@{\hspace{0pt}}c@{\hspace{2pt}}  
                c@{\hspace{4pt}}c@{\hspace{3pt}}c@{\hspace{4pt}}|@{\hspace{4pt}}                  
                                c@{\hspace{3pt}}c@{\hspace{3pt}}c@{\hspace{2pt}}                  
                c@{\hspace{0pt}}c@{\hspace{2pt}}r@{\hspace{0pt}}c@{\hspace{2pt}}                  
                c@{\hspace{4pt}}c@{\hspace{3pt}}c@{\hspace{4pt}}|@{\hspace{4pt}}                  
                                c@{\hspace{3pt}}c@{\hspace{3pt}}c@{\hspace{2pt}}                  
                c@{\hspace{0pt}}c@{\hspace{2pt}}r@{\hspace{0pt}}c@{\hspace{2pt}}r@{\hspace{0pt}}} 

& & $\mbox{\bf{\scshape pl}evel}^{\tiny \mbox{\bf wp}}$\hspace{4pt} & & \boldmath$\mathbf\tau$ & & $\mbox{\bf{\scshape cl}evel}^{\tiny \mbox{\bf wp}}$ & & \multicolumn{5}{c}{$\mbox{\bf Control-Level}^{\tiny \mbox{\bf wp}}$} & & \bf\scshape mape & & \multicolumn{1}{c}{\bf\scshape dmr} & &\multicolumn{5}{c}{$\mbox{\bf Control-Level}^{\tiny \mbox{\bf wp}}$} & & \bf\scshape mape & & \multicolumn{1}{c}{\bf\scshape dmr} & & \multicolumn{1}{c}{\bf\scshape rr} \\
\cline{3-3} \cline{5-5} \cline{7-7} \cline{9-13} \cline{15-15} \cline{17-17} \cline{19-23} \cline{25-25} \cline{27-27} \cline{29-29}

& & & & & & & & \multicolumn{3}{c}{$\sentencesw{3*10^5}$} & \multicolumn{2}{c}{$\sentencesw{5*10^5}$} & & & & & & \multicolumn{3}{c}{$\sentencesw{6*10^5}$} & \multicolumn{2}{c}{$\sentencesw{8*10^5}$} & & & & & & \rule{0pt}{3.5ex} \\ 
\cline{3-3} \cline{5-5} \cline{7-7} \cline{9-13} \cline{15-15} \cline{17-17} \cline{19-23} \cline{25-25} \cline{27-27} \cline{29-29}

& & & & & & & & \multicolumn{2}{c}{\bf Ac} & \bf EAc & \bf Ac & \bf EAc & & & & & & \multicolumn{2}{c}{\bf Ac} & \bf EAc & \bf Ac & \bf EAc & & & & & & \rule{0pt}{3ex} \\
\cline{3-3} \cline{5-5} \cline{7-7} \cline{9-13} \cline{15-15} \cline{17-17} \cline{19-23} \cline{25-25} \cline{27-27} \cline{29-29}

\multirow{2}{*}{\begin{sideways}{\bf \textsc{a}n\textsc{c}ora}\end{sideways}}
& \textsc{svmt}ool & 70,009\hspace{2pt} & & 0.76 & & 200,051 & & & 97.03 & 96.91 & 97.47 & 97.30 & & 0.14 & & \multirow{2}{*}{92.68} & & & & & & & & & & & & \rule{0pt}{3.5ex} 92.59 \\
& \textsc{t}n\textsc{t} & 60,019\hspace{2pt} & & 0.77 & & 175,002 & & & 97.11 & 97.03 & 97.47 & 97.35 & & 0.09 & & & & & & & & & & & & & & 62.50 \\[3.5ex]

\cline{3-3} \cline{5-5} \cline{7-7} \cline{9-13} \cline{15-15} \cline{17-17} \cline{19-23} \cline{25-25} \cline{27-27} \cline{29-29}

\multirow{2}{*}{\begin{sideways}{\bf \textsc{f}rown}\end{sideways}}
& \textsc{m}orfette & 100,009\hspace{2pt} & & 1.43 & & 240,053 & & & 95.65 & 95.65 & 96.23 & 96.27 & & 0.04 & & \multirow{2}{*}{80.49} & & & 96.39 & 96.47 & 96.69 & 96.75 & & 0.06 & & \multirow{2}{*}{79.21} & & \rule{0pt}{3.5ex} 51.72 \\
& \textsc{t}n\textsc{t} & 95,018\hspace{2pt} & & 1.51 & & 205,004 & & & 95.67 & 95.74 & 96.25 & 96.31 & & 0.09 & & & & & 96.42 & 96.48 & 96.63 & 96.74 & & 0.08 & & & & 100.00 \\[3.5ex]

\cline{3-3} \cline{5-5} \cline{7-7} \cline{9-13} \cline{15-15} \cline{17-17} \cline{19-23} \cline{25-25} \cline{27-27} \cline{29-29}

\multirow{10}{*}{\begin{sideways}{\bf \textsc{p}enn}~~~~~\end{sideways}}
& fn\textsc{tbl} & 95,007\hspace{2pt} & & 0.58 & & 285,005 & & & 96.10 & 96.04 & 96.43 & 96.35 & & 0.05 & & \multirow{2}{*}{87.80} & & & 96.53 & 96.45 & 96.72 & 96.58 & & 0.09 & & \multirow{2}{*}{95.05} & & \rule{0pt}{3.5ex} 74.36 \\
& \textsc{t}n\textsc{t} & 60,015\hspace{2pt} & & 0.51 & & 230,002 & & & 96.13 & 96.10 & 96.39 & 96.32 & & 0.04 & & & & & 96.45 & 96.39 & 96.57 & 96.47 & & 0.06 & & & & 45.71 \\

& \textsc{m}ax\textsc{e}nt & 95,007\hspace{2pt} & & 0.60 & & 270,033 & & & 95.95 & 95.88 & 96.34 & 96.18 & & 0.11 & & & & & 96.45 & 96.27 & 96.63 & 96.40 & & 0.17 & & \multirow{2}{*}{58.42} & & \rule{0pt}{3.5ex} 100.00 \\
& \textsc{t}n\textsc{t} & 60,015\hspace{2pt} & & 0.51 & & 230,002 & & & 96.13 & 96.10 & 96.39 & 96.32 & & 0.04 & & & & & 96.45 & 96.39 & 96.57 & 96.47 & & 0.06 & & & & 45.71 \\

& \textsc{mbt} & 75,035\hspace{2pt} & & 1.66 & & 195,007 & & & 95.24 & 95.32 & 95.76 & 95.91 & & 0.12 & & \multirow{2}{*}{31.71} & & & 95.89 & 96.10 & 96.13 & 96.37 & & 0.17 & & \multirow{2}{*}{37.62} & & \rule{0pt}{3.5ex} 80.00 \\
& \textsc{t}ree\textsc{t}agger & 60,015\hspace{2pt} & & 1.32 & & 220,032 & & & 95.13 & 95.07 & 95.83 & 95.60 & & 0.20 & & & & & 95.94 & 95.77 & 96.11 & 96.01 & & 0.18 & & & & 39.39 \\

& \textsc{m}orfette & 75,035\hspace{2pt} & & 0.52 & & 210,013 & & & 96.47 & 96.44 & 96.74 & 96.64 & & 0.07 & & \multirow{2}{*}{97.56} & & & 96.81 & 96.70 & 96.96 & 96.77 & & 0.11 & & \multirow{2}{*}{77.23} & & \rule{0pt}{3.5ex} 96.43 \\
& \textsc{s}tanford & 90,031\hspace{2pt} & & 0.98 & & 150,031 & & & 96.41 & 96.35 & 96.72 & 96.62 & & 0.08 & & & & & 96.81 & 96.70 & 96.95 & 96.82 & & 0.11 & & & & 53.85 \\

& \textsc{mxpost} & 85,013\hspace{2pt} & & 1.40 & & 140,002 & & & 96.11 & 96.18 & 96.52 & 96.54 & & 0.04 & & \multirow{2}{*}{97.56} & & & 96.59 & 96.64 & 96.74 & 96.79 & & 0.04 & & \multirow{2}{*}{99.01} & & \rule{0pt}{3.5ex} 100.00 \\
& \textsc{t}n\textsc{t} & 60,015\hspace{2pt} & & 0.51 & & 230,002 & & & 96.13 & 96.10 & 96.39 & 96.32 & & 0.04 & & & & & 96.45 & 96.39 & 96.57 & 96.47 & & 0.06 & & & & 45.71 \\[0.5ex]

\cline{3-3} \cline{5-5} \cline{7-7} \cline{9-13} \cline{15-15} \cline{17-17} \cline{19-23} \cline{25-25} \cline{27-27} \cline{29-29}

\end{tabular}
\end{scriptsize}
\end{center}
\caption{Monitoring of pairs of runs, without anchors, involving crossing
  learning curves along the control sequences.}
\label{table-statistics-on-relative-positions-no-anchor}
\end{table*}

\begin{table*}[htbp]
\begin{center}
\begin{scriptsize}
\begin{tabular}{@{\hspace{0pt}}l@{\hspace{3pt}}l@{\hspace{4pt}}
                r@{\hspace{0pt}}c@{\hspace{2pt}}  
                c@{\hspace{0pt}}c@{\hspace{2pt}}  
                c@{\hspace{0pt}}c@{\hspace{2pt}}  
                c@{\hspace{4pt}}c@{\hspace{3pt}}c@{\hspace{4pt}}|@{\hspace{4pt}}                  
                                c@{\hspace{3pt}}c@{\hspace{3pt}}c@{\hspace{2pt}}                  
                c@{\hspace{0pt}}c@{\hspace{2pt}}r@{\hspace{0pt}}c@{\hspace{2pt}}                  
                c@{\hspace{4pt}}c@{\hspace{3pt}}c@{\hspace{4pt}}|@{\hspace{4pt}}                  
                                c@{\hspace{3pt}}c@{\hspace{3pt}}c@{\hspace{2pt}}                  
                c@{\hspace{0pt}}c@{\hspace{2pt}}r@{\hspace{0pt}}c@{\hspace{2pt}}r@{\hspace{0pt}}} 

& & $\mbox{\bf{\scshape pl}evel}^{\tiny \mbox{\bf wp}}$\hspace{4pt} & & \boldmath$\mathbf\tau$ & & $\mbox{\bf{\scshape cl}evel}^{\tiny \mbox{\bf wp}}$ & & \multicolumn{5}{c}{$\mbox{\bf Control-Level}^{\tiny \mbox{\bf wp}}$} & & \bf\scshape mape & & \multicolumn{1}{c}{\bf\scshape dmr} & &\multicolumn{5}{c}{$\mbox{\bf Control-Level}^{\tiny \mbox{\bf wp}}$} & & \bf\scshape mape & & \multicolumn{1}{c}{\bf\scshape dmr} & & \multicolumn{1}{c}{\bf\scshape rr} \\
\cline{3-3} \cline{5-5} \cline{7-7} \cline{9-13} \cline{15-15} \cline{17-17} \cline{19-23} \cline{25-25} \cline{27-27} \cline{29-29}

& & & & & & & & \multicolumn{3}{c}{$\sentencesw{3*10^5}$} & \multicolumn{2}{c}{$\sentencesw{5*10^5}$} & & & & & & \multicolumn{3}{c}{$\sentencesw{6*10^5}$} & \multicolumn{2}{c}{$\sentencesw{8*10^5}$} & & & & & & \rule{0pt}{3.5ex} \\ 
\cline{3-3} \cline{5-5} \cline{7-7} \cline{9-13} \cline{15-15} \cline{17-17} \cline{19-23} \cline{25-25} \cline{27-27} \cline{29-29}

& & & & & & & & \multicolumn{2}{c}{\bf Ac} & \bf EAc & \bf Ac & \bf EAc & & & & & & \multicolumn{2}{c}{\bf Ac} & \bf EAc & \bf Ac & \bf EAc & & & & & & \rule{0pt}{3ex} \\
\cline{3-3} \cline{5-5} \cline{7-7} \cline{9-13} \cline{15-15} \cline{17-17} \cline{19-23} \cline{25-25} \cline{27-27} \cline{29-29}

\multirow{2}{*}{\begin{sideways}{\bf \textsc{a}n\textsc{c}ora}\end{sideways}}
& \textsc{svmt}ool & 70,009\hspace{2pt} & & 0.76 & & 200,051 & & & 97.03 & 96.90 & 97.47 & 97.28 & & 0.16 & & \multirow{2}{*}{92.68} & & & & & & & & & & & & \rule{0pt}{3.5ex} 100.00 \\
& \textsc{t}n\textsc{t} & 60,019\hspace{2pt} & & 0.77 & & 170,007 & & & 97.11 & 97.00 & 97.47 & 97.31 & & 0.13 & & & & & & & & & & & & & & 52.17 \\[3.5ex]

\cline{3-3} \cline{5-5} \cline{7-7} \cline{9-13} \cline{15-15} \cline{17-17} \cline{19-23} \cline{25-25} \cline{27-27} \cline{29-29}

\multirow{2}{*}{\begin{sideways}{\bf \textsc{f}rown}\end{sideways}}
& \textsc{m}orfette & 100,009\hspace{2pt} & & 1.43 & & 240,053 & & & 95.65 & 95.65 & 96.23 & 96.28 & & 0.05 & & \multirow{2}{*}{80.49} & & & 96.39 & 96.48 & 96.69 & 96.76 & & 0.07 & & \multirow{2}{*}{74.26} & & \rule{0pt}{3.5ex} 62.07 \\
& \textsc{t}n\textsc{t} & 95,018\hspace{2pt} & & 1.51 & & 210,009 & & & 95.67 & 95.77 & 96.25 & 96.35 & & 0.13 & & & & & 96.42 & 96.53 & 96.63 & 96.79 & & 0.12 & & & & 100.00 \\[3.5ex]

\cline{3-3} \cline{5-5} \cline{7-7} \cline{9-13} \cline{15-15} \cline{17-17} \cline{19-23} \cline{25-25} \cline{27-27} \cline{29-29}

\multirow{10}{*}{\begin{sideways}{\bf \textsc{p}enn}~~~~~\end{sideways}}
& fn\textsc{tbl} & 95,007\hspace{2pt} & & 0.58 & & 285,005 & & & 96.10 & 96.04 & 96.43 & 96.35 & & 0.05 & & \multirow{2}{*}{87.80} & & & 96.53 & 96.44 & 96.72 & 96.58 & & 0.09 & & \multirow{2}{*}{95.05} & & \rule{0pt}{3.5ex} 100.00 \\
& \textsc{t}n\textsc{t} & 60,015\hspace{2pt} & & 0.51 & & 230,002 & & & 96.13 & 96.10 & 96.39 & 96.32 & & 0.04 & & & & & 96.45 & 96.38 & 96.57 & 96.47 & & 0.07 & & & & 48.57 \\

& \textsc{m}ax\textsc{e}nt & 95,007\hspace{2pt} & & 0.60 & & 265,005 & & & 95.95 & 95.87 & 96.34 & 96.17 & & 0.12 & & & & & 96.45 & 96.26 & 96.63 & 96.38 & & 0.18 & & \multirow{2}{*}{58.42} & & \rule{0pt}{3.5ex} 100.00 \\
& \textsc{t}n\textsc{t} & 60,015\hspace{2pt} & & 0.51 & & 230,002 & & & 96.13 & 96.10 & 96.39 & 96.32 & & 0.04 & & & & & 96.45 & 96.38 & 96.57 & 96.47 & & 0.07 & & & & 48.57 \\

& \textsc{mbt} & 75,035\hspace{2pt} & & 1.66 & & 175,019 & & & 95.24 & 95.18 & 95.76 & 95.72 & & 0.06 & & \multirow{2}{*}{31.71} & & & 95.89 & 95.89 & 96.13 & 96.13 & & 0.03 & & \multirow{2}{*}{37.62} & & \rule{0pt}{3.5ex} 90.48 \\
& \textsc{t}ree\textsc{t}agger & 60,015\hspace{2pt} & & 1.32 & & 220,032 & & & 95.13 & 95.06 & 95.83 & 95.59 & & 0.21 & & & & & 95.94 & 95.76 & 96.11 & 95.99 & & 0.19 & & & & 69.70 \\

& \textsc{m}orfette & 75,035\hspace{2pt} & & 0.52 & & 210,013 & & & 96.47 & 96.44 & 96.74 & 96.64 & & 0.07 & & \multirow{2}{*}{97.56} & & & 96.81 & 96.70 & 96.96 & 96.77 & & 0.12 & & \multirow{2}{*}{73.27} & & \rule{0pt}{3.5ex} 100.00 \\
& \textsc{s}tanford & 90,031\hspace{2pt} & & 0.98 & & 150,031 & & & 96.41 & 96.36 & 96.72 & 96.63 & & 0.07 & & & & & 96.81 & 96.72 & 96.95 & 96.84 & & 0.09 & & & & 69.23 \\

& \textsc{mxpost} & 85,013\hspace{2pt} & & 1.40 & & 140,002 & & & 96.11 & 96.21 & 96.52 & 96.58 & & 0.08 & & \multirow{2}{*}{97.56} & & & 96.59 & 96.68 & 96.74 & 96.84 & & 0.08 & & \multirow{2}{*}{99.01} & & \rule{0pt}{3.5ex} 100.00 \\
& \textsc{t}n\textsc{t} & 60,015\hspace{2pt} & & 0.51 & & 230,002 & & & 96.13 & 96.10 & 96.39 & 96.32 & & 0.04 & & & & & 96.45 & 96.38 & 96.57 & 96.47 & & 0.07 & & & & 48.57 \\[0.5ex]

\cline{3-3} \cline{5-5} \cline{7-7} \cline{9-13} \cline{15-15} \cline{17-17} \cline{19-23} \cline{25-25} \cline{27-27} \cline{29-29}

\end{tabular}
\end{scriptsize}
\end{center}
\caption{Monitoring of pairs of runs, with anchors, involving crossing
  learning curves along the control sequences.}
\label{table-statistics-on-relative-positions-anchor}
\end{table*}

Finally, the delay in convergence observed in runs of $\mathcal G$
with respect to their homologues in $\mathcal F$ has a limited impact
on {\sc dmr}s, supporting again the use of anchoring as practical
mechanism to improve the robustness. Thus, only a 14.29\%
(resp. 5.26\%) of homologous {\sc dmr}s are higher for runs without
anchors in the interval $[\sentencesw{3*10^5}, \sentencesw{5*10^5}]$
(resp. $[\sentencesw{3*10^5}, \sentencesw{8*10^5}]$), while the
remaining 85.71\% (resp. 94.73\%) have identical values. That reduces
the average difference between homologous {\sc dmr}s to 2.08
(resp. 0.66).

\paragraph{Runs involving intersecting learning curves}

In the case of the collection ${\mathcal F}$ of runs based on learning
traces without anchors, {\sc rer}s for pairs in
Figure~\ref{fig-rer-mape-statistics-no-anchor} are taken from the data
in Table~\ref{table-statistics-on-relative-positions-no-anchor} and
range from 31.71 (resp. 37.62) to 97.56 (resp. 99.01) for the
control sequence in $[\sentencesw{3*10^5}, \sentencesw{5*10^5}]$
(resp. $[\sentencesw{3*10^5}, \sentencesw{8*10^5}]$). In addition,
while 66.67\% (resp. 33.33\%) of those values can be found in the
interval $[87.80,100]$, 83.33\% (resp. 66.67\%) of them are in
$[77.23,100]$. Results are similar for pairs of runs with anchoring
learning traces in the collection ${\mathcal G}$, which are shown in
Figure~\ref{fig-rer-mape-statistics-anchor} from the data compiled in
Table~\ref{table-statistics-on-relative-positions-anchor}. Maximum and
minimum {\sc rer} values are the same as those obtained without
anchors in both control sequences. Moreover, 66.67\% (resp. 33.33\%)
of them can be found in the interval $[87.80,100]$, climbing to
83.33\% (resp. 66.67\%) for the interval $[73.27,100]$, for the
control sequence in $[\sentencesw{3*10^5}, \sentencesw{5*10^5}]$
(resp. $[\sentencesw{3*10^5}, \sentencesw{8*10^5}]$). When the corpora
cover both control sequences, there is no significant difference
between their corresponding {\sc rer}s, confirming the reliability of
the calculations, which are also reasonably high for {\sc rer} in all
the case studies. In short, we can also argue the goodness of the
proposal on the qualitative plane.

Following the previous discussion for {\sc mape} and {\sc dmr}, the
impact of using anchoring also remains limited for {\sc rer}, allowing
us to definitively state its viability as an instrument for bettering
robustness. To prove this, we consider pairs of {\sc rer}s, one
computed for two runs in $\mathcal F$ with intersecting learning
traces, and the other for their homologues in $\mathcal G$. In all
cases (resp. 66.67\% of them) the {\sc rer}s are identical for the
control sequence in $[\sentencesw{3*10^5}, \sentencesw{5*10^5}]$
(resp. $[\sentencesw{3*10^5}, \sentencesw{8*10^5}]$). The remaining
33.33\% of pairs measured in $[\sentencesw{3*10^5},
  \sentencesw{8*10^5}]$ show higher {\sc rer} values for runs in
$\mathcal F$, namely without anchors, resulting in an average
difference between {\sc rer}s in the same pair of 1.48 in
$[\sentencesw{3*10^5}, \sentencesw{8*10^5}]$.

\subsubsection{The study of robustness}

\begin{figure}[htbp]
\begin{center}
\hspace*{-1cm}
\epsfxsize=1.025\linewidth
\epsffile{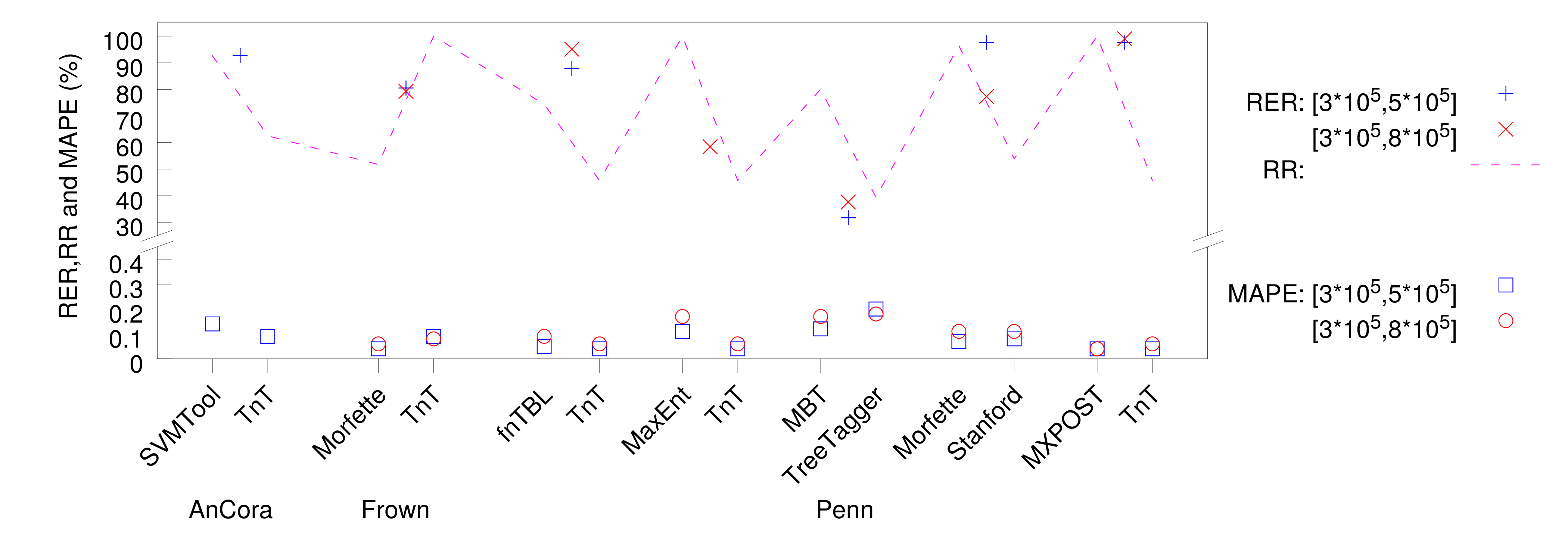}
\caption{{\sc mape}s, {\sc rr}s and {\sc rer}s for pairs of runs, without
  anchors, involving crossing learning curves along the control
  sequences.}
\label{fig-rer-mape-statistics-no-anchor}
\end{center}
\end{figure}

\begin{figure}[htbp]
\begin{center}
\hspace*{-1cm}
\epsfxsize=1.025\linewidth
\epsffile{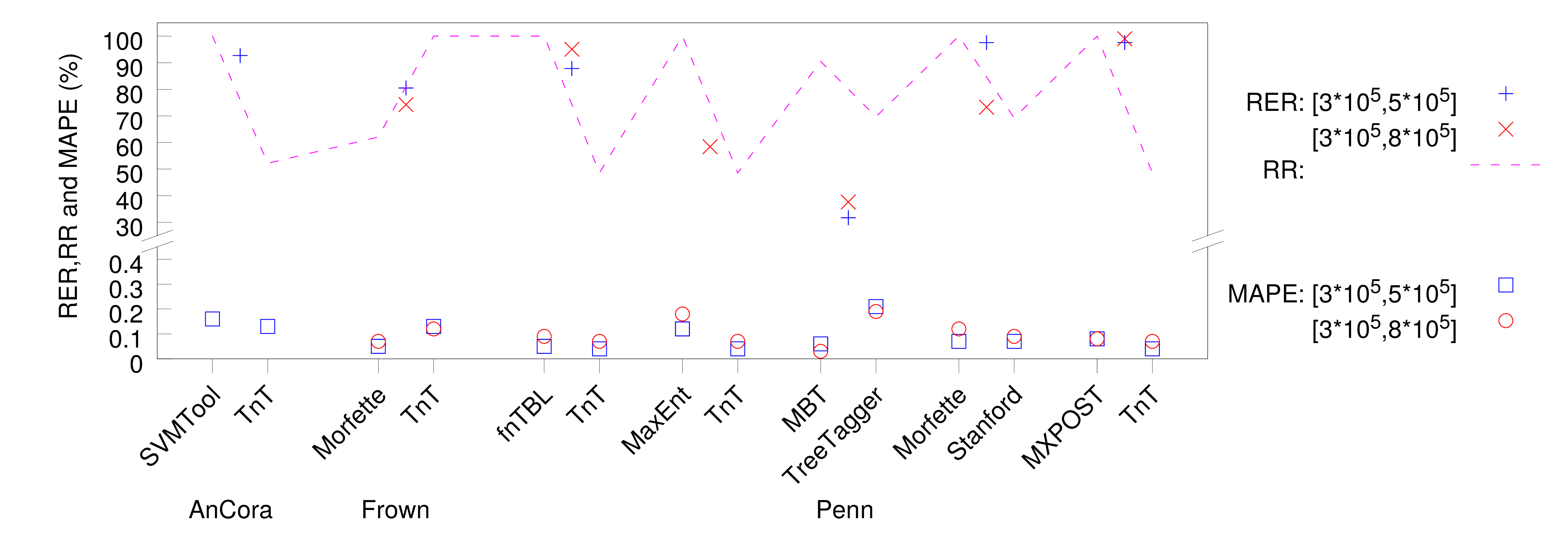}
\caption{{\sc mape}s, {\sc rr}s and {\sc rer}s for pairs of runs, with
  anchors, involving crossing learning curves along the control sequences.}
\label{fig-rer-mape-statistics-anchor}
\end{center}
\end{figure}

The reference metric is now {\sc rr}, and we are interested in values
as close as possible to 100, the maximum one.  Regarding the
collection of runs ${\mathcal F}$, when anchors are discarded, the
results are shown in both
Figs.~\ref{fig-dmr-mape-statistics-no-anchor}
and~\ref{fig-rer-mape-statistics-no-anchor} from the data compiled in
Table~\ref{table-statistics-on-accuracy-no-anchor}. While {\sc rr}
values range from 39.39 to 100, the latter is only reached in 28.57\%
of the runs. This percentage rises to 39.29\% for {\sc rr}s in the
interval $[90,100]$, reaching 67.86\% in $[80,100]$. Focusing now on
the collection of runs ${\mathcal G}$, with anchoring learning traces,
the results are shown in Figs.~\ref{fig-dmr-mape-statistics-anchor}
and~\ref{fig-rer-mape-statistics-anchor} from the data compiled in
Table~\ref{table-statistics-on-accuracy-anchor}. Minimum and maximum
{\sc rr}s are, respectively, 48.57 and 100, with 46.43\% of the runs
achieving the maximum value. This proportion reaches 64.29\% in the
interval $[90,100]$, and 75\% in $[80,100]$. With respect to the
behaviour of {\sc rr}s in pairs of homologous runs, higher values are
found when using anchoring in 57.14\% of those pairs, and lower ones
only in 14.29\% of them. The average difference between {\sc rr}
values for homologous runs is 9.43\%. These results illustrate not
only the good overall behaviour of the prediction model against
variations in its working hypotheses, but also the positive role that
the use of anchors play in this regard.

\section{Conclusions}
\label{section-conclusions}

Our proposal arises as a response to the challenge of reducing both
the training effort and the need for linguistic resources, in the
generation of learning-based {\sc pos} tagging systems in the {\sc
  nlp} domain. In this context, we introduce a prediction strategy for
learning curves, which we can exploit to deal with a variety of
practical uses beyond the mere estimation for the final value of the
accuracy associated to a tagger. Focusing on the most representative
ones, it is possible to estimate the extra accuracy increase between
any two levels in the corpus, which is helpful for evaluating the
training effort needed to attain a certain measurement
performance. Comparing the latter also becomes realistic at all
training levels, providing us with a useful instrument for choosing
the most efficient tagger in each case. Finally, accuracy prediction
below a certain degree of convergence fixed by the user can be
guaranteed, which gives us the possibility of evaluating the adequacy
of tagger configuration on the basis of a fraction of its generation
process. Altogether, these facilities involve both quantitative and
qualitative aspects, forming a powerful tool for reducing training
costs in tagger construction.

Formally, we have developed a technique whose generality permits it to
be applied in a much wider context. Based on a functional analysis, we
extend the classic discrete calculation of accuracy in {\sc ml} to a
continuous domain. The proposal is modelled as the uniform convergence
of a sequence of learning trends which iteratively approximates the
learning curve. Since the limit computed is a continuous curve, it is
guaranteed to be free of gaps, breaks and holes. This allows us to
make the predictions without disruptions due to instantaneous jumps
and over the entire training data base, while ensuring their
regularity. The correctness of the algorithm has been proved,
including a proximity criterion, with respect to our working
hypotheses. This permits us, once a point in the process called
prediction level has been passed, to identify the iteration from
which the estimates are below a convergence threshold fixed by the
user.

Regarding the robustness and given that the monotony of the asymptotic
backbone is at the basis of the correctness, our goal is to reduce the
fluctuations in that sequence, the anchoring of the learning trends
being the way to achieve this. With the aim of maximizing the
efficiency of this mechanism, the user can fix a verticality threshold
for determining the working level from which it is applied. Whilst
this is not a mandatory procedure, its implementation enables us to
limit the use of anchors to that part of the process where the slopes
are slight enough, avoiding an unnecessary deceleration of the
convergence.

In practice, the experimental results in the {\sc nlp} field of {\sc
  pos} tagging corroborate our expectations on a wide range of
particular cases for a representative sampling of taggers and corpora
on both English and Spanish. This opens the door to new applications
in {\sc ml}, particularly in the {\sc nlp} domain, this being the case
in a variety of well known tasks, such as {\sc mt}, text
classification, parsing or any other kind of activity requiring
linguistic annotation. All these are new fields of application we plan
to explore in a future work.

\section*{Acknowledgments}

\begin{small}
We thank Prof. Christian Mair, of the University of Freiburg, who
provided us with a copy of the tagged versions of the Brown family of
corpora, including the Freiburg-Brown corpus of American English. This
research has been partially funded by the Spanish Ministry of Economy
and Competitiveness through project FFI2014-51978-C2-1-R.
\end{small}

\begin{footnotesize}

\end{footnotesize}

\end{document}